\documentclass{article}

 \usepackage[main, final]{neurips_2026}

\usepackage[utf8]{inputenc} 
\usepackage[T1]{fontenc}    
\usepackage{hyperref}       
\usepackage{url}            
\usepackage{booktabs}       
\usepackage{amsfonts}       
\usepackage{nicefrac}       
\usepackage{microtype}      
\usepackage{xcolor}         
\usepackage{amsmath}
\usepackage{amsthm}
\usepackage{caption}
\usepackage{subcaption}
\usepackage{graphicx} 
\usepackage{booktabs}
\usepackage{tabularx}
\usepackage{caption}
\usepackage{multirow} 

\theoremstyle{plain}
\newtheorem{theorem}{Theorem}[section]

\newtheorem{lemma}[theorem]{Lemma}
\newtheorem{corollary}[theorem]{Corollary}
\theoremstyle{definition}

\theoremstyle{remark}
\RequirePackage{algorithmic}
\RequirePackage{algorithm}

\newcommand{\Adv}{{\mathbf{\bf Adv}}}

\newtheorem*{theorem31}{\textbf{Theorem 3.1}}
\newtheorem*{lemma32}{\textbf{Lemma 3.2}}
\newtheorem*{theorem33}{\textbf{Theorem 3.3}
}
\title{Leveraging Soft Prompts for Privacy Attacks in Federated Prompt Tuning}

\author{%
  Quan Minh Nguyen \\
  University of Florida\\
  \And
  Min-Seon Kim \\
  North Carolina State University\\
  \And
  Hoang M.~Ngo \\
  University of Florida\\
  \AND
  Trong Nghia Hoang \\
  Washington State University\\
  \And
  Hyuk-Yoon Kwon \\
  Seoul National University\\
  of Science and Technology\\
  \And
  My T.~Thai\thanks{Corresponding author.} \\
  University of Florida\\
  \texttt{mythai@cise.ufl.edu}\\
}

\begin{document}

\maketitle

\begin{abstract}
  Membership inference attacks (MIAs) pose a serious privacy threat in federated learning (FL).~While MIAs have been extensively studied in standard FL, the recent shift toward federated fine-tuning introduces new and largely unexplored attack surfaces.~In this work, we show that federated prompt-tuning, which adapts pre-trained foundation models using lightweight input prefixes, exposes a novel and effective vector for membership inference.~We propose \textsc{PromptMIA}, a membership inference attack tailored to federated prompt-tuning, in which a malicious server introduces adversarially crafted prompts and exploits their updates during collaborative training to determine whether a target data point belongs to a client’s private dataset.~We formalize this threat via a security game and demonstrate that \textsc{PromptMIA} achieves consistently high attack advantage across diverse benchmark datasets, substantially outperforming current SOTA federated MIAs.~We also provide a theoretical lower bound on the attack advantage that explains the observed empirical behavior.~Finally, we show that existing MIA defenses are often ineffective against \textsc{PromptMIA}, highlighting the need for defense mechanisms specifically tailored to prompt-tuning in federated settings. 
\end{abstract}

\section{Introduction} 
\label{sec:intro}

Federated Learning (FL)~\cite{mcmahan2017communication} is a prominent training paradigm that enables data holders to collaboratively train a shared model without exposing their private data.~Despite this, FL methods are vulnerable to adversarial threats \citep{munoz2017towards, zhu2019deep}.~One prominent threat is membership inference attack (MIA) ~\citep{shokri2017membership}, where the adversarial server attempts to determine whether a specific data record was included in a client’s training dataset.~Existing research has predominantly studied such privacy threats in settings where the attack surface is tied to clients' gradients or full model parameters exchanged during training~\citep{melis2019exploiting,li2023effective}.~This classical view has, however, overlooked new attack surfaces that arise in more recent FL paradigms that instead communicate and aggregate external adaptation modules of pre-trained foundation models, e.g. soft-prompts prepended to input embeddings.~\citep{weng2024probabilistic}.

To raise awareness of the potential existence of unexplored privacy threats in recent FL paradigms, this paper reveals a novel attack surface via manipulating the clustering and aggregation of local sets of soft prompts in federated prompt-tuning (FPT)~\citep{weng2024probabilistic}.~To the best of our knowledge, this is the first work that identifies an unexplored privacy risk in soft prompt optimization for representation fine-tuning, particularly in FPT.~This work is orthogonal to and complement existing research on privacy risks in prompting large language models with exemplars and/or task instructions that contain private data for in-context learning (ICL). Such studies focus largely on designing adversarial queries that expose whether a particular data point was included in private instruction prompts used for ICL~\citep{wen2024membership, duan2023flocks}.~In contrast, soft prompt optimization does not include private data in the prompts themselves.~Moreover, the attack model in in-context learning originates from end users external to the training system whereas in FPT, it arises from the server within the training network.~Existing defenses against prompt risk in in-context learning~\citep{wu2023privacy, hong2023dp, tang2023privacy} are therefore not applicable to soft prompt risk in FPT.

To highlight the privacy risk that arises within soft prompt aggregation in FPT, we develop~\textsc{PromptMIA}, a novel membership inference attack (MIA) that exploits FPT’s prompt update and selection mechanism as a new attack surface.~This is substantiated with the following contributions: 

{\bf (I)}~We develop an algorithm that optimizes adversarial prompts to be preferentially selected and updated by local clients when their datasets contain a target data point, while remaining unselected otherwise.~\textsc{PromptMIA} is substantiated via running this algorithm on the server to create and add adversarial prompts to the shared prompt pools.~The server can keep track of changes made to these prompts between communication iterations to infer membership information in a single communication round, without the need for training a shadow model or external data (Section~\ref{sec:promptmia}).

{\bf (II)}~We analyze the theoretical attack advantage of~\textsc{PromptMIA} through the lens of a security game.~In this view, the advantage of the attacker is characterized in terms of the true and false positive rates which measure the chance that the adversarial prompts are correctly and incorrectly selected, respectively.~Our analysis shows that ~\textsc{PromptMIA} achieves perfect true positive rate and provably low false positive rate, resulting in a high attack advantage (Section~\ref{sec:theory}). 

{\bf (III)}~We run extensive experiments to validate the effectiveness of \textsc{PromptMIA} across $7$ datasets including CIFAR-10, CIFAR-100 \citep{krizhevsky2009learning}, TinyImageNet \citep{le2015tiny}, MNIST-M\citep{ganin2016domain}, Fashion-MNIST\citep{xiao2017fashion}, CINIC-10 \citep{darlow2018cinic}, MMAFEDB\citep{mahmoudi2020mma} and $3$ vision transformer architectures
including ViT-B/32~\citep{dosovitskiy2020image}, DeiT-B/16~\citep{touvron2021training}, and ConViT~\citep{d2021convit}.~Our empirical results consistently show that \textsc{PromptMIA} outperforms SOTA federated MIAs \citep{zhu2025fedmia, gu2022cs,yeom2018privacy}, achieving $> 90 \%$ attack success rate on all datasets, which corroborates the aforementioned high attack advantage (Section~\ref{subsec:advantage_attack_success_rate_measurement}). Our code is available at \url{https://github.com/GivralNguyen/Prompt-MIA}.

{\bf (IV)}~While the primary objective of this work is not to propose novel defense mechanisms or conduct an exhaustive evaluation of all possible defense strategies against \textsc{PromptMIA}, we conduct additional analyses and experiments to examine how \textsc{PromptMIA} interacts with FL clients implementing standard defenses such as outlier detection \citep{liu2008isolation, breunig2000lof, estimator1999fast, manevitz2001one}, input noise perturbation \citep{lecuyer2019certified}, regularization-based defenses \citep{wangni2018gradient, huang2020instahide, shorten2019survey} (Section~\ref{sec:MIA_DEF}), as well as defenses that modify the prompt selection or aggregation protocol, such as Secure Aggregation \citep{bonawitz2017practical}, randomized key indices, and randomized prompt selection via dropout, and monitoring protocols such as cross-round key-drift auditing
(Appendix~\ref{appx:systemdefenses}). We show that these defenses are often either provably or empirically ineffective. We do not claim that no future audit or secure protocol can defeat the attack. Rather, these results underscore the urgent need for more research on privacy defenses against this new prompt-based attack surface.
\section{Federated Prompt-Tuning}
\label{sec:FPT}

To address the resource bottleneck in traditional full model fine-tuning, recent literature in federated learning (FL) has increasingly adopted parameter-efficient tuning methods, notably prompt tuning. Prompt-tuning~\citep{lester2021power, wang2022learning} reformulates the downstream task adaptation problem as input modification rather than weight updates. This method focuses on optimizing learnable soft-prompts which are appended to the input embedding in a transformer architecture. These prompts provide beneficial context for downstream tasks, similar to how hints can be provided to assist puzzle solving. In Federated Prompt-Tuning (FPT), a central server maintains a global prompt pool that clients collaboratively optimize, while keeping the underlying large-scale foundation model frozen. Let the global prompt pool of size $M$ be denoted as  $\mathcal P_{\textsc{global}}=\{P_i\}_{i=1}^M$ with a corresponding set of learnable keys $\mathcal K_{\textsc{global}}=\{k_i\}_{i=1}^M$. During training or inference, given an input $x$, the frozen pre-trained model $f$ extracts the \texttt{[CLS]} representation to form a deterministic query $q(x) = f(x)[0,:]$. Using cosine distance $\gamma$, the client retrieves the top-$N$ contextually relevant prompts from $\mathcal{P}_{\textsc{global}}$ (Eq. \ref{eq:cosinedistance}):
\begin{equation}
\label{eq:cosinedistance}
\hat{\mathcal K}_x=\operatorname*{argmin}_{\{s_i\}_{i=1}^{N}\subseteq [M]}
\sum_{i=1}^{N}\gamma\!\big(q(x),k_{s_i}\big) =\operatorname*{argmax}_{\{s_i\}_{i=1}^{N}\subseteq [M]}
\sum_{i=1}^{N}\kappa\!\big(q(x),k_{s_i}\big), 
\qquad [M]=\{1,\dots,M\}.   
\end{equation}

i.e., the $N$ prompts whose associated keys are closest to the query feature $q(x)$ under the cosine distance $\gamma$, or equivaliently, the $N$ prompts whose associated keys have the highest cosine similarity to $q(x)$. These selected prompts are prepended to the input embedding $x_e$, forming the adapted input $x_p=[P_{s_1};\cdots;P_{s_N};x_e]$. Let $h_{\mathrm{CLS}}(x_p)$ denote the final \texttt{[CLS]}
representation of the prompt-augmented input.
Let $g$ denote the optional frozen pretrained classification head.
With the trainable classifier $f_c^{\phi}$ parameterized by
$\phi$, the training objective is defined as:
\begin{equation}
\label{eq:trainl2p}
\min_{\mathcal P,\mathcal K,\phi}\;
\mathcal L\!\left(
    f_c^{\phi}\!\left(g\!\left(h_{\mathrm{CLS}}(x_p)\right)\right),
    y
\right)
+\lambda\!\sum_{i\in\hat{\mathcal K}_x}
\gamma\!\big(q(x),k_i\big).
\end{equation}

\begin{figure*}[!t]
    \centering
    \includegraphics[width=0.68\textwidth]{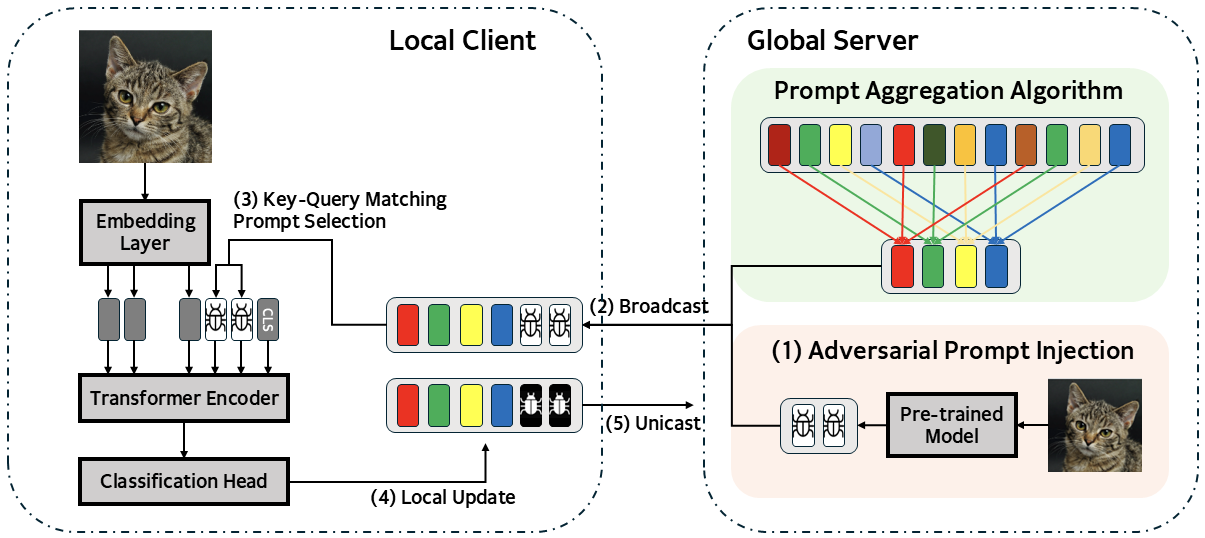}
    \caption{
    \textsc{PromptMIA} workflow: (1) the server injects adversarial prompts designed for a target sample into the global prompt pool; (2) the modified pool is broadcast to clients; (3) each client performs query-key matching, which selects all adversarial prompts if the target sample is present; (4) selected prompts are locally updated and (5) returned to the server. By monitoring which prompts are updated, the server infers the target’s membership in client data.
    }
    \label{fig:mia}
\end{figure*}

The first term is the softmax cross-entropy, while the second is a surrogate that pulls the selected keys to align with the corresponding query features.  Given input image $x$ and global key-prompt pool $(\mathcal{K}_{\textsc{global}}, \mathcal{P}_{\textsc{global}})$, the client selects $\hat{\mathcal K}_{x}$ using Eq.~\ref{eq:cosinedistance} and updates $\hat{\mathcal K}_{x}$, the corresponding prompts $\hat{\mathcal P}_{x}$ along with ${f_{c}}^{\phi}$ according to Eq.~\ref{eq:trainl2p}. The server aggregates the trained prompts from participating clients to update $(\mathcal K_{\textsc{global}},\mathcal P_{\textsc{global}})$ using any of the existing FL algorithms (e.g., \textsc{FedAvg}~\citep{mcmahan2017communication}, \textsc{FedProx}~\citep{li2020federated} or \textsc{PFPT}~\citep{weng2024probabilistic}). See Appendix \ref{appx:pfpt} for more details.

\section{MIAs against Federated Prompt-Tuning}\label{sec:mia}
\label{headings}

In this section, we present the formulation and workflow of \textsc{PromptMIA}. An overview of the attack is shown in Fig.~\ref{fig:mia}. Unlike prior approaches that rely on model weights, gradients \citep{zhu2025fedmia, gu2022cs}, or auxiliary datasets \citep{yeom2018privacy} to train shadow models, \textsc{PromptMIA} operates by injecting adversarial prompts $\mathcal{P_\textsc{adv}}$ into the shared prompt pool $\mathcal{P_\textsc{global}}$ prior to a training round. These prompts are designed to be selected only when target sample $\mathcal{T}$ is present in a client's data. As clients select and update prompts (Eq. \ref{eq:cosinedistance}), the server can monitor which injected prompts are modified and use this as membership signal. \textsc{PromptMIA} exploits prompt updates as a privacy channel to infer membership in a single round, reducing both computational cost and adversarial assumptions. ~\textsc{PromptMIA} achieves perfect true positive rate and provably low false positive rate, resulting in a high attack success rate.

\subsection{Prompt based AMIs as Security Games}\label{sec:securitygame}
We formalize the prompt-based active membership inference attack (AMI) threat models through security games between a challenger (client) and an adversary (server) in the FPT setting.

{\bf (I)}~ The adversary is denoted by \( \mathcal{A} \), and the corresponding security games are represented as \( \mathsf{Exp}^{\textup{AMI}}(\mathcal{A}) \). In \( \mathsf{Exp}^{\textup{AMI}}(\mathcal{A}) \), the adversarial server $\mathcal{A}$ consists of three components: $\mathcal{A}_{\mathsf{INIT}}$, $\mathcal{A}_{\mathsf{ATTACK}}$, and $\mathcal{A}_{\mathsf{GUESS}}$. At the beginning of the games, a random bit $b$ determines whether a target sample $\mathcal{T}$ belongs to the client's data $\mathcal{D}$ (with $b$ = 1 indicating membership). 

{\bf (II)}~In the initialization phase $\mathcal{A}_{\mathsf{INIT}}$, the server constructs the current round's aggregated global key-prompt pool $(\mathcal{K}_{\textsc{global}}, \mathcal{P}_{\textsc{global}}) = \{\, (k_{i}, P_{i}) \,\}_{i=1}^M$ using any existing FL aggregation algorithm. 

{\bf (III)}~In the attack phase $\mathcal{A}_{\mathsf{ATTACK}}$, the server constructs a set of $N$ adversarial keys $\mathcal{K}_{\textsc{adv}} = \{\, k_{a_m} \,\}_{m=1}^N ,$ where each $k_{a_m}$ is an adversarial key vector explicitly generated for target sample $\mathcal{T}$ given query function $q$. It then selects a subset of $N$ prompts $\{(k_j, P_j)\}_{j \in S} \subseteq (\mathcal{K}_{\textsc{global}}, \mathcal{P}_{\textsc{global}})$ where $S \subseteq \{1, \dots, M\}$ and $|S| = N$. For each index $j \in S$, the server replaces the original key $k_j$ with an adversarial key from $\mathcal{K}_{\textsc{adv}}$. While only the key values are changed, for ease of notation, we define the set of adversarial prompts as $(\mathcal{K}_{\textsc{adv}}, \mathcal{P}_{\textsc{adv}}) = \{\, (k_{a_m}, P_{a_m}) \,\}_{m=1}^N .$ Correspondingly, the set of remaining prompts are benign prompts $(\mathcal{K}_{\textsc{benign}}, \mathcal{P}_{\textsc{benign}}) = \{\, (k_{b_n}, P_{b_n}) \,\}_{n=1}^{M-N} .$ Together, we obtain the modified global prompt pool $(\tilde{\mathcal{K}}, \tilde{\mathcal{P}}) = \bigl( \mathcal{K}_{\textsc{adv}} \cup \mathcal{K}_{\textsc{benign}},\; \mathcal{P}_{\textsc{adv}} \cup \mathcal{P}_{\textsc{benign}} \bigr).$ 

$\mathcal{A}_{\mathsf{ATTACK}}$ then distributes $(\tilde{\mathcal{K}}, \tilde{\mathcal{P}})$ to participating clients (instead of $(\mathcal{K}, \mathcal{P})$ like in regular FPT). Each client $t$ selects $(\hat{\mathcal{K}}_{t},\hat{\mathcal{P}}_{t})$ from $(\tilde{\mathcal{K}}, \tilde{\mathcal{P}})$ using Eq.~\ref{eq:cosinedistance} and updates them based on their local data $\mathcal{D}$. 

{\bf (IV)}~In $\mathcal{A}_{\mathsf{GUESS}}$, the server uses $\hat{\mathcal{P}_t}$ to guess $b$, effectively identifying whether $\mathcal{T} \in \mathcal{D}$. The advantage of the adversarial server $\mathcal{A}^\mathcal{D}$ in the security game is given by:
\begin{align}\label{eq:advantage_def}
      \Adv^{\textup{AMI}}(\mathcal{A}) 
      = \Pr[b'=1|b=1] - \Pr[b'=1|b=0]
\end{align} 
where $b$ is the true membership label, and $b'$ is the adversary’s prediction. We define the \emph{attack success rate} as:
\begin{equation}\label{eq:asr_def}
\textbf{ASR}^{\textup{AMI}}(\mathcal{A}) 
= \frac{1}{2} (\Pr[b'=1 \mid b=1]+\Pr[b'=0 \mid b=0])
= \tfrac{1}{2}\bigl(1+\Adv^{\textup{AMI}}(\mathcal{A})\bigr).
\end{equation}

\subsection{MIA using Adversarial Prompt Injection}
Given the security game laid out in Section~\ref{sec:securitygame}, the adversary’s objective is to inject a set of $N$ adversarial keys-prompts $(\mathcal{K}_{\textsc{adv}} ,\mathcal{P}_{\textsc{adv}})$ into the global prompt pool 
such that, if $\mathcal{T} \in \mathcal{D}$, the top-$N$ selected prompts will always be the adversarial set.
The membership signal is defined as the event that \textbf{all adversarial prompts are selected and subsequently updated}. To build intuition, we first introduce a \emph{Naive Prompt Injection} attack and analyze its weaknesses, before introducing the more robust method, \textsc{PromptMIA}, which achieves perfect true positive rate and provably low false positive rate, resulting in a high attack advantage ( Section \ref{sec:theory}).
\subsubsection{Naive Prompt Injection Attack}
\begin{figure*}[!t]
  \centering
  \begin{subfigure}{0.32\textwidth}
    \includegraphics[width=\linewidth]{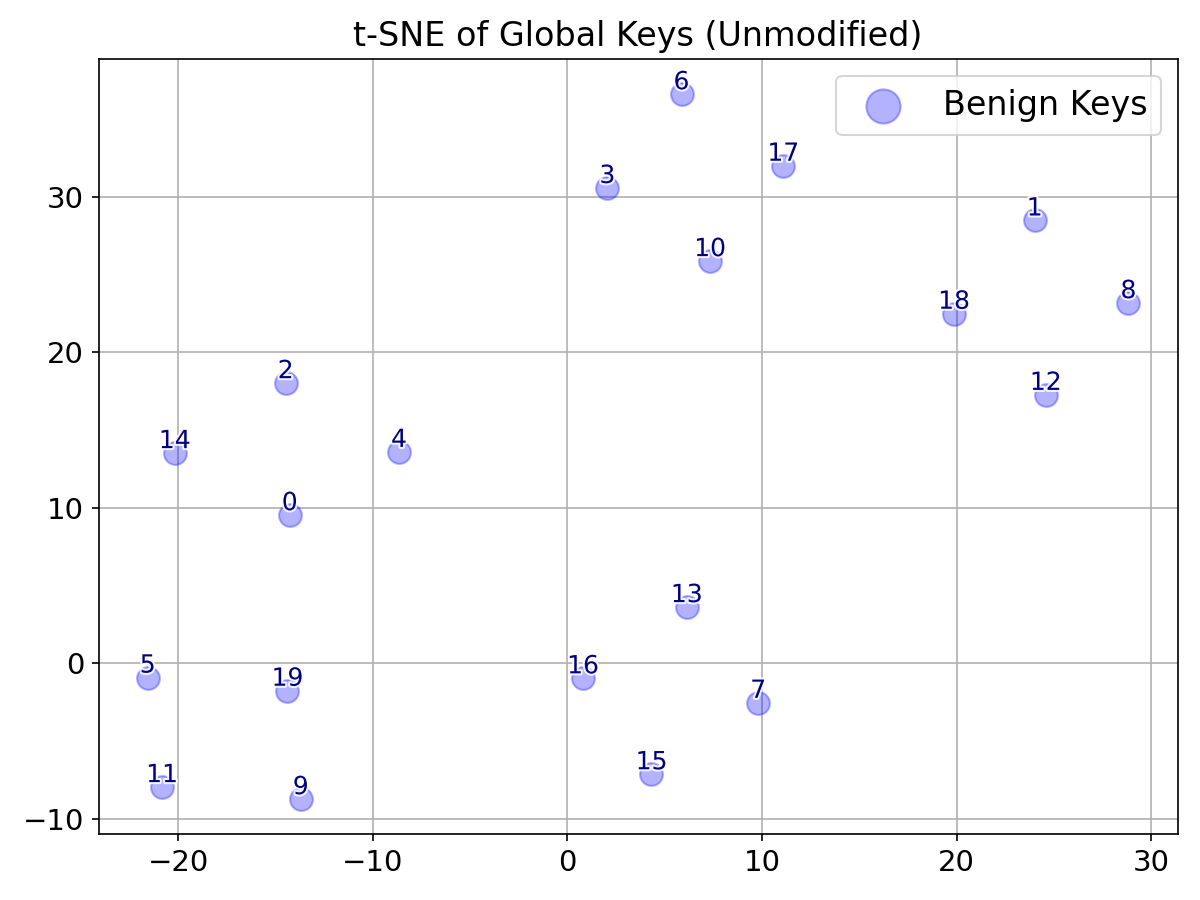}
    \caption{t-SNE of global keys without adversarial key injection. \newline}
    \label{fig:tsebenign}
  \end{subfigure}\hfill
  \begin{subfigure}{0.32\textwidth}
    \includegraphics[width=\linewidth]{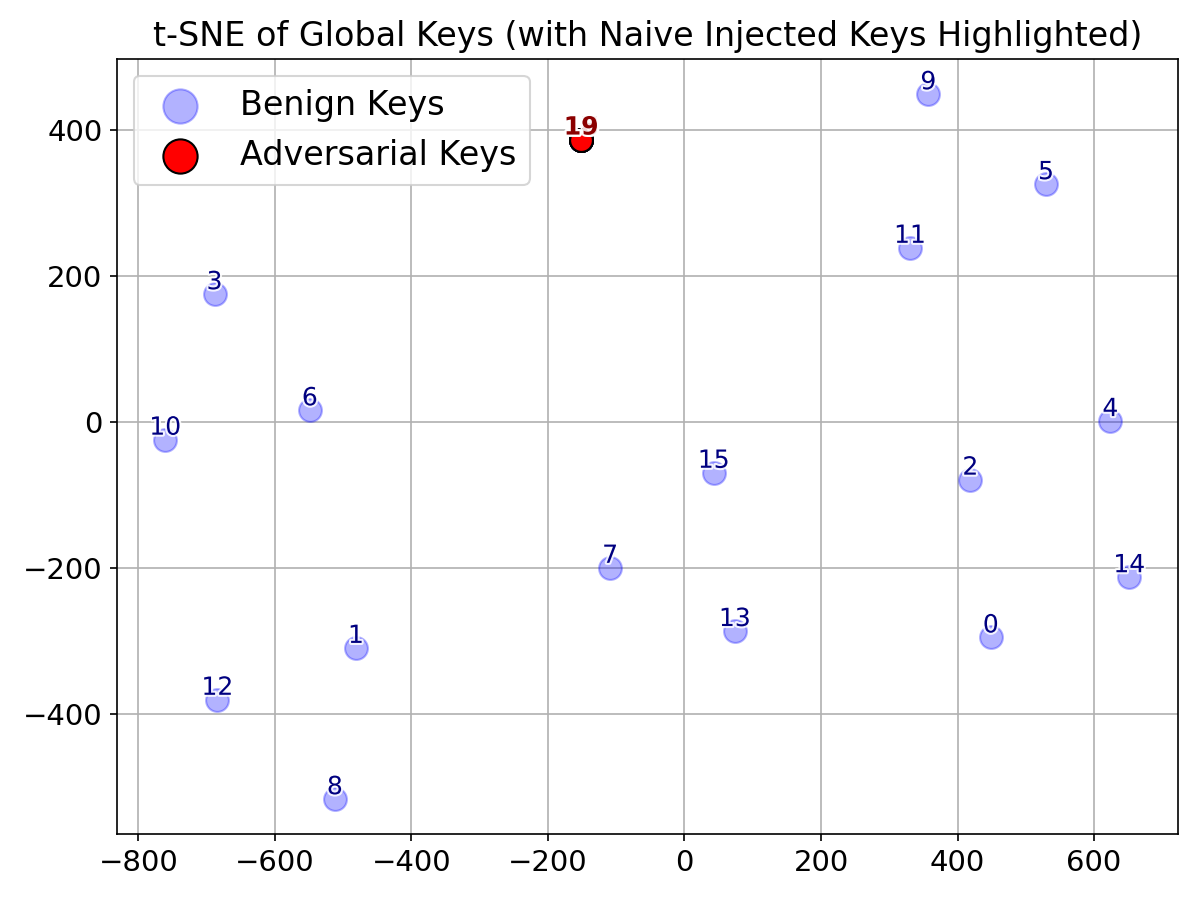}
    \caption{t-SNE visualization shows naively injected adversarial keys overlapping}
    \label{fig:tsenaive}
  \end{subfigure}\hfill
  \begin{subfigure}{0.32\textwidth}
    \includegraphics[width=\linewidth]{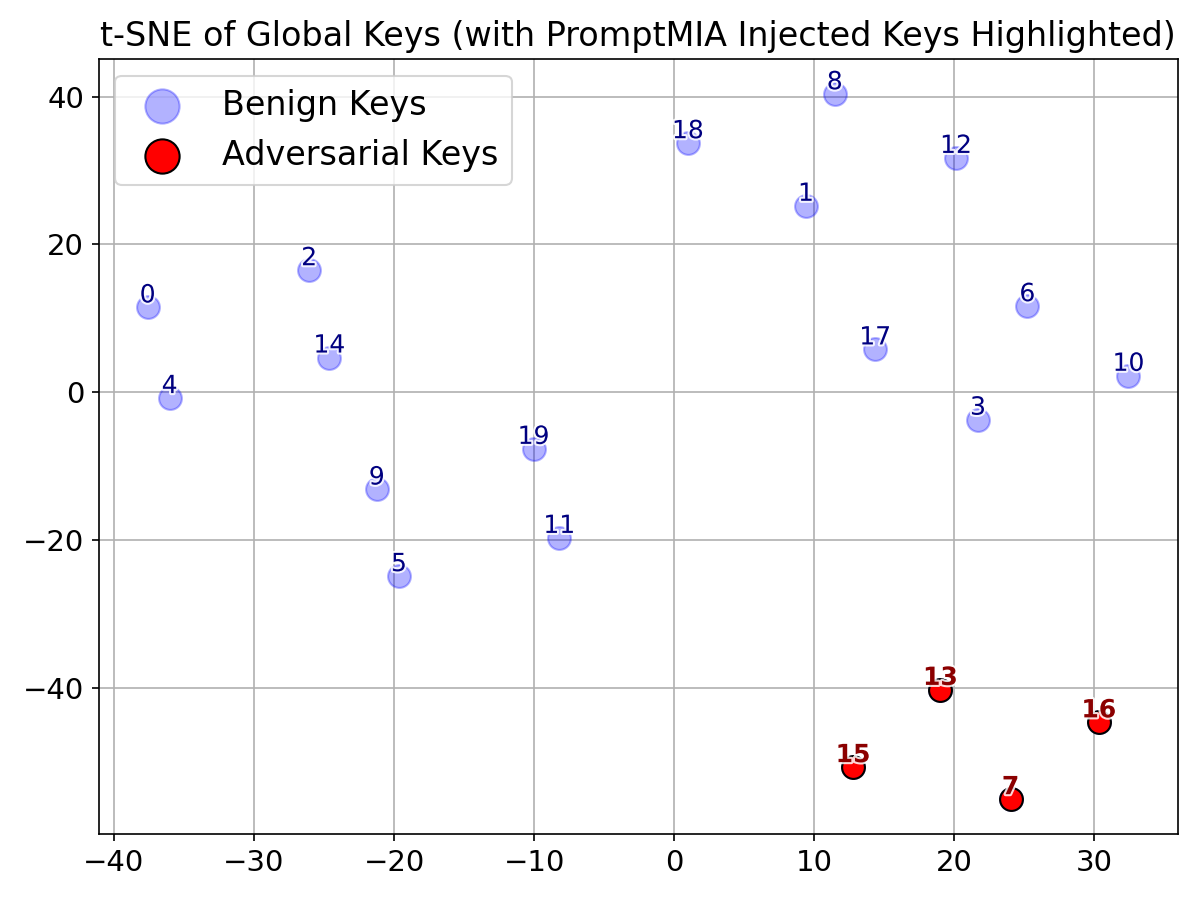}
    \caption{t-SNE of global keys after PromptMIA adversarial keys injection.}
    \label{fig:tsepromptmia}
  \end{subfigure}
  \caption{Comparison of the global key distributions produced by a ViT-B/32 model trained on CIFAR-10 after 60 global epochs, visualized using t-SNE. Blue keys are benign keys, and red keys are adversarial keys.}
  \label{fig:tsneall}
\end{figure*}

In this approach, the adversary constructs adversarial keys $\mathcal{K}_{\textsc{adv}} = \{\, k_{a_m} \,\}_{m=1}^N$  and inserts them into the global prompt pool. The goal is to ensure that if target data $\mathcal{T} \in \mathcal{D}$, the top-$N$ selected prompts using Eq. \ref{eq:cosinedistance} always coincide with $\mathcal{K}_{\textsc{adv}}$. A straightforward strategy is to align each adversarial key $k_{a_m}$ directly with the client’s query vector $q(\mathcal{T})$ so as to maximize cosine similarity. We ensure that:\\
\begin{equation}
\label{eq:naive}
\kappa\!\left(q(\mathcal{T}), k_{a_m}\right) 
= 1 \;\text{ or }\; k_{a_m} = q(\mathcal{T}), 
\quad \forall\, k_{a_m} \in \mathcal{K}_{\textsc{adv}} .
\end{equation}
where $\kappa$ is the cosine similarity operator. The server selects a subset $S \subseteq [M]$ with $|S|=N$, 
and for each $j \in S$ performs key modificaton:
\begin{equation}
\label{eq:keymod}
 (k_j, P_j) \;\mapsto\; (k_{a_m}, P_j), \qquad m=1,\dots,N,   
\end{equation}
However, this naive attack suffers from fundamental weaknesses. Since all adversarial keys collapse to the same vector $q(\mathcal{T})$ due to Eq. \ref{eq:naive}, clients can easily detect this redundancy through dimensionality reduction techniques (e.g., t-SNE, PCA) (see Fig. \ref{fig:tsenaive}), which would reveal a tight cluster of identical keys, or simply by directly inspecting key values. In addition, simple defenses such as discarding any key whose similarity exceeds a suspicious threshold 
(e.g., excluding $\kappa(q(\mathcal{T}), k_{a_m}) > 1-r$ for small $r>0$) can render the attack ineffective. 

Finally, because all adversarial keys are identical, the attack suffers from a high false positive rate ( see Section \ref{subsec:advantage_attack_success_rate_measurement}) : even when the target sample is absent, adversarial prompts may still be selected whenever the similarity of one adversarial key ( and as a result, all adversarial keys) to some non-target query $q(x)$, $x \in \mathcal{D},\, x \neq \mathcal{T}$, is higher than all benign keys, or $\kappa\!\left(q(x), k_{a_m}\right) > \max_{k_b \in \mathcal{K}_{\textsc{benign}}} \kappa\!\left(q(x), k_b\right)$, significantly increasing FPR and reducing the attack's advantage.

\subsubsection{PromptMIA}
\label{sec:promptmia}
To build up on and overcome the limitations of the naive prompt injection attack, we impose two requirements on adversarial key generation. First, adversarial keys $\mathcal{K}_{\textsc{adv}}$ must achieve higher cosine similarity with the target query $q(\mathcal{T})$ 
than any benign key $\mathcal{K}_{\textsc{benign}} $. Second, adversarial keys must be sufficiently diverse from one another to not be trivially exposed. Formally, the server selects a subset $S \subseteq [M]$ of size $N$, and for each index $j \in S$, the original key $k_j$ is replaced with an adversarial key $k_{a_m} \in \mathcal{K}_{\textsc{adv}}$ ( see Eq. \ref{eq:keymod}). $k_{a_m}$ is constrained to lie within a cosine similarity interval with $q(\mathcal{T})$:
\begin{equation}
\label{eq:adv_constraint}
\kappa\!\left(q(\mathcal{T}), k_{a_m}\right) \in 
\Bigl[\max_{k_b \in \mathcal{K}_{\textsc{benign}}} \kappa\!\left(q(\mathcal{T}), k_b\right) + \delta_{\min}, \;\;
\max_{k_b \in \mathcal{K}_{\textsc{benign}}} \kappa\!\left(q(\mathcal{T}), k_b\right) + \delta_{\min} + \Delta \Bigr], \forall\, k_{a_m} \in \mathcal{K}_{\textsc{adv}}
\end{equation}

where $\delta_{\min}$ ensures that all adversarial keys have higher cosine similarity to $q(\mathcal{T})$ than benign keys, while $\Delta$ introduces controlled variability to prevent all adversarial keys from collapsing to the same vector. Algorithm~\ref{alg:gen_key_with_similarity} describes the procedure for generating a single adversarial key with a specified cosine similarity $s$ to the target query. Building on this primitive, Algorithm~\ref{alg:gen_adv_key_set} constructs a set of $N$ adversarial keys $\mathcal{K}_{\textsc{ADV}}$ that satisfies Eq. \ref{eq:adv_constraint}. Details of these algorithms are given in Appendix \ref{appx:genkey}. \textsc{PromptMIA} allows the adversary to generate an adversarial key set $\mathcal{K}_{\textsc{adv}}$ that will always be selected and updated when $\mathcal{T} \in \mathcal{D}$, while not being easily detectable (see Fig. \ref{fig:tsepromptmia} and Section \ref{sec:MIA_DEF}).
The server generates and distributes the modified global prompt pool $(\tilde{\mathcal{K}}, \tilde{\mathcal{P}}) 
= \bigl( \mathcal{K}_{\textsc{adv}} \cup \mathcal{K}_{\textsc{benign}},\; 
           \mathcal{P}_{\textsc{adv}} \cup \mathcal{P}_{\textsc{benign}} \bigr)$ to all participating clients. Given client $t$ with local dataset $\mathcal{D}$ and input data $x \in \mathcal{D}$, the client computes $q(x)$ and selects top-$N$ keys $\hat{\mathcal K}_x$ using Eq. \ref{eq:cosinedistance} and update only the selected prompts using Eq. \ref{eq:trainl2p}. If $x \equiv \mathcal{T}$, then we have the top-$N$ keys are exactly the adversarial keys $\mathcal{K}_{\textsc{adv}}$, or $\hat{\mathcal K}_x \equiv \mathcal{K}_{\textsc{adv}}$ , which corresponds precisely to the membership condition.

\subsection{Theoretical Analysis of PromptMIA}\label{sec:theory}

In this section, we provide theoretical guarantees that \textsc{PromptMIA} achieves a perfect true positive rate (TPR) alongside a provably low false positive rate (FPR). Specifically, TPR is the adversary's success in identifying the target sample $\mathcal{T}$ when it is a member ($b=1$), and FPR is the adversary's error in identifying $\mathcal{T}$ as a member when it is a non-member ($b=0$).

{\bf (I)}~First, the \textsc{PromptMIA} attack is constructed to ensure perfect identification when the target sample is present. Theorem~\ref{thm:tpr} establishes that the TPR of \textsc{PromptMIA} equals 1. The proof follows directly from the construction of adversarial keys in Algorithm~\ref{alg:gen_adv_key_set}, which ensures that $\min_{k_a \in \mathcal{K}_{\textsc{adv}}} \kappa(q(\mathcal{T}), k_a) > \max_{k_b \in \mathcal{K}_{\textsc{benign}}} \kappa(q(\mathcal{T}), k_b)$.
Thus, if the client possesses $\mathcal{T} \in \mathcal{D}$, the top-$N$ selected keys must be exactly the set $\mathcal{K}_{\textsc{adv}}$.

\begin{theorem} [True Positive Rate]
\label{thm:tpr}
Let $\mathcal{K}_{\textsc{adv}} = \{k_{a_m}\}_{m=1}^N$ be the set of $N$ adversarial keys generated by Algorithm \ref{alg:gen_adv_key_set} with parameters $\delta_{\min} > 0$ and $\Delta \ge 0$. Let $\mathcal{K}_{\textsc{benign}}$ be the set of $M-N$ benign keys. If the client's dataset $\mathcal{D}$ contains the target sample $\mathcal{T}$ (i.e., $b=1$) and the client's selection mechanism (Eq. \ref{eq:cosinedistance}) selects the top-$N$ prompts based on highest cosine similarity (lowest cosine distance), the set of selected keys $\hat{\mathcal{K}}_\mathcal{T}$ for the query $q(\mathcal{T})$ will be exactly the adversarial set:
$\hat{\mathcal{K}}_\mathcal{T} = \mathcal{K}_{\textsc{adv}}$. Consequently, the True Positive Rate (TPR) of \textsc{PromptMIA} is 1, or $\text{TPR} = \Pr[b'=1 \mid b=1] = 1$
\end{theorem}

{\bf (II)}~We now analyze the FPR, $\Pr[b'=1 \mid b=0]$, which is more complex. A false positive arises whenever a non-member query $q(x)$ selects all $N$ adversarial keys $\{k_{a_j}\}_{j=1}^{N}$ as its top-N choices than to any benign key  $\{k_{b_i}\}_{i=1}^{M-N}$. 
To mathematically derive a closed-form bound for this probability, it is necessary to formalize the geometric and statistical properties of the non-member query space. Because the exact distribution of high-dimensional query vectors is analytically intractable, we establish two foundational assumptions grounded in the inherent optimization dynamics of FPT.

\begin{figure*}[ht!]
  \centering
  \begin{minipage}{0.32\textwidth}
    \centering
    \includegraphics[width=\linewidth]{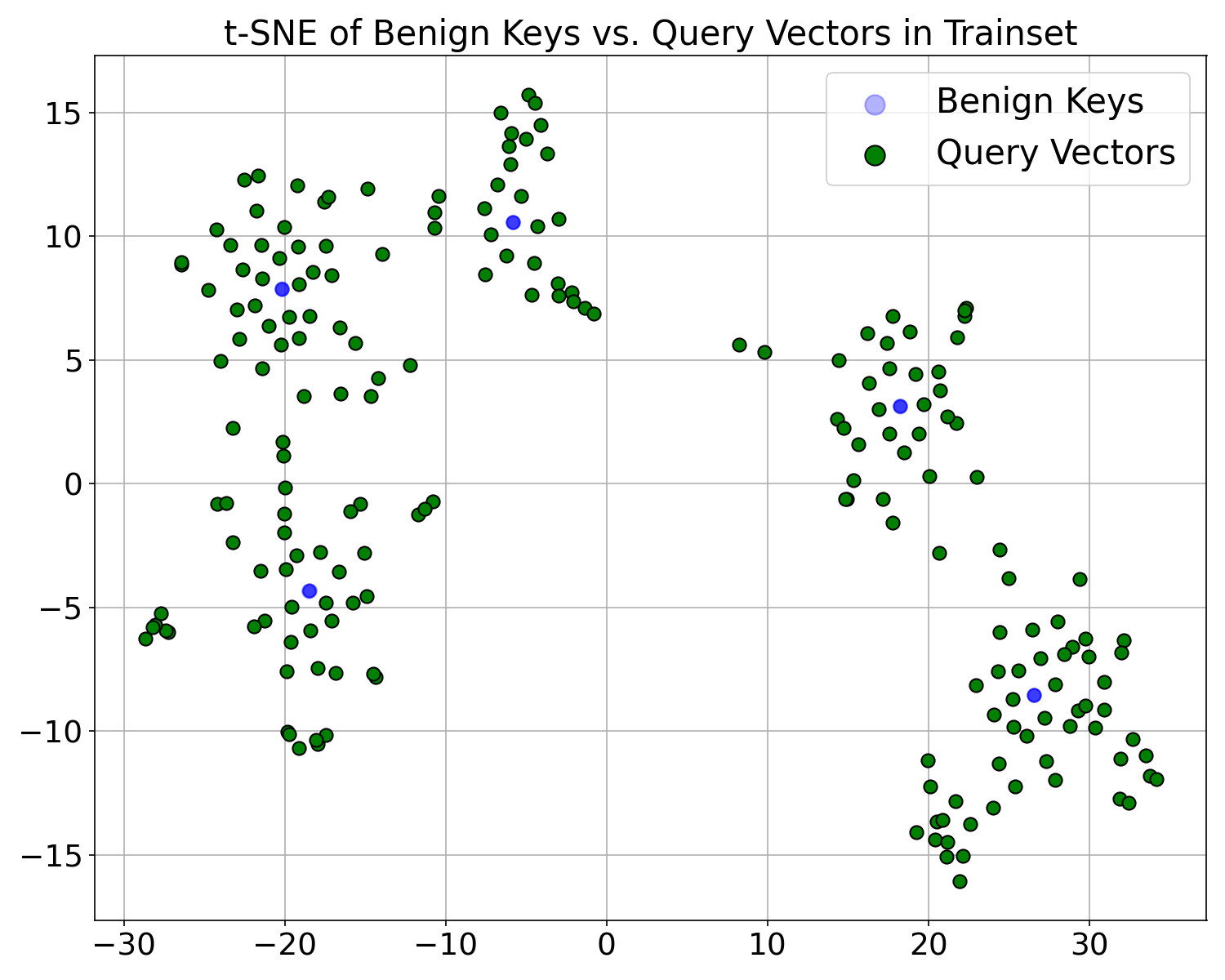}
  \end{minipage}
  \hfill
  \begin{minipage}{0.32\textwidth}
    \centering
    \includegraphics[width=\linewidth]{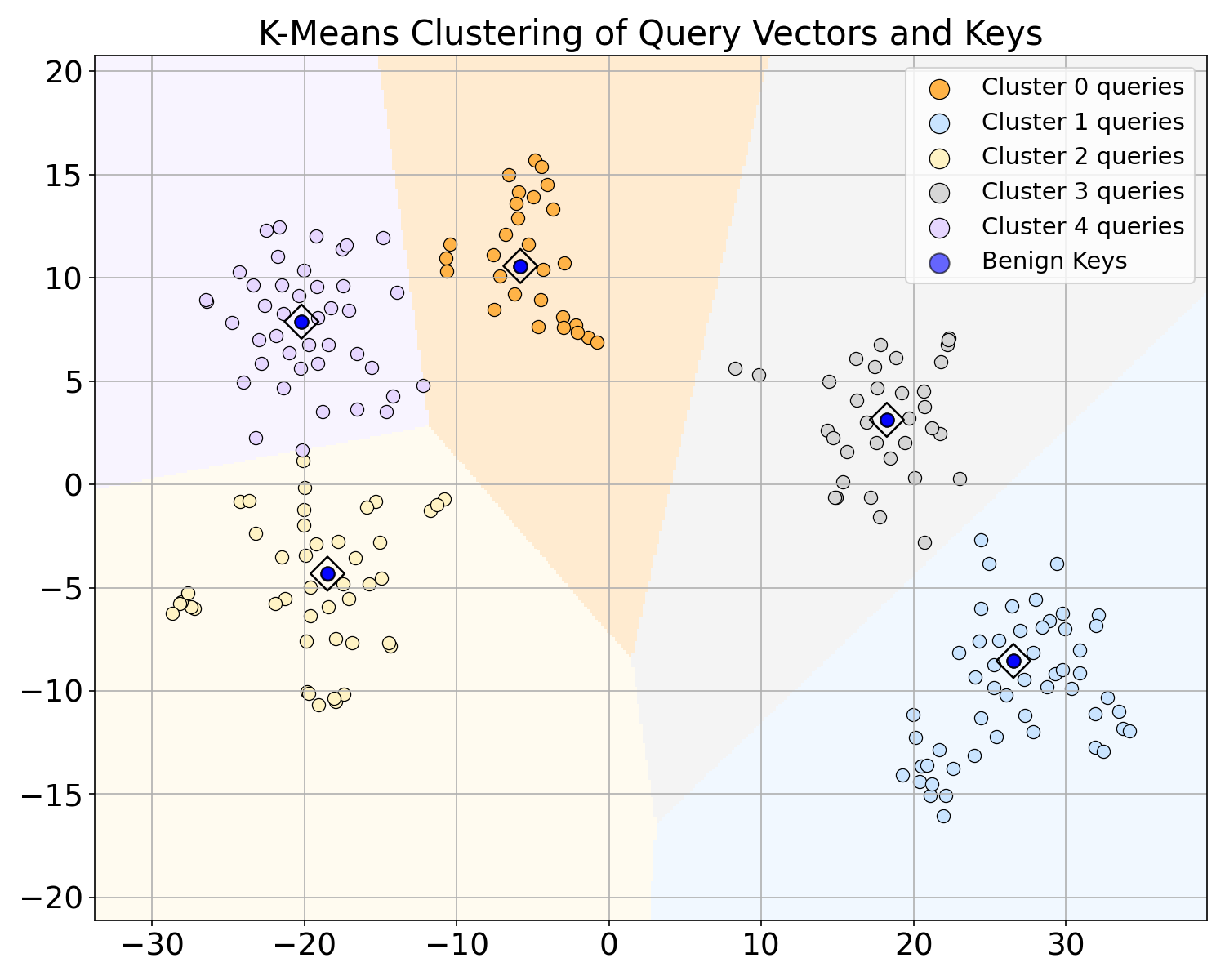}
  \end{minipage}
  \hfill
  \begin{minipage}{0.32\textwidth}
    \centering
    \includegraphics[width=\linewidth]{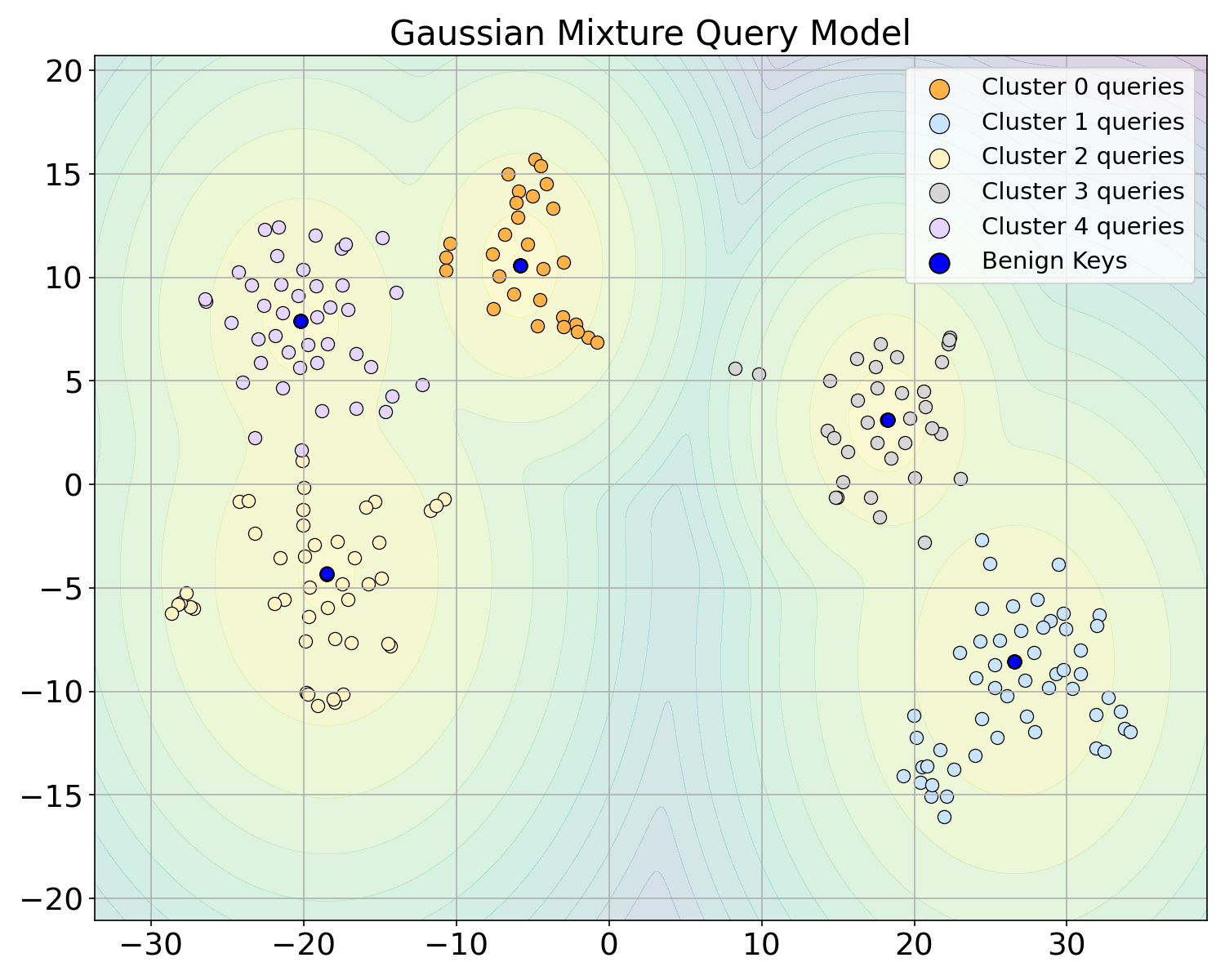}
  \end{minipage}
  \caption{\textbf{Left:} t-SNE projection of the global keys and query vectors from the train set. \textbf{Center:} K-Means clustering of keys with queries. \textbf{Right:} Gaussian Mixture Query Model. Visualizations are generated from a ViT-B32 model trained on CIFAR-10 for 60 global epochs.}
  \label{fig:key_query_overview}
\end{figure*}

\textbf{Assumption 1 (Benign Keys as Cluster Centroids).}
\textit{We assume that the $K = M-N$ benign keys $k_{b_i} \in \mathcal{K}_{\textsc{benign}}$ act as the effective centroids of the non-member query vector distribution. Each non-member query $q(x)$ is assumed to belong to the cluster of its nearest benign key.}

First, the objective (Eq.~\ref{eq:trainl2p}) includes a surrogate loss $\gamma(q(x), k_i)$ that explicitly pulls selected keys to align with their corresponding query features. Over multiple training rounds, this optimization forces the benign keys to stabilize and act as cluster centers, which formally justifies Assumption 1.  




\textbf{Assumption 2 (Gaussian Mixture Query Model).}
\textit{We model the global marginal distribution of non-member queries $q(x)$ as a Gaussian Mixture Model (GMM). Specifically, the conditional distribution of queries belonging to benign cluster $i$ is modeled as a spherical Gaussian centered at that key: $q(x) \mid (x \in \text{cluster } i) \sim \mathcal{N}(k_{b_i}, \sigma_i^2 I)$, where $\sigma_i^2$ is the local cluster variance.} 

Second, Assumption 2 is motivated by the local alignment structure induced by benign keys. Queries assigned to the same benign key tend to concentrate around that key in the embedding space, while different benign keys define distinct local regions. Thus, we approximate each local region by a spherical Gaussian centered at its benign key, forming a tractable Gaussian mixture model for analysis. This is consistent with the standard use of Gaussian mixtures for density estimation and clustering~\citep{li1999mixture,Lianmeng2022}. In addition, this approximation captures the empirical multi-cluster geometry visualized in Fig.~\ref{fig:key_query_overview}.



Next, we introduce the adversarial key into this setting. For a non-member query $q(x)$ drawn from cluster $i$, we define a \emph{cluster-flip} event $E_i$ as the case when $q(x)$ selects all $N$ adversarial keys as its $N$ nearest centroids. Formally, $E_i$ is the intersection of $N \times (M-N)$ \emph{race} events $A_{jl}$, where each $A_{jl}$ denotes that $q(x)$ is closer to an adversarial key $k_{a_j}$ than to a benign key $k_{b_l}$. A cluster-flip event thus arises only if the adversary wins all of these races simultaneously. The probability of the joint event $E_i$ can then be upper bounded by the probability of the single race with the lowest success probability, as formally stated in Lemma~\ref{lem:flip_prob_main}.

\begin{lemma}
\label{lem:flip_prob_main}
Let $q(x)$ be a non-member query drawn from the $i$-th component of the GMM, such that its conditional distribution is $q(x) \mid (x \in \text{cluster } i) \sim \mathcal{N}(k_{b_i}, \sigma_i^2 I)$. The conditional probability $\Pr(E_i)$, that $q(x)$ selects all $N$ adversarial keys $\mathcal{K}_{ADV} = \{k_{a_1}, \dots, k_{a_N}\}$ as its $N$ closest centroids, is bounded by:
$$\Pr(E_i) \le \min_{\substack{1 \le j \le N \\ 1 \le l \le M-N}} \Phi\left( \frac{(k_{a_j}-k_{b_l})^T k_{b_i}}{\sigma_i\|k_{a_j} - k_{b_l}\|} \right)$$
where $\Phi(\cdot)$ is the Cumulative Distribution Function (CDF) of the standard normal distribution.
\end{lemma}

With this lemma, we can bound the FPR. The FPR is the probability of the event $E_{FP}$ that a single, random non-member query $q(x)$ results in a cluster-flip event. Using the Law of Total Probability, FPR can be expressed as a weighted sum of cluster-flip probabilities, where the weights correspond to the prior probabilities of clusters. Based on the bound on the probability of cluster-flip events for each cluster established in Lemma~\ref{lem:flip_prob_main}, we can derive a bound on the FPR which is the largest (worst-case) cluster-flip probability across all clusters, stated in Theorem~\ref{thm:fpr_main}.

\begin{theorem} [False Positive Rate]
\label{thm:fpr_main}
The per-sample False Positive Rate (FPR) is bounded by:
$$FPR = \Pr[b'=1 \mid b=0] \le \max_{1\leq i\leq M-N} \left( \min_{\substack{1 \le j \le N \\ 1 \le l \le M-N}} \Phi\left( z_{ijl} \right) \right)$$
where $\mathcal{K}_{ADV} = \{k_{a_1}, \dots, k_{a_N}\}$ is the set of $N$ adversarial keys, and 
$z_{ijl} = \frac{(k_{a_j} - k_{b_l})^T k_{b_i}}{\sigma_i \|k_{a_j} - k_{b_l}\|}$
\end{theorem}

Theorem~\ref{thm:fpr_main} provides some insights on conditions under which the attack is most effective. First, the bound is tighter when the adversarial target $q(\mathcal{T})$ is highly distinctive. A distinctive target ensures that its corresponding adversarial keys are geometrically well separated from benign key clusters, resulting in small $\Phi\left( z_{ijl} \right)$. In addition, the FPR is lower when the benign data forms tight and compact clusters in the query space around the benign keys (i.e., minimizing $\sigma_i^2$). That will decreases the likelihood that any non-member query will randomly stray into the adversary's region. Additionally, combining Theorems~\ref{thm:tpr} and~\ref{thm:fpr_main} yields the \textbf{bound on the Advantage} stated in Corollary~\ref{cor:advantage}. The detailed proofs are provided in Appendix~\ref{appx:proofs}.

\textbf{On the tightness of FPR bound established Theorem~\ref{thm:fpr_main}:} In practice, the upper bound for the per-sample FPR remains $\approx 0$ across all datasets, which corresponds to our $100\%$ per-sample attack success rate ( Sec. \ref{subsec:advantage_attack_success_rate_measurement}). We examine the upper FPR bound accross various settings in Appendix \ref{appx:upper_noniid_bound}.

\section{Experimental Results}\label{sec:fullexp}

We evaluate \textsc{PromptMIA} on four datasets: CIFAR-10, CIFAR-100 ~\citep{krizhevsky2009learning}, TinyImageNet ~\citep{le2015tiny}, and a synthetic 4-dataset benchmark constructed by pooling MNIST-M \citep{lee2021dranet}, Fashion-MNIST ~\citep{xiao2017fashion}, CINIC-10 ~\citep{darlow2018cinic}, and MMAFEDB \footnote{\url{https://www.kaggle.com/datasets/yuulind/mmafedb-clean}} and three different models: ViT-B32 ~\citep{dosovitskiy2020image}, ConViT ~\citep{d2021convit} and DeiT ~\citep{touvron2021training}. Experiments follow three axes: (1) measuring attack effectiveness via advantage and success rate and comparison against other federated MIAs; (2) testing robustness of \textsc{PromptMIA} against MIA defenses; and (3) conducting ablations on key hyperparameters ($M$, $N$, $\delta_{\min}$, $\Delta$) as well as federated settings. Details about experimental settings are given in Appx. \ref{appx:experiments}. 

\subsection{\textsc{PromptMIA Performance Evaluation}}
\label{subsec:advantage_attack_success_rate_measurement}
\begin{figure*}[t!]
    \centering
    \includegraphics[width=\textwidth]{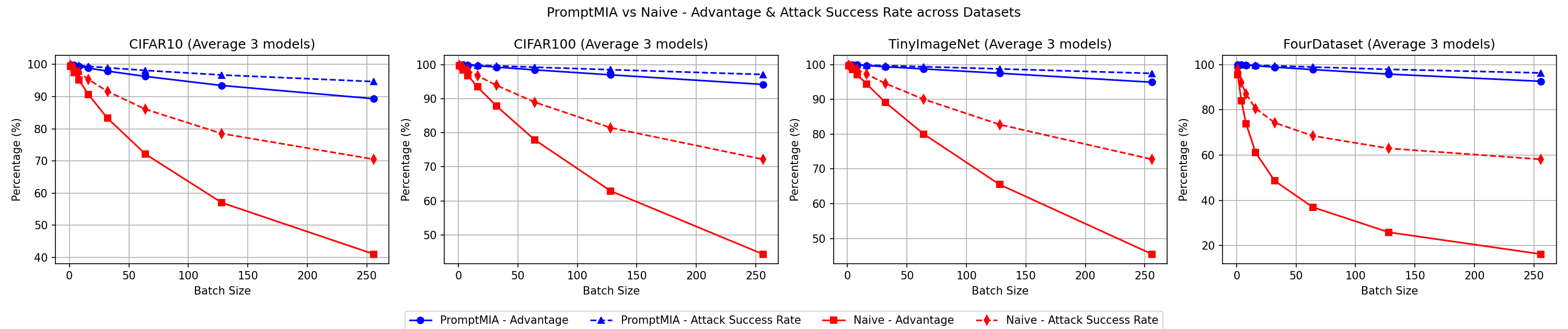}
    \caption{Performance of PromptMIA vs Naive \textbf{averaged across three models}. 
             Each subplot shows Advantage and A.S.R w.r.t Batch Size 
             across CIFAR10, CIFAR100, TinyImageNet, and FourDataset.}
    \label{fig:promptmia_vs_all_models}
\end{figure*}

We evaluate the performance of \textsc{PromptMIA} against FPT using two metrics: Advantage (Eq.~\ref{eq:advantage_def}) and Attack Success Rate (Eq.~\ref{eq:asr_def}). Unless otherwise specifically noted, we set $\delta_\text{min} = 0.02$ and $\Delta = 0.05$. Following \citep{wang2022learning}, the inital global prompt pool size is set at $M=20$, and the prompt selection size at $N=4$. 
In the case of batched update, given batch $B=\{(x_i,y_i)\}_{i=1}^\ell$, for each sample the client computes the per-sample selected key set $\hat{\mathcal K}_{x_i}$ and corresponding per-sample loss $\mathcal{L}_{x_i}$ (defined in Eq. \ref{eq:trainl2p}) and updates the batch-level set of chosen keys and prompts as: $\hat{\mathcal K}_B = \hat{\mathcal K}_B \bigcup \hat{\mathcal K}_{x_i}$ and $\hat{\mathcal P}_B = \hat{\mathcal P}_B \bigcup \mathcal{P}_{x_i}$. Batch-wise loss $\mathcal{L}_{B}$ is calculated by accumulating $\mathcal{L}_{x_i}$. The client updates the selected keys and prompts as $\hat{\mathcal K}_{B} \leftarrow \mathcal{K}_{B} - \mu \nabla_{\hat{\mathcal K}_{B}}\mathcal{L}_B$ and $\hat{\mathcal P}_{B} \leftarrow \hat{\mathcal P}_{B} - \mu \nabla_{\hat{\mathcal P}_{B}}\mathcal{L}_B$, where $\mu$ is the learning rate. After receiving the client’s updates, the server infers membership by checking whether all adversarial prompts are selected, i.e. $\mathbb{I}_{\{\mathcal{T} \in \mathcal{D}\}} = 1$ if \(\mathcal{K}_{\textsc{ADV}} \subseteq \hat{\mathcal K}_{B}\), and \(0\) otherwise.
Fig. \ref{fig:promptmia_vs_all_models} shows average results of three models over four different datasets. \textbf{Detailed results and error bands on individual models} are given in Appx. \ref{appx:res_models}. \textsc{PromptMIA} achieves near-perfect attack success rates across all models and datasets at small batch sizes, and maintains $>90\%$ ASR at larger batch sizes. In contrast, \textit{naive prompt injection  attack} collapses as the batch size increases, reaching Advantage of only $\approx 20\%$ against FourDataset when batch size $=256$.

\begin{figure}[t!]
    \centering
    \begin{minipage}[t]{0.60\textwidth}
        \centering
        \includegraphics[width=\textwidth]{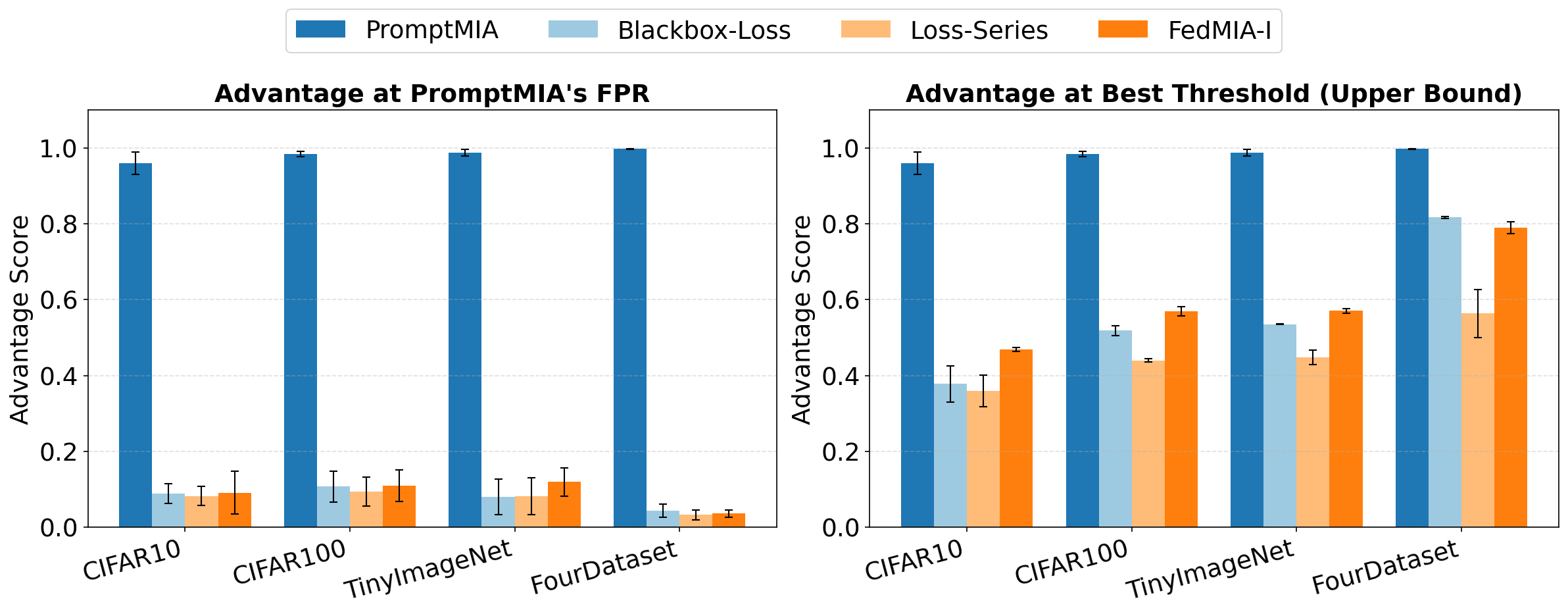}
        \caption{Comparison of \textsc{PromptMIA} vs.\ existing federated MIAs (batch size = 64).}
        \label{fig:compareattack}
    \end{minipage}\hfill
    \begin{minipage}[t]{0.37\textwidth}
        \centering
        \includegraphics[width=\textwidth]{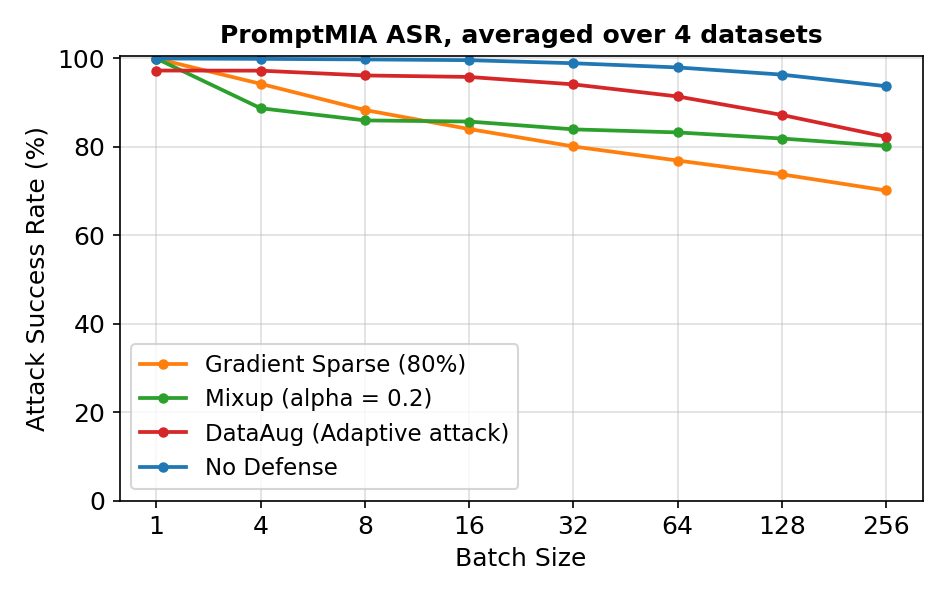}
        \caption{Impact of regularization-based MIA defenses.}
        \label{fig:comparedef}
    \end{minipage}
\end{figure}

\textbf{\textsc{PromptMIA} vs.\ existing FedMIAs.} We also compare \textsc{PromptMIA} against existing MIAs. Most MIAs rely on the observation that ML models behave differently on training data and on unseen data \citep{zhu2025fedmia}, and exploit signals such as gradient norms, gradient cosine similarity, or loss values. Gradient-based methods are not directly applicable to FPT, where clients exchange prompts rather than gradients. We therefore compare \textsc{PromptMIA} against MIAs that use loss values as the membership signal: Blackbox-Loss \citep{yeom2018privacy}, Loss-Series \citep{gu2022cs}, and FedMIA-I \citep{zhu2025fedmia}. Since these attacks output a continuous score, their Advantage depends on the decision threshold, and we report two settings (Fig.~\ref{fig:compareattack}). (i) At \textsc{PromptMIA}'s operating point, each baseline's threshold is set to match \textsc{PromptMIA}'s FPR on the same model. Here \textsc{PromptMIA} reaches an Advantage of 0.96--1.00, whereas the baselines remain at or below 0.12. (ii) At the best threshold, each baseline uses the threshold that maximizes its Advantage, chosen using the true membership labels. This is an upper bound that no real attacker can achieve. Even then, the baselines reach only 0.36--0.82, and \textsc{PromptMIA}, which uses no such tuning, outperforms all of them on every dataset. Structured comparision of \textsc{PromptMIA} and existing FedMIAs are given in Appx. \ref{appx:comparemia}.

\subsection{MIA Defenses against \textsc{PromptMIA}}\label{sec:MIA_DEF}

Although the primary objective of this work is not to propose novel defense mechanisms or conduct an exhaustive evaluation of all possible defense strategies against \textsc{PromptMIA}, we examine how standard MIA defenses interact with \textsc{PromptMIA}. Details about theses defenses are given in Appendix \ref{appx:miadefenses} and \ref{appx:systemdefenses}.

{\bf (I)}~\textbf{Modern MIA Defenses:} Most existing MIAs rely on models overfitting their training data \citep{zhu2025fedmia,gu2022cs,yeom2018privacy,li2023effective}, so modern MIA defenses are mostly regularization-based: gradient sparsification \citep{wangni2018gradient}, mixup \citep{huang2020instahide}, and data augmentation \citep{shorten2019survey}. Fig.~\ref{fig:comparedef} evaluates these defenses in a realistic setting, where the client applies the defense both during training and in the attacked round. Results are averaged over four datasets (CIFAR-10, CIFAR-100, TinyImageNet, and FourDataset) with ViT-B/32. \emph{Gradient sparsification} keeps only the top 20\% of gradient entries, which can prune the updates to the injected prompts. \emph{Mixup} and \emph{data augmentation} represent challenging scenarios, since they depart from the security game of Sec.~\ref{sec:securitygame}: the client no longer trains on the target itself, but on a modified version of it (a blend with another image, or an augmented copy). Even so, \textsc{PromptMIA} remains a significant threat under both gradient sparsification and mixup, including at large batch sizes. For data augmentation, we propose a simple modification that makes the attack adaptive to the defense by constructing adversarial keys using augmented copies of the target sample (Appx.~\ref{appx:adaptive-promptmia}). Overall, modern regularization-based MIA defenses do not eliminate the threat posed by \textsc{PromptMIA}.

{\bf (II)} Another line of defense is to use \textbf{Anomaly Detection Methods} to filter out adversarial prompts. We run four classical outlier detectors on the $M$-key global pool to flag the $N$ injected adversarial keys. We use an injection rate of $10\%$, i.e., each round is attacked with probability $0.1$ (Appendix \ref{appx:outlier}). Since \textsc{PromptMIA} succeeds in a single round, the true injection rate, and thus the attainable precision, is even lower. As shown in Table\ref{tab:outlier}, these methods are ineffective against \textsc{PromptMIA}, with Figures~\ref{fig:outlier} and \ref{fig:outlierbenign} visualizing their behavior. Although evaluating deep learning-based anomaly detection algorithms \citep{doswift, jiang2023adgym} would be interesting, we leave this for future work; moreover, the prompt pool is typically small, which raises doubts about the effectiveness of such data-hungry methods.
Moreover, the number of prompts in the prompt pool is typically small, which raises doubts about the effectiveness of deep learning based methods.

\begin{figure*}[ht!]
  \centering
  \includegraphics[width=\textwidth]{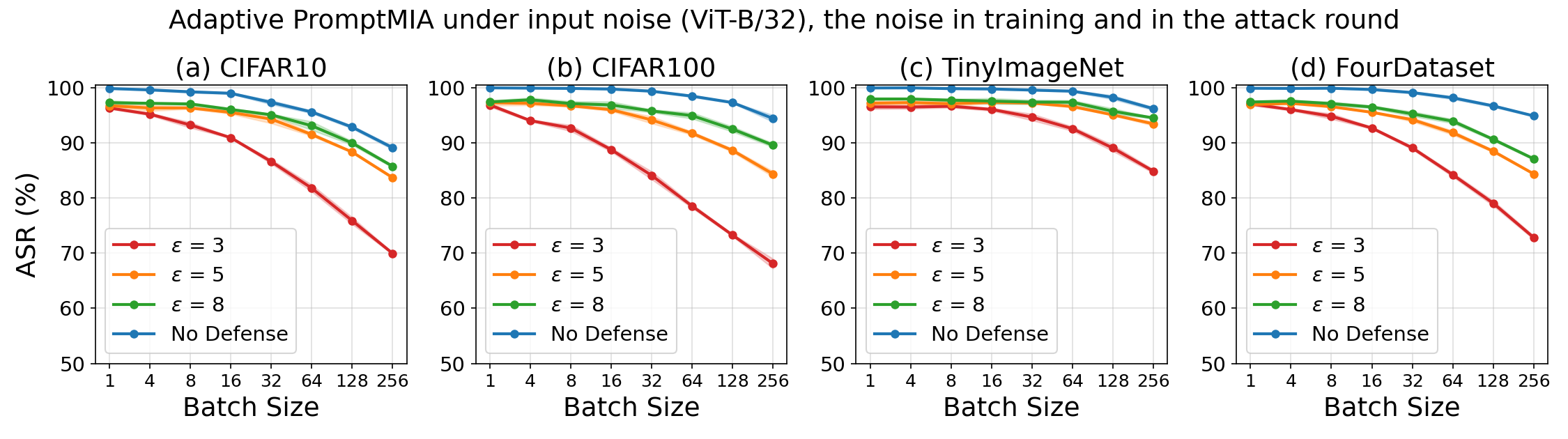}
  \caption{ASR of adaptive \textsc{PromptMIA} (ViT-B/32) under Input Noise Perturbation with different $\epsilon$.}
  \label{fig:dpres}
\end{figure*}

\noindent\textbf{(III) Input Noise Perturbation.}
Inspired by PixelDP \citep{lecuyer2019certified}, we add fresh Gaussian noise $\eta \sim \mathcal{N}(0,\sigma^2 I)$ to normalized input images \citep{lecuyer2019certified} during every local training step and during attack evaluation. We set $\sigma = L\sqrt{2\ln(1.25/\delta)}/\epsilon$, with $\delta = 1/n_{\mathrm{train}}$ and $L=1$, where $n_{\mathrm{train}}$ denotes the number of training samples. Smaller values of $\epsilon$ produce stronger input perturbations. Similar to the data augmentation setting, we evaluate an adaptive variant of \textsc{PromptMIA} that constructs adversarial keys using simulated noisy copies of the target (see Appendix~\ref{appx:adaptive-promptmia}). In particular, we measure the attack success rate (ASR) under $\epsilon \in \{3,5,8\}$ using ViT-B/32 across four datasets, as shown in Fig.~\ref{fig:dpres}. Table~\ref{tab:dp} reports the corresponding privacy--utility trade-off, assessed through attack success and classification accuracy. Defending against \textsc{PromptMIA} with input noise perturbation presents a pronounced privacy--utility trade-off. Stronger noise (e.g., $\epsilon=3$) reduces attack success at larger batch sizes, but at a substantial cost to model accuracy: on CIFAR-100, accuracy drops from $73.5\%$ without noise to $42.5\%$. Conversely, moderate noise levels ($\epsilon=5$ or $8$) better preserve accuracy but offer limited protection, leaving the model highly vulnerable to adaptive \textsc{PromptMIA}.

{\bf (IV)}~\textbf{System-level Defenses:}  We also analyze the impact of system-level defenses that modify the prompt selection or aggregation mechanism against \textsc{PromptMIA} such as randomized key indices, secure aggregation \cite{bonawitz2017practical}, randomized prompt selection via prompt dropout and cross-round key-drift auditing. In Appendix \ref{appx:systemdefenses}, we show that these defenses are either provably or empirically ineffective.


\begin{table}[ht!]
\begin{minipage}{0.48\linewidth}
\centering
\caption{Outlier detection results \textbf{averaged over datasets and models}. See Appx. \ref{appx:outlier}. }
\label{tab:outlier}
\scriptsize{\begin{tabular}{lccc}
\hline
\textbf{Method} & \textbf{Precision} & \textbf{Recall} & \textbf{F1} \\
\hline
IsolationForest\citep{liu2008isolation}    & 0.04(0.01) & 1.00(0.00) & 0.08(0.02) \\
LocalOutlierFactor\citep{breunig2000lof} & 0.00(0.00) & 0.00(0.00) & 0.00(0.00) \\
OneClassSVM\citep{manevitz2001one}        & 0.05(0.01) & 0.49(0.03) & 0.08(0.02) \\
EllipticEnvelope\citep{estimator1999fast}   & 0.03(0.01) & 0.24(0.03) & 0.05(0.02) \\
\hline
\end{tabular}
}
\end{minipage}
\hfill
\begin{minipage}{0.48\linewidth}
\centering
\caption{Model accuracy (\%) under input noise training with different $\epsilon$ (ViT-B32).}
\label{tab:dp}
\scriptsize{\begin{tabular}{lcccc}
\hline
\textbf{Dataset} & $\epsilon=3$ & $\epsilon=5$ & $\epsilon=8$ & \textbf{Non-DP} \\
\hline
CF-10     & 82.7 (0.6) & 87.9 (0.5) & 89.9 (0.5) & 93.5 (0.1) \\
CF-100    & 42.5 (0.2) & 54.4 (0.8) & 61.8 (0.8) & 73.5 (0.6) \\
TinyIN    & 69.3 (0.1) & 74.3 (0.1) & 77.7 (0.4) & 81.5 (0.2) \\
4-dataset & 38.9 (1.1) & 53.3 (1.3) & 60.2 (0.8) & 68.1 (1.3) \\
\hline
\end{tabular}
}
\end{minipage}

\end{table}

\subsection{Additional Experiments}
In Appx. \ref{appx:hyperparam_analysis}, we analyze how key hyperparameters impact \textsc{PromptMIA}’s performance. The most interesting finding is that the ASR of \textsc{PromptMIA} is much higher on the models that have been trained for a few epochs rather than randomly initialized keys (Fig. \ref{fig:ablation_traininground}). This corresponds to our theoretical findings in Section \ref{sec:theory} that FPR is lower when the benign data forms tight and compact clusters in the query space around the benign keys, which happens naturally during the training process (Appx. \ref{appx:trainingdynamic}). We provide additional experiments on various settings such  as non-iid and imbalalanced data (Appx. \ref{appx:hetero}), very large batch size (Appx. ~\ref{appx:larger_batch}), text and multimodel data (Appx. ~\ref{appx:modality}), large number of clients (Appx. \ref{appx:numclients}), target distinctiveness (Appx. \ref{appx:target_distinct}) and downstream task accuracy (Appx. \ref{appx:impact_acc}). We also discuss additional metrics such as Attack Sparsity, Computational Overhead and Scalability in Appx. \ref{appx:metrics}. We discuss \textsc{PromptMIA}'s limitations and future research directions in Appx. \ref{appx:limitations}.


\section{Related Work} \label{headings}
{\bf Membership Inference Attacks against Federated Learning.} 
The goal of MIAs in FL is to identify if a specific data point was part of a
client’s training set. Active Membership Inference (AMI) attacks involve a dishonest server poisoning the global models, e.g., maliciously modifying model parameters, before dispatching them to clients \citep{nguyen2023active, vu2024analysis}. Our work follows the standard AMI protocol. Recently, FedMIA \cite{zhu2025fedmia} proposed a three-step attack that leverages updates from all clients that can be integrated as an extension to existing attacks.

{\bf Federated Fine-Tuning of Foundation Models.} Federated fine-tuning avoids updating the entire model and instead allows clients to fine-tune only a subset of parameters, such as LoRA-\citep{qi2024fdlora, wang2024flora, fan2025helora} and adapter-based approaches \citep{cai2022fedadapter, ghiasvand2024communication}. Recently, prompt-based federated fine-tuning approaches have been proposed \citep{su2024federated, weng2024probabilistic, bai2024diprompt, feng2024cp}, which update soft prompts instead of full model weights. Our work focuses on MIAs under FPT setting since these approaches are architecturally orthogonal.

{\bf Relation to broader privacy attacks.} In in-context learning, adversaries can craft malicious prompts to extract confidential information in discrete prompts \citep{wen2024membership, duan2023flocks}. Active attacks such as backdoor injection \citep{zhao2024survey} or representation steering \citep{wu2025axbench, wu2025improved} are typically designed to manipulate model behavior or internal representations rather than infer membership (Appx. \ref{appx:backdoor}). Our work is the first to investigate the privacy risks of soft prompts in FPT.
\section{Ethical Considerations}\label{sec:ethics}
\textbf{Intended use.} \textsc{PromptMIA} is intended solely for privacy auditing and security evaluation of federated prompt-tuning (FPT) systems. We publish it to expose a new attack surface and to dispel the assumption that FPT guarantees data protection. While FPT reduces privacy risk by not sharing raw data, it does not eliminate that risk. All our experiments use public benchmark datasets in simulated federations; no deployed system was attacked, and no inference was performed against any real individual's data.

\textbf{Practical safeguards.} Clients should not treat a central aggregator as a trusted party, and should adopt additional safeguards in sensitive deployments despite their costs. Mechanisms that prevent the server from observing individual client updates, such as secure aggregation (Appendix~\ref{appx:systemdefenses}) or multi-party management of the prompt pool, reduce the attack from client-level to group-level membership inference, at additional communication and computation cost. 

\textbf{Responsible disclosure.} \textsc{PromptMIA} targets the design of the prompt-selection protocol in FPT rather than a defect in a particular implementation, and we are not aware of a commercial FPT deployment with a maintainer to whom conventional embargoed disclosure would apply. In place of vendor notification, we publish the attack so that the risk is known to researchers and implementers before such systems reach wider deployment. We release the code at \url{https://github.com/GivralNguyen/Prompt-MIA} with a clear intended-use statement, and we document the safeguards above together with their costs and limits, so that practitioners are not left with a demonstrated attack and no guidance.

\section{Conclusion}
In conclusion, we show that FPT introduces a new and critical privacy vulnerability. Through \textsc{PromptMIA}, we demonstrate that a malicious server can leverage adversarial prompts to reliably infer client membership, achieving high attack success rates across multiple datasets. Our theoretical analysis explains the attack's robustness by establishing a lower bound on the attack's advantage. Finally, evaluation of \textsc{PromptMIA} against existing MIA defenses reveals their limitations in this setting, underscoring the need for new defense strategies specifically designed for FPT.

\begin{ack}
This work is partially supported by NSF grant IIS-2416606 and NIH/NIDDK grant R01DK142809. Any opinions, findings, and conclusions expressed in this paper are those of the authors and do not necessarily reflect the views of the funding agencies.
\end{ack}

\bibliography{neurips_2026}
\bibliographystyle{unsrt}






\newpage
\appendix

\section{Appendix}
\subsection{Federated Prompt Tuning}
\label{appx:pfpt}
\begin{figure}[h!]
    \centering
    \includegraphics[width=0.85\textwidth]{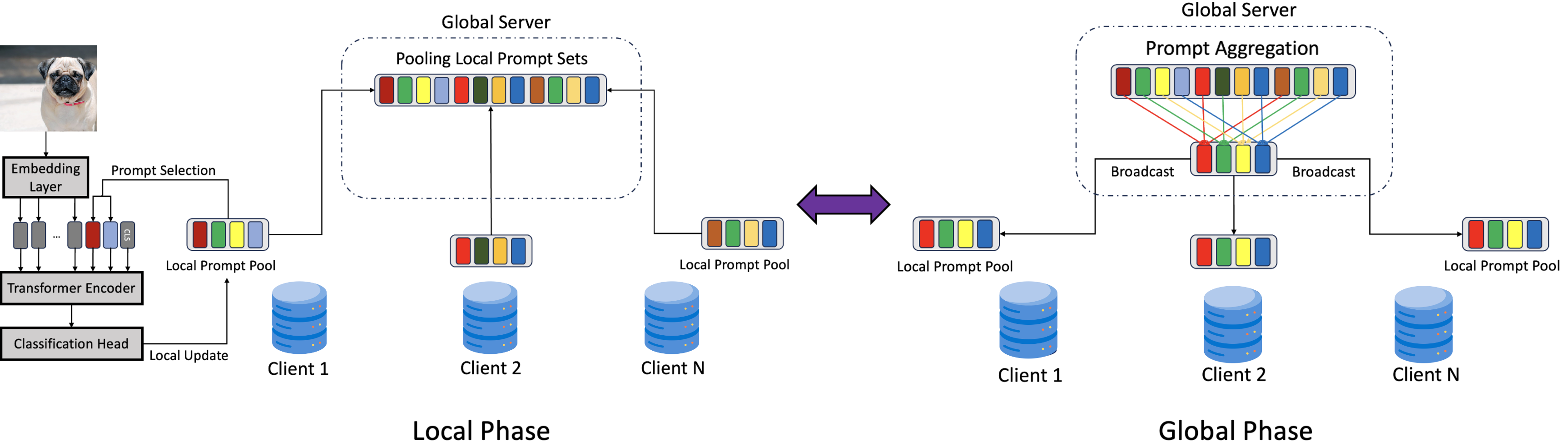}
    \caption{
   (Local Phase) each client samples and fine-tunes a subset of global summarizing prompts using a prompt-selection strategy; (Global Phase) the server aggregates all local prompt sets to refine the global prompt pool.
    }
    \label{fig:pfpt_workflow}
\end{figure}

Figure \ref{fig:pfpt_workflow} illustrates the workflow of federated prompt tuning. In federated prompt tuning, a central server maintains a global pool of prompts and keys. Each client selects the top-$N$ prompts most similar to its input features, updates those prompts and a lightweight classifier locally, and sends the updated prompts back. The server then aggregates the clients’ prompts to refine the global prompt pool without sharing raw data.
\subsection{Generating adversarial keys set $\mathcal{K}_\textsc{adv}$}
\label{appx:genkey}
To mount \textsc{PromptMIA}, the adversarial server must generate a set of keys
$\mathcal{K}_{\textsc{ADV}}$ that are more similar to the target query vector
$q(\mathcal{T})$ than all benign keys $\mathcal{K}_{\textsc{benign}}$ while remaining diverse enough to avoid detection.

\textbf{\textsc{GenKeyWithSimilarity}} (Alg.~\ref{alg:gen_key_with_similarity})
constructs a single adversarial key with a desired cosine similarity $s$ to the target query.
It first \emph{normalizes} the target query $q(\mathcal{T})$ to obtain the unit vector 
$\hat{q} = q(\mathcal{T}) / \|q(\mathcal{T})\|$. 
Then it \emph{samples a random vector} $r \in \mathbb{R}^{D_k}$ and 
\emph{removes the component of $r$ that lies along $\hat{q}$} by computing 
$o = r - \langle r, \hat{q}\rangle \hat{q}$. 
This ensures that $o$ is orthogonal to the target direction. 
Next, $o$ is normalized to $\hat{o} = o / \|o\|$, producing a unit vector exactly 
perpendicular to $\hat{q}$. 
To construct a vector with a desired cosine similarity $s$ to $\hat{q}$, the algorithm forms
\[
\hat{v} = s \cdot \hat{q} + \sqrt{1 - s^2} \cdot \hat{o},
\]
Finally, $\hat{v}$ is \emph{rescaled} to match the original norm of $q(\mathcal{T})$, 
producing the adversarial key 
$k_a = \hat{v} \cdot \|q(\mathcal{T})\|$.

\textbf{\textsc{GenAdvKeySet}} (Alg.~\ref{alg:gen_adv_key_set})
builds the entire adversarial key set $\mathcal{K}_{\textsc{ADV}}$.
It computes the maximum similarity $s_{\max}$ between the target query and existing
benign keys, then samples $N$ similarity scores uniformly from the interval
\[
[s_{\max}+\delta_{\min}, \; s_{\max}+\delta_{\min}+\Delta],
\]
ensuring that each adversarial key is slightly closer to the target query
than any benign key by at least $\delta_{\min}$ but not so close that all keys collapse to the target query.
Each sampled similarity $s_m$ is used to generate a key via
\textsc{GenKeyWithSimilarity}, producing a diverse set of keys that satisfy the
required similarity bounds.
\begin{algorithm}[H]
\caption{\textsc{GenKeyWithSimilarity}$(q(\mathcal{T}), s)$}
\label{alg:gen_key_with_similarity}
\begin{algorithmic}[1]
\REQUIRE Target query vector $q(\mathcal{T}) \in \mathbb{R}^{D_k}$, desired cosine similarity $s \in (0,1)$
\ENSURE Adversarial key $k_a \in \mathbb{R}^{D_k}$ such that $\kappa(k_a, q(\mathcal{T})) \approx s$ and $\|k_a\| = \|q(\mathcal{T})\|$
\STATE $\hat{q} \gets q(\mathcal{T}) / \|q(\mathcal{T})\|$ \hfill \COMMENT{Normalize the target vector}
\STATE $r\sim \mathcal{U}(0,1)^{D_k}$; $\hat{r} \gets r / \|r\|$ \hfill \COMMENT{Sample a random vector}
\STATE $o \gets \hat{r} - \langle \hat{r}, \hat{q} \rangle \cdot \hat{q}$ \hfill \COMMENT{Remove component along $\hat{q}$}
\STATE $\hat{o} \gets o / \|o\|$ \hfill \COMMENT{Normalize orthogonal component}
\STATE $\hat{v} \gets s \cdot \hat{q} + \sqrt{1 - s^2} \cdot \hat{o}$ \hfill \COMMENT{Combine to enforce similarity $s$}
\STATE $k_a \gets \hat{v} \cdot \|q(\mathcal{T})\|$ \hfill \COMMENT{Rescale to match original norm}
\STATE \textbf{return} $k_a$ \hfill \COMMENT{Adversarial key}
\end{algorithmic}
\end{algorithm}

\begin{algorithm}[H]
\caption{\textsc{GenAdvKeySet}$(q(\mathcal{T}), \mathcal{K}_{\textsc{benign}}, \delta_{\min}, \Delta, N)$}
\label{alg:gen_adv_key_set}
\begin{algorithmic}[1]
\REQUIRE Target query vector $q(\mathcal{T}) \in \mathbb{R}^{D_k}$, benign key set $\mathcal{K}_{\textsc{benign}}$, margins $\delta_{\min}, \Delta$, number of adversarial keys $N$
\ENSURE $\kappa\!\left(q(\mathcal{T}), k_{a_m}\right) \in 
\Bigl[\max_{k_b \in \mathcal{K}_{\textsc{benign}}} \kappa\!\left(q(\mathcal{T}), k_b\right) + \delta_{\min}, \;\;
\max_{k_b \in \mathcal{K}_{\textsc{benign}}} \kappa\!\left(q(\mathcal{T}), k_b\right) + \delta_{\min} + \Delta \Bigr], \forall\, k_{a_m} \in \mathcal{K}_{\textsc{adv}}$
\STATE $\hat{q} \gets q(\mathcal{T}) / \|q(\mathcal{T})\|$ \hfill \COMMENT{Normalize target query}
\STATE $s_{\max} \gets \max_{k_b \in \mathcal{K}_{\textsc{benign}}} \kappa(\hat{q}, k_b)$ \hfill \COMMENT{Maximum similarity to benign keys}
\FOR{$m = 1$ to $N$}
    \STATE Sample $s_m \sim \mathcal{U}(s_{\max} + \delta_{\min}, \; s_{\max} + \delta_{\min} + \Delta)$
    \STATE $k_{a_m} \gets$ \textsc{GenKeyWithSimilarity}$(q(\mathcal{T}), s_m)$
\ENDFOR
\STATE \textbf{return} Adversarial key set $\mathcal{K}_{\textsc{adv}} = \{k_{a_m}\}_{m=1}^N$
\end{algorithmic}
\end{algorithm}

\subsection{Adaptive \textsc{PromptMIA} against input-transforming defenses}\label{appx:adaptive-promptmia}
Input-transforming defenses modify each training sample before it reaches the model, so a client selects prompts (Eq.~\ref{eq:cosinedistance}) using the query of a \emph{transformed} target rather than the clean query $q(\mathcal{T})$ around which the adversarial keys $\mathcal{K}_{\textsc{adv}}$ were built (Eq.~\ref{eq:adv_constraint}). Because the transformation is random and redrawn at every local step, a key tuned to $q(\mathcal{T})$ may no longer be selected, which weakens the attack. Following Kerckhoffs's principle \citep{shannon1949communication} that a system's security must not rely on the secrecy of its mechanism (i.e., security through obscurity), as well as the standard assumption in modern membership inference that the adversary has full knowledge of the client's training protocol \citep{nasr2019comprehensive, carlini2022membership}, we adapt \textsc{PromptMIA} as follows. The server knows the client's defense pipeline, since it is part of the training protocol, but not the specific random transformation applied in any given step. In both settings below, the server estimates the \emph{expected} transformed query and builds keys that remain selected under the transformation's randomness; the two settings differ only in how the transformed copies are generated.

Concretely, the server draws $K$ transformed copies $\tilde{\mathcal{T}}_1,\dots,\tilde{\mathcal{T}}_K$ of the target and takes their mean query direction
\begin{equation}\label{eq:adaptive_center}
c = \frac{\bar{q}}{\lVert \bar{q}\rVert}, \qquad \bar{q} = \frac{1}{K}\sum_{k=1}^{K} \frac{q(\tilde{\mathcal{T}}_k)}{\lVert q(\tilde{\mathcal{T}}_k)\rVert}.
\end{equation}
It overwrites the $N$ keys nearest $c$ with adversarial keys constructed as in Alg.~\ref{alg:gen_adv_key_set}, but centered on $c$ instead of $q(\mathcal{T})$. The only remaining choice is how strongly to align the keys with $c$: rather than fixing the margin at the lower end of Eq.~\ref{eq:adv_constraint}, the server raises the target similarity to the \emph{smallest} value at which at least $95\%$ of the $K$ copies still select all $N$ adversarial keys, , falling back to fully
aligned keys (similarity $1$) if no value reaches this coverage. Requiring high coverage makes the keys robust to the random transformation, while choosing the smallest sufficient similarity keeps the keys as far from $c$ as possible, limiting the false positives induced by non-member batches. We use $K=32$.

\subsubsection{Data augmentation}\label{appx:adaptive-augment}
For data augmentation \citep{shorten2019survey}, each copy $\tilde{\mathcal{T}}_k$ is an independent augmentation of the target drawn from the client's augmentation pipeline, so the client selects prompts with $q(\tilde{\mathcal{T}})$ rather than $q(\mathcal{T})$. The server applies the construction above directly, sampling the $K$ copies from the known augmentation distribution.

\subsubsection{Input noise perturbation}\label{appx:adaptive-noise}
For the input-noise defense of Sec.~\ref{sec:MIA_DEF} \citep{lecuyer2019certified}, each copy is a noisy target $\tilde{\mathcal{T}}_k = \mathcal{T} + \eta_k$ with $\eta_k \sim \mathcal{N}(0,\sigma^2 I)$ added to the normalized input, using the same $\sigma$ as the client's trained model. The adaptive construction is otherwise identical: the server forms $c$ from Eq.~\ref{eq:adaptive_center} over the $K$ noisy copies, overwrites the $N$ keys nearest $c$, and calibrates the similarity to the $95\%$-coverage threshold, again with $K=32$.

\subsection{Comparision of \textsc{PromptMIA} and existing Federated MIAs.}

In Table \ref{tab:mia_comparison}, we provide a structured comparison summarizing existing MIA approaches in federated learning, along with their assumptions, attack surfaces, and key differences from PromptMIA. To summarize, the attack surface of PromptMIA is fundamentally different from traditional  (sets of global prompts v.s. gradient vectors/ loss values). Existing MIAs are based on the observation that ML models behave differently on training data and unseen data and can be circumvented through regularization-based defenses. PromptMIA also significantly outperforms these methods ( Fig. \ref{fig:compareattack}). To ensure a fair comparison, we report Blackbox-Loss, Loss-Series, and FedMIA-I both at their best threshold, which maximizes Advantage over the full ROC curve and is an upper bound for these attacks since it is chosen using the evaluation labels, and at a threshold matched to \textsc{PromptMIA}'s FPR on the same model, using the largest FPR that does not exceed it.

\begin{table}[h]
\centering
\small
\begin{tabularx}{\textwidth}{@{} c l l X X @{}}
\toprule
\# & \textbf{Attack} & \textbf{Attack Surface} & \textbf{Attack Methodology} & \textbf{Assumptions} \\ \midrule
1 & \begin{tabular}[c]{@{}l@{}}Blackbox-Loss \citep{yeom2018privacy}\\ Loss-Series \citep{gu2022cs}\end{tabular} & Data Loss & Model tends to have lower loss on training points than on non-training points. [2] uses loss values over multiple rounds. & Bounded loss and Gaussian error distributions \\ \addlinespace
2 & Grad-Cosine \citep{li2023effective} & Gradient vector & The distributions of the cosine similarity of instances from the training data (members) and from non-members are different. & Gradient vectors of different instances are orthogonal \& their cosine similarity is Gaussian. \\ \addlinespace
3 & FedMIA \citep{zhu2025fedmia} & \begin{tabular}[c]{@{}l@{}}Data Loss (Fed-MIA-I),\\ Gradient vector (Fed-MIA-II)\end{tabular} & Update distribution of clients trained with target data is different from non-target clients & Update distribution follows Gaussian distribution \\ \addlinespace
4 & \textbf{PromptMIA (ours)} & Global prompt pool in FPT & Attacker injects adversarial prompts that will always be updated when target data is present. & Each query vector is modeled as a Gaussian mixture over benign keys \\ \bottomrule
\end{tabularx}
\caption{Comparison of \textsc{PromptMIA} and existing Membership Inference Attacks.}
\label{tab:mia_comparison}
\end{table}
\label{appx:comparemia}

\subsection{Membership Inference Defenses}
\label{appx:miadefenses}
Although the primary objective of this work is not to propose novel defense mechanisms or conduct an exhaustive evaluation of all possible defense strategies against \textsc{PromptMIA}, we examine how standard MIA defenses interact with \textsc{PromptMIA}. 


\textbf{Noise Perturbation}
Rather than obfuscating gradient updates, clients can add calibrated noise directly to the input image, which provides $(\epsilon,\delta)$-DP per noisy access with respect to input changes of bounded $\ell_2$ norm \citep{lecuyer2019certified}. This introduces randomness to Eq. \ref{eq:cosinedistance}:
\begin{equation}
\label{eq:cosinedistancenoisy}
\mathcal{K}_x
= \operatorname*{argmax}_{\{s_i\}_{i=1}^{N} \subseteq [M]}
\sum_{i=1}^{N} \kappa\!\big(q(x + \eta), k_{s_i}\big),
\qquad \eta \sim \mathcal{N}(0, \tilde{\sigma} I).
\end{equation}
Increasing the noise variance strengthens the protection by making the prompt selection less predictable and reducing the effectiveness of adversarially crafted keys. However, this also comes at the high cost of model utility.

\textbf{Regularization-based MIA Defenses:} We evaluate \textsc{PromptMIA} against regularization-based MIA Defenses, namely Gradient Sparsification \citep{wangni2018gradient}, MixUp \citep{huang2020instahide} and Data Augmentation\citep{shorten2019survey}. For Gradient Sparsification, we use $80\%$ sparsification rate. For MixUp, we use $\alpha_\text{mixup}=0.2$. For Data Augmentation, we employ a combination of random cropping, horizontal flipping, and color jittering.

\textbf{Anomaly detection techniques}
We test commonly used anomaly detection techniques in machine learning against $\textsc{PromptMIA}$. 
Specifically, we consider the following classical approaches: 
\texttt{IsolationForest} \citep{liu2008isolation}, \texttt{LocalOutlierFactor} \citep{breunig2000lof}, \texttt{OneClassSVM} \citep{manevitz2001one}, and \texttt{EllipticEnvelope} \citep{estimator1999fast}. 
These methods are widely used, well established in the literature, and directly available in \textsc{scikit-learn} library \citep{kramer2016scikit}. In our implementation, we use \textsc{scikit-learn} version 1.5.2. Although it would be interesting to evaluate deep learning-based anomaly detection algorithms \citep{doswift, jiang2023adgym} 
(e.g., autoencoders, VAEs, GAN-based models) against $\textsc{PromptMIA}$, 
we leave this for future work. 
Moreover, the number of prompts in the prompt pool is typically small (20 in our experiments), 
which raises doubts about the effectiveness of deep learning based methods in this setting. 
We therefore restrict our evaluation to the aforementioned classical methods. We briefly describe the anomaly detection algorithms used in this paper:
\begin{itemize}
\item \textbf{Isolation Forest} detects anomalies by recursively partitioning the feature space with random splits. Each sample’s path length, or the number of splits needed to isolate it in a random tree is shorter for outliers, as they are easier to separate from the bulk of the data. Averaging this path length over a forest of random trees yields an anomaly score: points with shorter average paths are more likely to be anomalous. 

\item \textbf{Local Outlier Factor (LOF)} detects anomalies by comparing the local density of each sample to that of its $k$-nearest neighbors. Normal points have similar density to their neighbors, while outliers lie in sparser regions. The LOF score is the ratio of the average neighbor density to the sample’s own density; values $\text{LOF}\!>\!1$ indicate potential outliers. LOF captures both local and global structure, making it effective in datasets with varying densities.

\item \textbf{One-Class Support Vector Machine (OCSVM)} is an unsupervised anomaly detection method derived from the Support Vector Machine framework. Instead of separating multiple classes, OCSVM learns a decision boundary that encloses the majority (normal) data in feature space, labeling points outside this region as anomalies or novelties. It works by maximizing the margin around normal data to create a robust “normalcy region” using a kernel function (commonly the radial basis function) to capture non-linear patterns. 

\item \textbf{Elliptic Envelope} is an outlier detection method that assumes inliers follow a known distribution, typically Gaussian. It fits a robust estimate of the data’s mean and covariance (using the Minimum Covariance Determinant estimator) to capture the central elliptical shape of normal data while ignoring outliers. Points are then scored by their Mahalanobis distance from this fitted ellipse, with distant points flagged as anomalies. This approach is effective when the normal data distribution is approximately Gaussian.
\end{itemize}

\subsection{System-level Defenses}
\label{appx:systemdefenses}
We also analyze the impact of system-level defenses that modify the prompt selection or aggregation mechanism against \textsc{PromptMIA}. 
\begin{itemize}
    \item \textbf{Randomized key indices.} We consider the setting where, after training, clients randomly permute the indices of their prompt keys before sending updates to the server to prevent the server from knowing exactly which prompts were updated. ~However, this defense is fundamentally ineffective against \textsc{PromptMIA}. The server already stores the previous prompt pool and can check whether each adversarial prompt appears in the client-updated pool via content matching, which is unaffected by index permutation. It then knows that the client selected and updated all adversarial prompts when no match is found. Index randomization is thus provably ineffective as it cannot prevent the server from knowing which prompts were updated.

    \item \textbf{Secure Aggregation.} Another strategy is to apply secure aggregation \cite{bonawitz2017practical} so that individual client updates are hidden from the server. The intuition is that, if the server only observes an aggregated result rather than each client's prompt updates, it may be unable to determine whether a particular adversarial prompt was selected. However, current secure aggregation protocols are developed for linear aggregation (that is, weighted averaging) of local model parameters, and do not extend to probablistic aggregation methods commonly used in federated prompt tuning. Weighted averaging is not sufficient in non-IID settings where local parameters must first be aligned before aggregation in order to prevent important features from collapsing into less informative representations due to semantic misalignment~\cite{weng2024probabilistic}.~To address this, the recent PFPT work~\cite{weng2024probabilistic} developed an aligned and aggregated mechanism based on probabilistic non-parametric clustering which was shown to achieve substantially better performance than weighted averaging in non-IID settings.~However, both the alignment and the aggregation steps in PFPT are inherently non-linear, making it unclear how existing secure aggregation methods could be generalized to support such a mechanism. Furthermore, while Secure aggregation is generally effective at defending against modern MIAs; however, the associated computational and communication costs remain a significant barrier to practical implementation. \citep{zhu2025fedmia}. Based on these observations, a possible future resereach direction is to develop SecAgg-compatible probablistic aggregation mechanisms that are optimized for optimizing for communication and computation efficiency.
      
    \item \textbf{Prompt Dropout.} Since \textsc{PromptMIA} relies on the adversarial keys being selected, introducing randomness into the prompt selection (e.g., via prompt dropout) can reduce the attack success rate. To account for this, when clients apply dropout, the server predicts a target as a member if at least 50\% of the adversarial keys are selected, rather than all of them. With the same dropout rate of 0.2, 0.3 or 0.4, the attack success rate averaged over the four datasets remains substantial (Fig. \ref{fig:asr_dropout}). While there are certainly more complex mechanisms for adapting to dropout, such as tuning the decision threshold to the dropout rate or combining key selections across multiple rounds, this simple rule is enough to keep the attack effective.
    \begin{figure}[!h]
        \centering
        \includegraphics[width=0.8\linewidth]{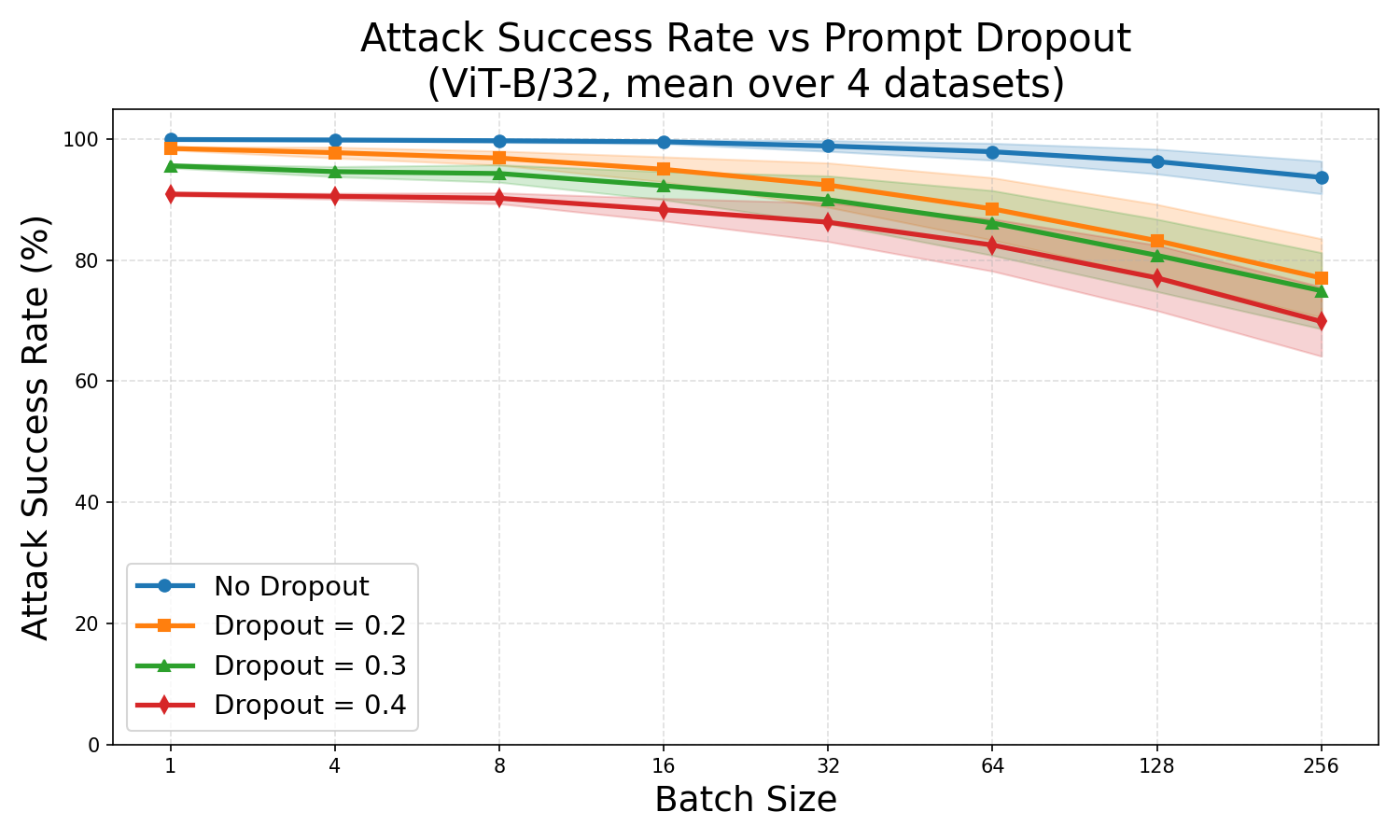}
        \caption{
            \textbf{Impact of Prompt Dropout on attack success rate.}
        }
        \label{fig:asr_dropout}
    \end{figure}
    \item \textbf{Cross-round key-drift auditing.} We evaluate \textsc{PromptMIA} against a stronger system-level defense in which clients audit the global prompt pool for anomalous key drift across consecutive rounds, a defense we term \emph{cross-round key-drift auditing}. Since a client already receives the global prompt pool each round, it can store the previous round's pool and compute the per-key round-over-round displacement $d_i = \|k_i^{(t)} - k_i^{(t-1)}\|$, flagging keys with $d_i > \mathrm{median}(d) + c \cdot \mathrm{MAD}(d)$, where MAD is the Median Absolute Deviation. This appears to target \textsc{PromptMIA}'s fingerprint, since injection overwrites $N$ keys while honest aggregation evolves each key smoothly. Two effects defeat it. First, because the selection mechanism (Eq.~\ref{eq:cosinedistance}) depends only on cosine similarity, which is invariant to key norm, the adversary rescales each adversarial key to match the norm of the benign key it replaces, so the injected keys are not unusually large and the attack success rate is unchanged. Second, and more fundamentally, the auditor has an irreducible false-positive rate on honest training drift: benign keys do not all move equally between rounds, so a MAD-based threshold always flags the upper tail even when no adversarial key is present. Since \textsc{PromptMIA} injects in a \emph{single} communication round, these false positives on the other rounds dominate precision, exactly as for the outlier detectors in Table~\ref{tab:outlier} and Fig.~\ref{fig:outlierbenign}. Table~\ref{tab:driftaudit} reports the F1 of the auditor against norm-rescaled \textsc{PromptMIA} when accounting for the injection rate: at a conservative rate of one attacked round in ten, and at the realistic rate of a single attacked round in the $60$-round schedule.
    \begin{table}[h!]
    \centering
    \caption{F1 of cross-round key-drift auditing against norm-rescaled \textsc{PromptMIA}, averaged over four datasets (ViT-B/32). Low F1 indicates the audit fails to reliably detect the injected keys. Accounting for clean (un-attacked) rounds, in which the auditor still flags benign keys, sharply lowers F1; at the realistic single-round injection rate the auditor is essentially ineffective.}
    \label{tab:driftaudit}
    \scriptsize{\begin{tabular}{lccc}
    \hline
    \textbf{Injection rate} & $c=1$ & $c=2$ & $c=3$ \\
    \hline
    $1/10$ (conservative) & 0.136 & 0.256 & 0.400 \\
    $1/60$ (single round) & 0.025 & 0.052 & 0.099 \\
    \hline
    \end{tabular}}
    \end{table}

    \item We do not claim that no future audit or secure protocol can defeat the attack. Rather, these results
    underscore the urgent need for more research on privacy defenses against this new prompt-based
    attack surface
    
    \end{itemize}

\subsection{Experimental Settings}
\label{appx:experiments}
\paragraph{Datasets.} 
We evaluate our methods on four widely used vision benchmarks: 
CIFAR-10 and CIFAR-100~\citep{krizhevsky2009learning}, 
TinyImageNet~\citep{le2015tiny}, and a synthetic benchmark referred to as 
\emph{4-dataset}. The 4-dataset benchmark is constructed by pooling 
four diverse datasets: 
1) MNIST-M~\citep{lee2021dranet}, 
2) Fashion-MNIST~\citep{xiao2017fashion}, 
3) CINIC-10~\citep{darlow2018cinic}, and 
4) MMAFEDB\footnote{\url{https://www.kaggle.com/datasets/yuulind/mmafedb-clean}}. 
For 4-dataset, we use a total of 118{,}629 training and 10{,}006 test samples, 
with 30{,}000 training and 2{,}500 test examples drawn from each dataset (except for MMAFEDB which contributes 28,629/2,506). For CIFAR-10, CIFAR-100, and TinyImageNet, 
we adopt the standard train/test splits provided by the official datasets. 

\paragraph{Federated learning setup.} 
To simulate heterogeneous client data, we partition datasets using a Dirichlet 
distribution with concentration parameter $\alpha=0.5$, which produces non-i.i.d. 
label distributions across clients. We consider a federation of 80 clients, with 
10 randomly selected in each communication round, and aggregate client updates with \textsc{FedAvg}~\citep{mcmahan2017communication}. For FourDataset, each of the four source datasets is partitioned among 20 clients. Each selected client trains with batch size 64, using the Adam optimizer with learning rate $1\!\times\!10^{-3}$ for the keys and $1\!\times\!10^{-4}$ for the prompts and the classifier. Training is conducted for 60 communication rounds. The prompt pool has $M=20$ key-prompt pairs, of which the top $N=4$ are selected for each input. We set hyperparameter $\lambda$ in Eq.~\ref{eq:trainl2p} to 0.5, consistent with \citep{wang2022learning}. We use three different frozen pretrained backbones: ViT-B/32 \citep{dosovitskiy2020image}, DeiT-B/16 \citep{touvron2021training}, and ConViT-Base \citep{d2021convit}. These settings apply unless stated otherwise.

\paragraph{PromptMIA setup.} 
 Unless otherwise specifically noted, we set $\delta_\text{min} = 0.02$ and $\Delta = 0.05$. Following \citep{wang2022learning}, the inital global prompt pool size is set at $M=20$, and the prompt selection size at $N=4$. 

\paragraph{Implementation details.} 
All experiments were conducted on a Linux workstation running Ubuntu 20.04 LTS, 
equipped with an Intel(R) Xeon(R) CPU E5-2697 v4 @ 2.30GHz (18 cores 36 threads), 
384GB RAM, and two NVIDIA RTX A6000 GPU (48GB VRAM each). Our implementation is 
based on PyTorch~2.0 with CUDA~12.2. 

\subsection{Proofs}
\label{appx:proofs}
\begin{theorem31} [True Positive Rate]
Let $\mathcal{K}_{\textsc{adv}} = \{k_{a_m}\}_{m=1}^N$ be the set of $N$ adversarial keys generated by Algorithm \ref{alg:gen_adv_key_set} with parameters $\delta_{\min} > 0$ and $\Delta \ge 0$. Let $\mathcal{K}_{\textsc{benign}}$ be the set of $M-N$ benign keys. If the client's dataset $\mathcal{D}$ contains the target sample $\mathcal{T}$ (i.e., $b=1$) and the client's selection mechanism (Eq. \ref{eq:cosinedistance}) selects the top-$N$ prompts based on highest cosine similarity (lowest cosine distance), the set of selected keys $\hat{\mathcal{K}}_\mathcal{T}$ for the query $q(\mathcal{T})$ will be exactly the adversarial set:
$\hat{\mathcal{K}}_\mathcal{T} = \mathcal{K}_{\textsc{adv}}$. Consequently, the True Positive Rate (TPR) of \textsc{PromptMIA} is 1.
$$TPR = \Pr[b'=1 \mid b=1] = 1$$
\end{theorem31}

\begin{proof}
Based on Algorithm \ref{alg:gen_adv_key_set}, every generated adversarial key $k_{a_m} \in \mathcal{K}_{\textsc{adv}}$ have a cosine similarity $s_m$ to $q(\mathcal{T})$ such that:
$$s_m \ge s_{\max} + \delta_{\min}, \qquad \text{where } s_{\max} = \max_{k_b \in \mathcal{K}_{\textsc{benign}}} \kappa(q(\mathcal{T}), k_b)$$

Since $\delta_{\min} > 0$, we have:

$$\min_{k_a \in \mathcal{K}_{\textsc{adv}}} \kappa(q(\mathcal{T}), k_a) > \max_{k_b \in \mathcal{K}_{\textsc{benign}}} \kappa(q(\mathcal{T}), k_b)$$

Because cosine distance $\gamma(\cdot, \cdot)$ is a monotonically decreasing function of cosine similarity $\kappa(\cdot, \cdot)$, we have an equivalent expression:

$$\max_{k_a \in \mathcal{K}_{\textsc{adv}}} \gamma(q(\mathcal{T}), k_a) < \min_{k_b \in \mathcal{K}_{\textsc{benign}}} \gamma(q(\mathcal{T}), k_b)$$
When the client processes $\mathcal{T} \in \mathcal{D}$, it computes $q(\mathcal{T})$ and selects the top-$N$ keys with the smallest distance $\gamma$ (Eq. \ref{eq:cosinedistance}). Since all $N$ adversarial keys have a smaller distance to $q(\mathcal{T})$ than all $M-N$ benign keys, the top-$N$ selected keys must be exactly the set $\mathcal{K}_{\textsc{adv}}$.

The adversary's guessing rule $\mathcal{A}_{\mathsf{GUESS}}$ predicts $b'=1$ if all adversarial prompts $\mathcal{P}_{\textsc{adv}}$ are updated. Since $b=1$, these prompts will be selected and updated. Therefore, $\Pr[b'=1 \mid b=1] = 1$.
\end{proof}

\begin{lemma32}\label{lem:flip_prob}
Let $q(x)$ be a non-member query drawn from the $i$-th component of the GMM, such that its conditional distribution is $q(x) \mid (x \in \text{cluster } i) \sim \mathcal{N}(k_{b_i}, \sigma_i^2 I)$. The conditional probability $\Pr(E_i)$, that $q(x)$ selects all $N$ adversarial keys $\mathcal{K}_{ADV} = \{k_{a_1}, \dots, k_{a_N}\}$ as its $N$ closest centroids, is bounded by:
$$\Pr(E_i) \le \min_{\substack{1 \le j \le N \\ 1 \le l \le M-N}} \Phi\left( \frac{(k_{a_j}-k_{b_l})^T k_{b_i}}{\sigma_i\|k_{a_j} - k_{b_l}\|} \right)$$
where $\Phi(\cdot)$ is the Cumulative Distribution Function (CDF) of the standard normal distribution.
\end{lemma32}

\begin{proof}
Let $E_i$ be the event that a query $q(x)$, drawn from the $i$-th component of the GMM (i.e., $q(x) \mid (x \in \text{cluster } i) \sim \mathcal{N}(k_{b_i}, \sigma_i^2I)$), selects all $N$ adversarial keys. This occurs when all adversarial keys have a higher similarity to $q(x)$ than all $M-N$ benign keys. This is formally defined as:
$$E_i = \left\{ \min_{j=1...N} q(x)^T k_{a_j} > \max_{l=1...M-N} q(x)^T k_{b_l} \right\}$$
Here, we omit the normalization factors, since both the query and all keys are $\ell_2$-normalized. This is an intersection of $N \times (M-N)$ pairwise events, $E_i = \bigcap_{j=1}^{N} \bigcap_{l=1}^{M-N} A_{jl}$, where:
    $$A_{jl} = \left\{ q(x)^T k_{a_j} > q(x)^T k_{b_l} \right\} \iff A_{jl} = \left\{ (k_{a_j} - k_{b_l})^T q(x) > 0 \right\}$$

The probability of an intersection of events is less than or equal to the probability of the single least likely event in that set. This gives us a strict upper bound:
    $$\Pr(E_i) = \Pr\left(\bigcap_{j,l} A_{jl}\right) \le \min_{\substack{1 \le j \le N \\ 1 \le l \le M-N}} \Pr(A_{jl})$$

Now, we find the probability of a single pairwise event $A_{jl}$.
Let $Y_{jl} = (k_{a_j} - k_{b_l})^T q(x)$. Based on the GMM defined in Assumption 2, the conditional distribution of $q(x)$ is a multivariate Gaussian $q(x) \mid (x \in \text{cluster } i) \sim \mathcal{N}(k_{b_i}, \sigma_i^2I)$.
The variable $Y_{jl}$ is a linear projection of $q(x)$, so it also follows a 1D Gaussian distribution. We find its mean $E[Y_{jl}]$ and variance $\text{Var}[Y_{jl}]$:

\textbf{Mean:} 
\begin{align*}
    E[Y_{jl}] &= E[(k_{a_j} - k_{b_l})^T q(x)] = (k_{a_j} - k_{b_l})^T E[q(x)] = (k_{a_j} - k_{b_l})^T k_{b_i} 
\end{align*}

\textbf{Variance:}
\begin{align*}
    \text{Var}[Y_{jl}] &= \text{Var}((k_{a_j} - k_{b_l})^T q(x)) = (k_{a_j} - k_{b_l})^T \text{Cov}(q(x)) (k_{a_j} - k_{b_l}) \\
    &=(k_{a_j} - k_{b_l})^T (\sigma_i^2I) (k_{a_j} - k_{b_l}) = \sigma_i^2 \|k_{a_j} - k_{b_l}\|^2
\end{align*}

We want to find $\Pr(A_{jl}) = \Pr(Y_{jl} > 0)$. We standardize $Y_{jl}$ to $Z \sim \mathcal{N}(0, 1)$ as:
$$\Pr(Y_{jl} > 0) = \Pr\left( Z > \frac{0 - E[Y_{jl}]}{\sqrt{\text{Var}(Y_{jl})}} \right) = \Pr\left( Z > -\frac{E[Y_{jl}]}{\sqrt{\text{Var}(Y_{jl})}} \right)$$

Using the identity $\Pr(Z > -z) = \Phi(z)$, the probability is $\Phi(z_{ijl}(\mathcal{K}_{ADV}))$, where the $z$-score is:
        $$z_{ijl}(\mathcal{K}_{ADV}) = \frac{E[Y_{jl}]}{\sqrt{\text{Var}(Y_{jl})}} = \frac{(k_{a_j} - k_{b_l})^T k_{b_i}}{\sigma_i \|k_{a_j} - k_{b_l}\|}$$

Substituting this result into the original inequality, we have the final bound:
$$\Pr(E_i) \le \min_{\substack{1 \le j \le N \\ 1 \le l \le M-N}} \Phi\left( \frac{(k_{a_j} - k_{b_l})^T k_{b_i}}{\sigma_i \|k_{a_j} - k_{b_l}\|} \right)$$
\end{proof}

\begin{theorem33} [False Positive Rate]
\label{thm:fpr}
The per-sample False Positive Rate (FPR) is bounded by:
$$FPR = \Pr[b'=1 \mid b=0] \le \max_{1\leq i\leq M-N} \left( \min_{\substack{1 \le j \le N \\ 1 \le l \le M-N}} \Phi\left( z_{ijl} \right) \right)$$
where $\mathcal{K}_{ADV} = \{k_{a_1}, \dots, k_{a_N}\}$ is the set of $N$ adversarial keys drawn from the shell $S$, and $z_{ijl}$ is the $z$-score: 
$$z_{ijl} = \frac{(k_{a_j} - k_{b_l})^T k_{b_i}}{\sigma_i \|k_{a_j} - k_{b_l}\|}$$
\end{theorem33}

\begin{proof}
With batch size of $1$, the client has a single non-member sample $x$. The FPR is the probability of the event $E_{FP}$ that $q(x)$ selects all $N$ adversarial keys.
    $$FPR = \Pr(E_{FP}) = \Pr\left( \min_{j=1...N} q(x)^T k_{a_j} > \max_{l=1...M-N} q(x)^T k_{b_l} \right)$$

Let $p_i$ be the prior probability (or GMM mixing weight) that a single non-member sample $x$ belongs to benign cluster $i$ (such that $\sum_{i=1}^{M-N} p_i = 1$). We find the total global probability by marginalizing over all $M-N$ components of the GMM that $q(x)$ could be drawn from:
    $$FPR = \sum_{i=1}^{M-N} \Pr(E_{FP} \mid q(x) \in \text{cluster } i) \cdot p_i$$
    where $p_i = \Pr(q(x) \in \text{cluster } i)$ is the mixing weight for GMM component $i$.

The term $\Pr(E_{FP} \mid q(x) \in \text{cluster } i)$ is exactly the probability $P(E_i)$ in Lemma 3.2. We substitute its bound:
    $$\Pr(E_{FP} \mid i) \le \min_{\substack{1 \le j \le N \\ 1 \le l \le M-N}} \Phi\left( z_{ijl}(\mathcal{K}_{ADV}) \right)$$

 We place this bound into the above sum:
    $$FPR \le \sum_{i=1}^{M-N} p_i \cdot \left( \min_{\substack{1 \le j \le N \\ 1 \le l \le M-N}} \Phi\left( z_{ijl}(\mathcal{K}_{ADV}) \right) \right) \le \max_{1\leq i\leq M-N} \left( \min_{\substack{1 \le j \le N \\ 1 \le l \le M-N}} \Phi\left( z_{ijl}(\mathcal{K}_{ADV}) \right) \right)$$
\end{proof}

\begin{corollary}[Attack Advantage]
\label{cor:advantage}
The advantage of the adversarial server $\mathcal{A}$, which is defined as $\Adv^{\textup{AMI}}(\mathcal{A}) = TPR - FPR$, is lower bounded by:
$$
\Adv^{\textup{AMI}}(\mathcal{A}) \ge 1 - \max_{1\leq i\leq M-N} \left( \min_{\substack{1 \le j \le N \\ 1 \le l \le M-N}} \Phi\left( z_{ijl} \right) \right)
$$
\end{corollary}

\subsection{On the upper bound of the per-sample False Positive Rate.} \label{appx:upper_noniid_bound}
In Fig. \ref{fig:fprbound_a}, we plot the upper bound of per-sample FPR across CIFAR10, CIFAR100, TinyImageNet, and FourDataset, using DeiT-B16 as the feature extractor after 60 training rounds. Following \citep{wang2022learning}, the global prompt pool size is set at $M=20$, and the prompt selection size at $N=4$. In practice, the upper bound for the per-sample False Positive Rate remains $\approx 0$ across all datasets, which corresponds to our $100\%$ per-sample attack success rate:


\begin{figure}[htbp]
     \centering
     \begin{subfigure}[b]{0.48\textwidth}
         \centering
         \includegraphics[width=\linewidth]{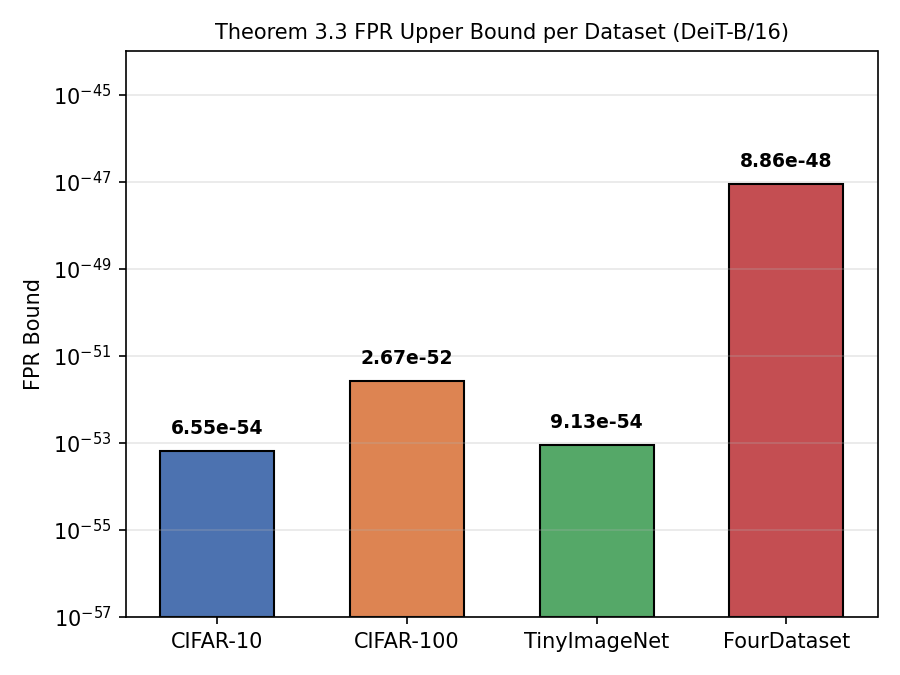}
         \caption{FPR bound on different datasets with $\alpha = 0.5$}
         \label{fig:fprbound_a}
     \end{subfigure}
     \hfill
     \begin{subfigure}[b]{0.48\textwidth}
         \centering
         \includegraphics[width=\linewidth]{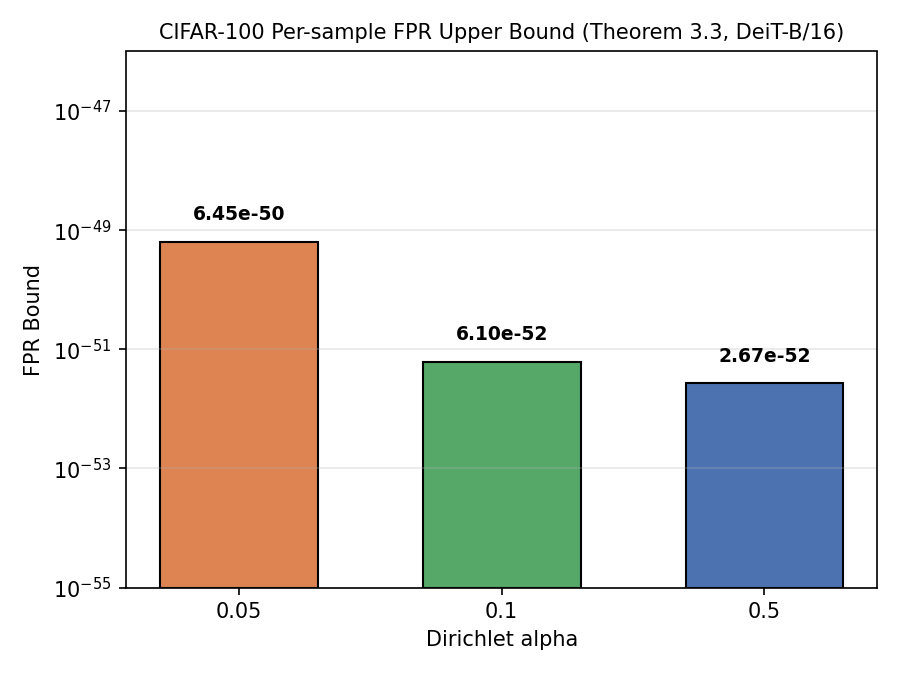}
         \caption{FPR bound on CIFAR-100 with varying $\alpha$}
         \label{fig:fprbound_b}
     \end{subfigure}
     \caption{Theorem~\ref{thm:fpr_main} FPR bound (DeiT-B/16, mean over 100 targets) under varying non-IID conditions.}
     \label{fig:fprbound_comparison}
\end{figure}

\subsection{\textsc{PromptMIA} under non-iid settings}
\label{appx:hetero}
 With non-IID datapoints, the theoretical bound remains tight as long as the prompt pool size ($M$) is sufficiently large to allocate keys to these non-IID points. To clarify, the tightness of the theoretical bound depends on the optimization dynamics and the capacity of the global prompt pool ($M$). Due to the FPT clustering optimization defined in Equation \ref{eq:trainl2p}, the surrogate loss actively pulls benign keys to align with the corresponding query features. As a result, when non-IID data regions exist, benign keys dynamically migrate to cover these regions. Therefore, as long as the prompt pool size ($M$) is sufficiently large to allocate keys to these distinct non-IID modes, the clusters remain locally compact and the bound is tight. On the other hand, if the number of benign keys is insufficient, a single key must represent a wider set of queries. That increases the variance $\sigma^2$, and then loosens the bound. In practice (Fig. \ref{fig:fprbound_b}), the per-sample FPR bound is still tight under extreme non-iid, which corresponds to our high per-sample Attack Success Rate.

\begin{figure}[!h]
    \centering
    \includegraphics[width=0.70\linewidth]{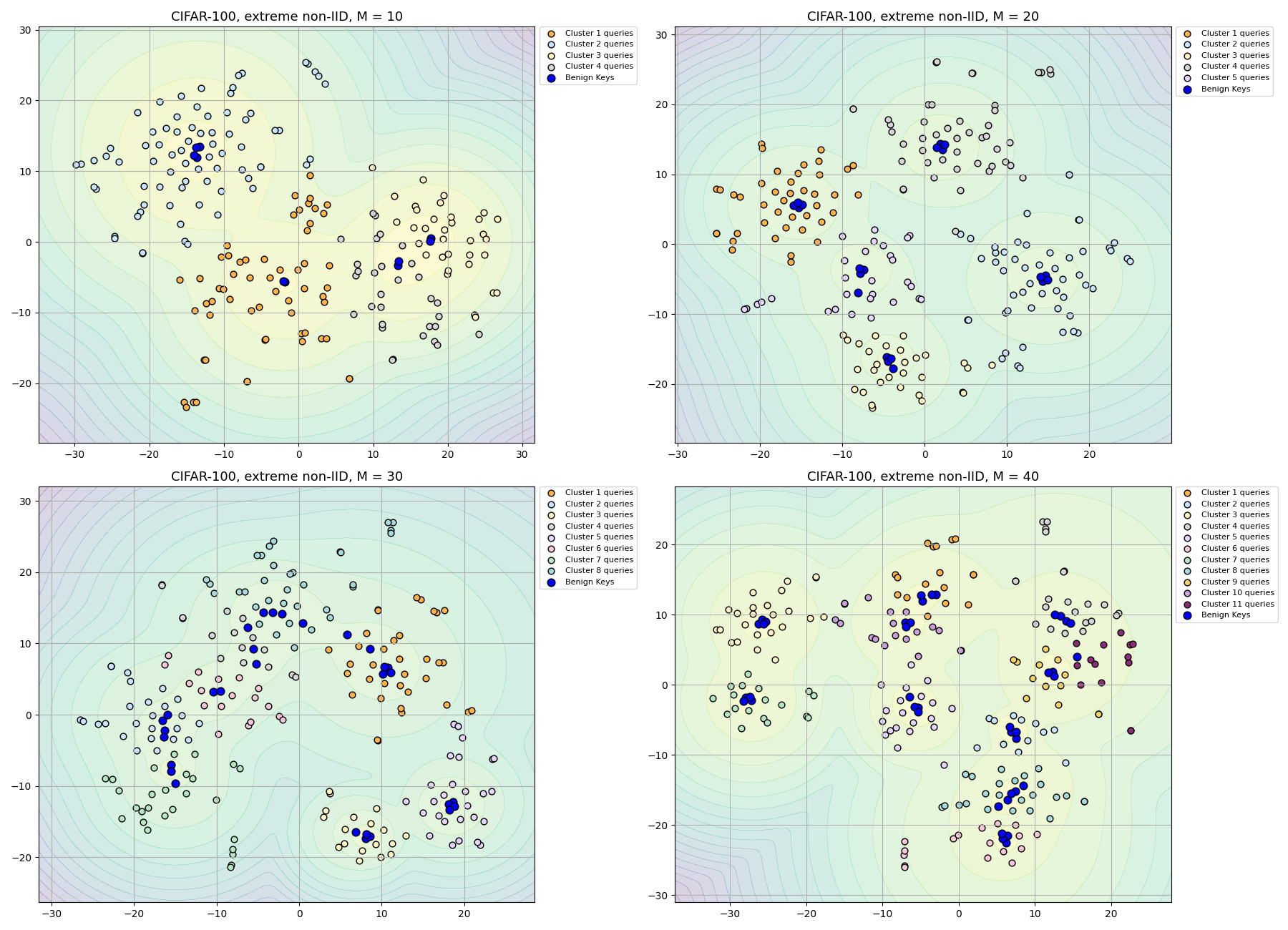}
    \caption{
        Gaussian mixture query model on CIFAR-100 under extreme non-IID with varying $M$.
        }
    \label{fig:gaussian_query_extreme}
\end{figure}

 In Fig. \ref{fig:gaussian_query_extreme}, we vary $M$ from 10 to 40 on TinyImageNet and CIFAR100 with ViT-B/32 under extreme non-IID settings (Fig.~\ref{fig:gaussian_query_extreme}). For the extreme non-IID setting, we follow prior work on federated learning with non-IID and imbalanced data \citep{weng2024probabilistic}: each client's training data is dominated by a subset of classes comprising 10\% of the total classes, with about 99\% of the client's local samples drawn from that subset. For each class, 99\% of its samples are split evenly among the clients dominated by that class, and the remaining 1\% are distributed among the other clients in Dirichlet($\alpha=1$) proportions, resulting in 80 clients with highly label-imbalanced local datasets.
In practice, the prompt pool size $M$ required to effectively assign keys to non-IID data points can be adaptively determined through a probabilistic aggregation mechanism \citep{weng2024probabilistic}. Even under extreme non-IID settings, the necessary $M$ remains sufficiently small ($M \le 40$) ( Fig. \ref{fig:gaussian_query_extreme}). Correspondingly, ASR increases as $M$ grows (Table~\ref{tab:pool_size_asr}).

\begin{table}[h!]
\centering
\begin{tabular}{ccccc}
\toprule
 & \multicolumn{2}{c}{CIFAR-100} & \multicolumn{2}{c}{TinyImageNet} \\
\cmidrule(lr){2-3} \cmidrule(lr){4-5}
$M$ & $B=64$ & $B=256$ & $B=64$ & $B=256$ \\
\midrule
10 & 98.0 & 91.4 & 93.4 & 74.0 \\
20 & 99.1 & 96.6 & 99.1 & 95.6 \\
30 & 99.4 & 97.5 & 99.4 & 97.6 \\
40 & 99.5 & 97.9 & 99.5 & 97.9 \\
\bottomrule
\end{tabular}
\caption{Attack success rate (\%) of \textsc{PromptMIA} under extreme non-IID settings for different prompt pool sizes $M$ and batch sizes $B$ (ViT-B/32).}
\label{tab:pool_size_asr}
\end{table}

\subsection{Advantage and Attack Success Rate of Individual models}
\label{appx:res_models}
\begin{figure}[h!]
    \centering
    \includegraphics[width=\textwidth]{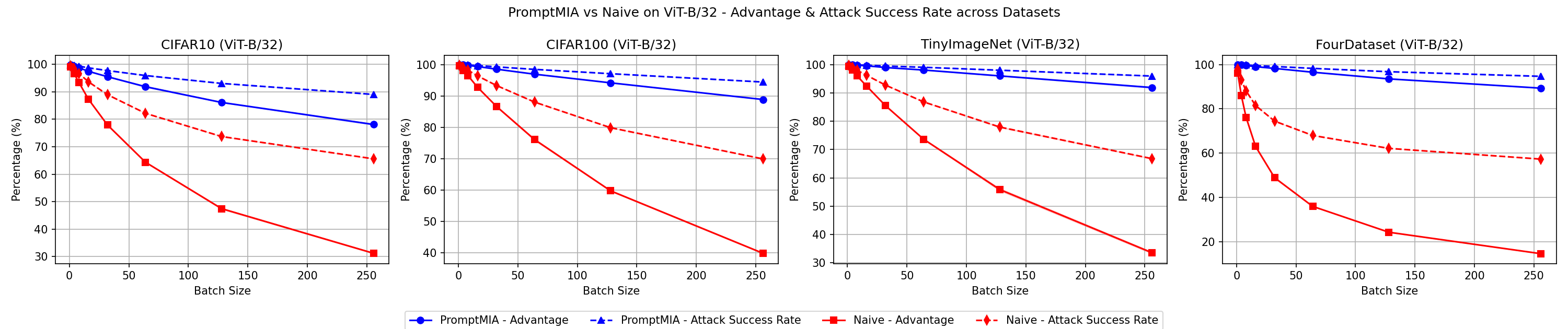}
    \caption{PromptMIA vs Naive attack results against ViT-B/32. 
             Each subplot shows Advantage and Attack Success rate w.r.t Batch Size 
             across CIFAR10, CIFAR100, TinyImageNet, and FourDataset.}
    \label{fig:promptmia_vs_naive_vit}
\end{figure}

\begin{figure}[h!]
    \centering
    \includegraphics[width=\textwidth]{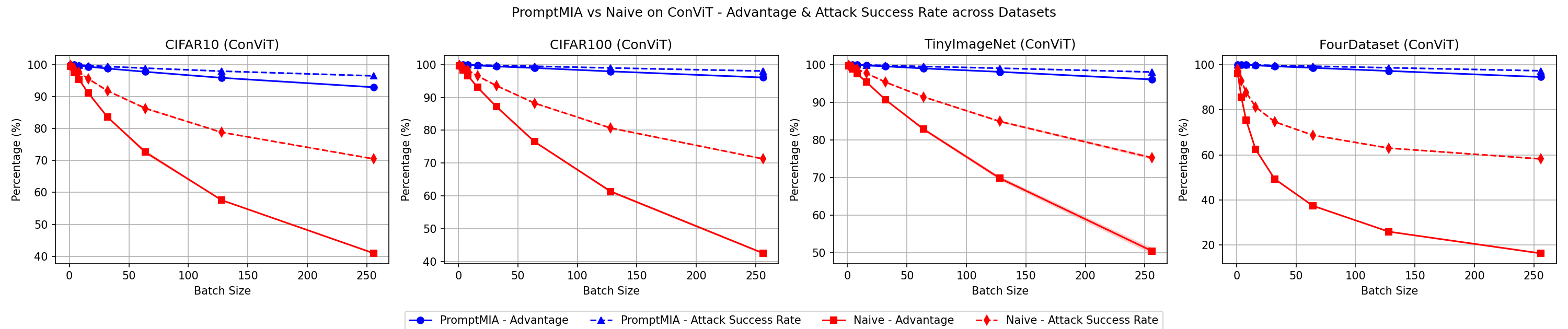}
    \caption{PromptMIA vs Naive attack results against ConViT. 
             Each subplot shows Advantage and Attack Success rate w.r.t Batch Size 
             across CIFAR10, CIFAR100, TinyImageNet, and FourDataset.}
    \label{fig:promptmia_vs_naive_convit}
\end{figure}

\begin{figure}[h!]
    \centering
    \includegraphics[width=\textwidth]{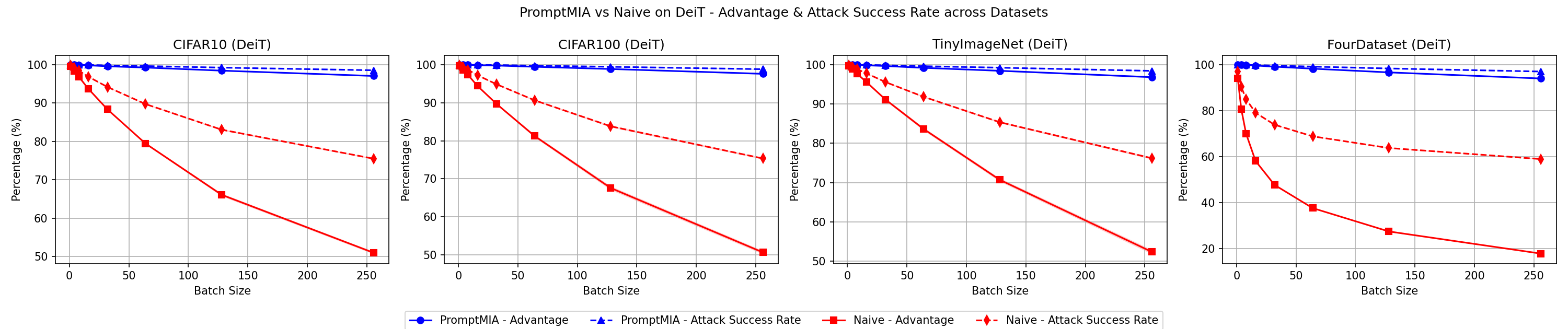}
    \caption{PromptMIA vs Naive attack results against Deit. 
             Each subplot shows Advantage and Attack Success rate w.r.t Batch Size 
             across CIFAR10, CIFAR100, TinyImageNet, and FourDataset.}
    \label{fig:promptmia_vs_naive_deit}
\end{figure}

Figures~\ref{fig:promptmia_vs_naive_vit}--\ref{fig:promptmia_vs_naive_deit} present a detailed comparison between our proposed \textsc{PromptMIA} and a naive membership inference baseline across three backbone models (ViT-B/32, ConViT, and DeiT) on all four datasets (CIFAR10, CIFAR100, TinyImageNet, and FourDataset). Each subplot reports both the Advantage and the Attack Success Rate (ASR) as a function of the client batch size. Results are averaged across 5 repeats of the security game on one trained model. 

\newpage
\subsection{Detailed results on outlier detection}
\label{appx:outlier}

\begin{figure}[h!]
  \centering
  \begin{subfigure}{0.24\textwidth}
    \includegraphics[width=\linewidth]{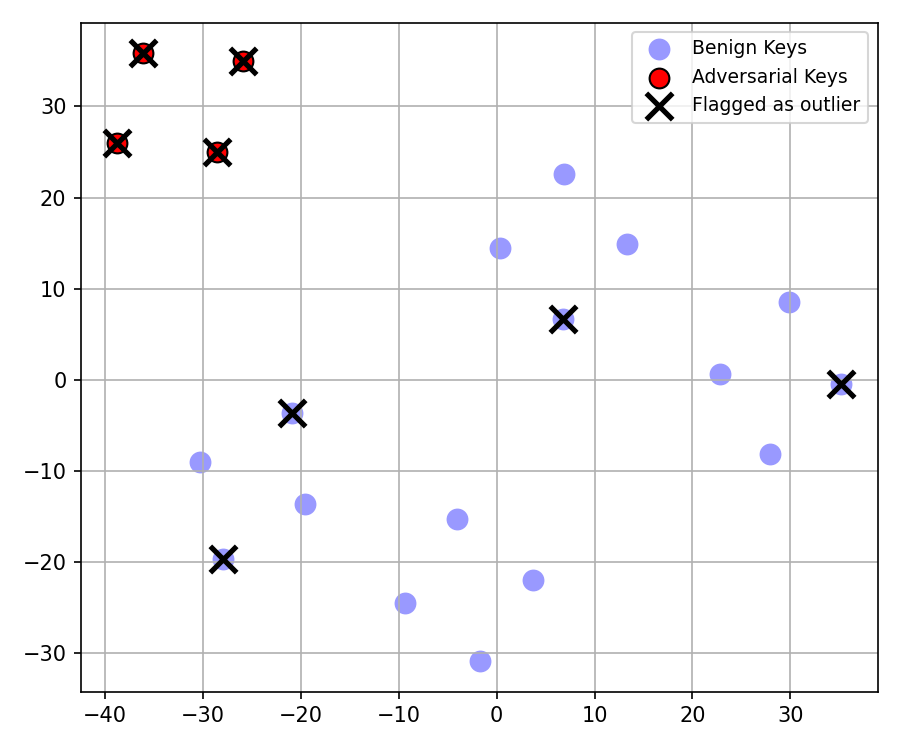}
    \caption{Isolation Forest}
  \end{subfigure}\hfill
  \begin{subfigure}{0.24\textwidth}
    \includegraphics[width=\linewidth]{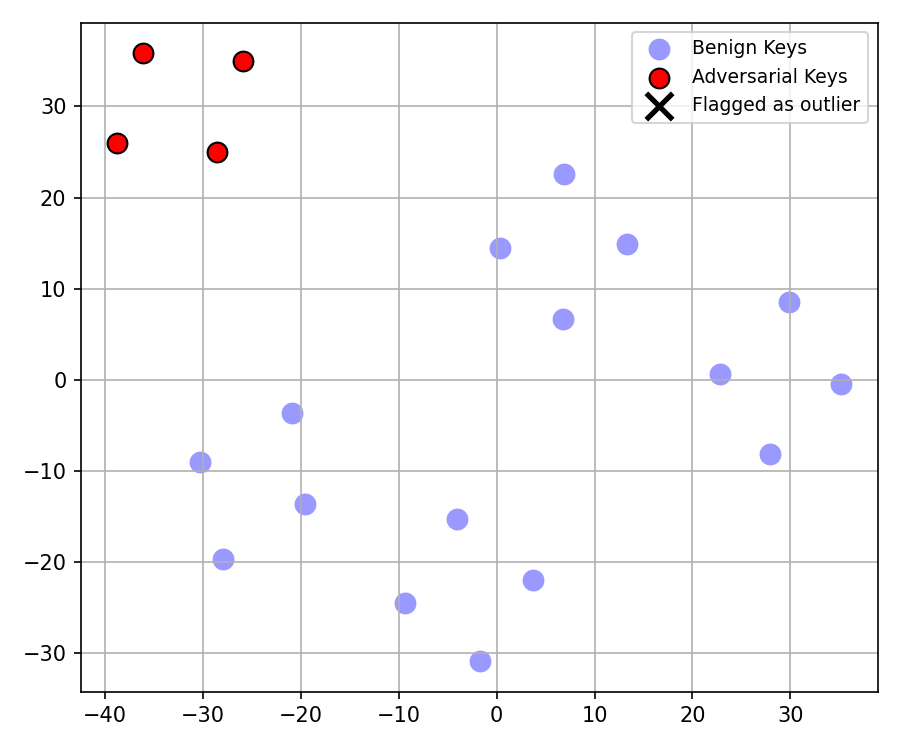}
    \caption{Local Outlier Factor}
  \end{subfigure}\hfill
  \begin{subfigure}{0.24\textwidth}
    \includegraphics[width=\linewidth]{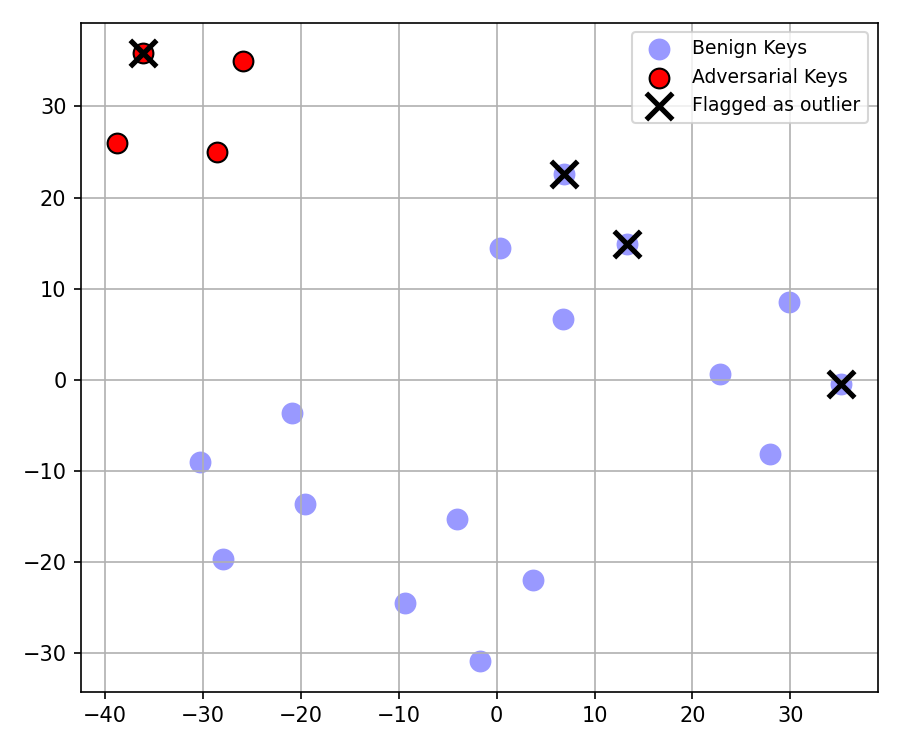}
    \caption{OneClassSVM}
  \end{subfigure}
  \begin{subfigure}{0.24\textwidth}
    \includegraphics[width=\linewidth]{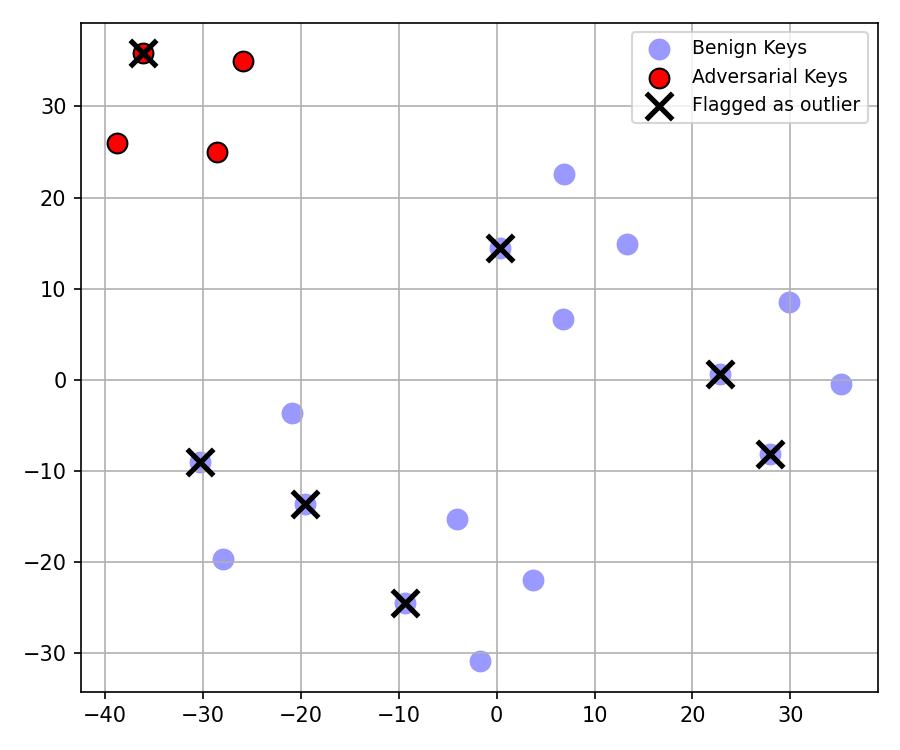}
    \caption{EllipticEnvelope}
  \end{subfigure}
  \caption{Visualization of outlier detection methods on CIFAR-10 trained Vit-B32. Blue keys are benign keys. Red keys are adversarial keys. Crossed keys are flagged as outliers. }
  \label{fig:outlier}
\end{figure}

\begin{figure}[h!]
  \centering
  \begin{subfigure}{0.24\textwidth}
    \includegraphics[width=\linewidth]{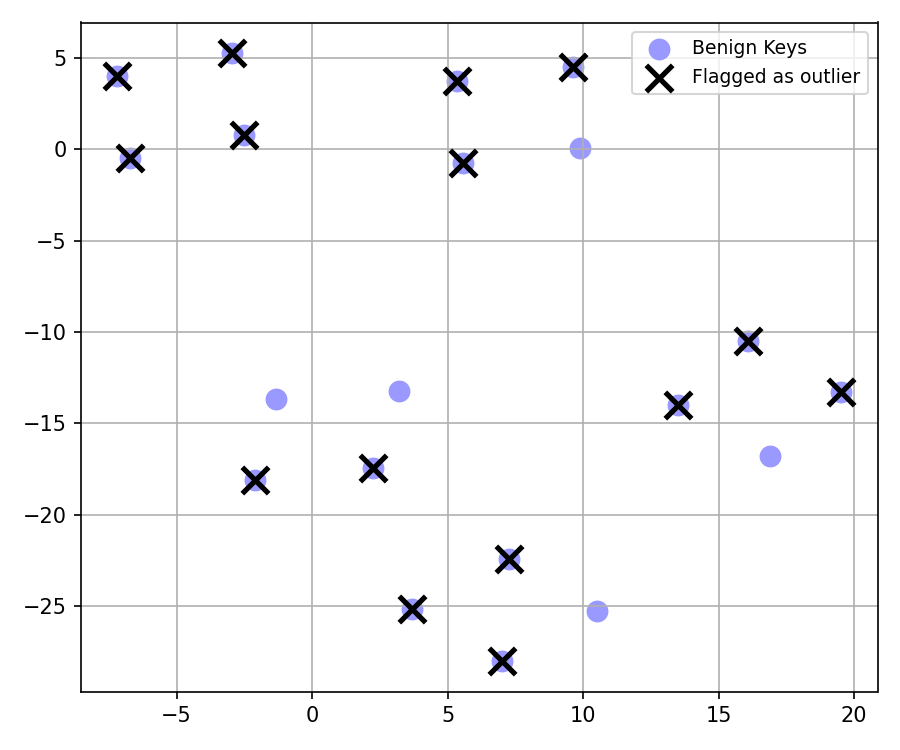}
    \caption{Isolation Forest}
  \end{subfigure}\hfill
  \begin{subfigure}{0.24\textwidth}
    \includegraphics[width=\linewidth]{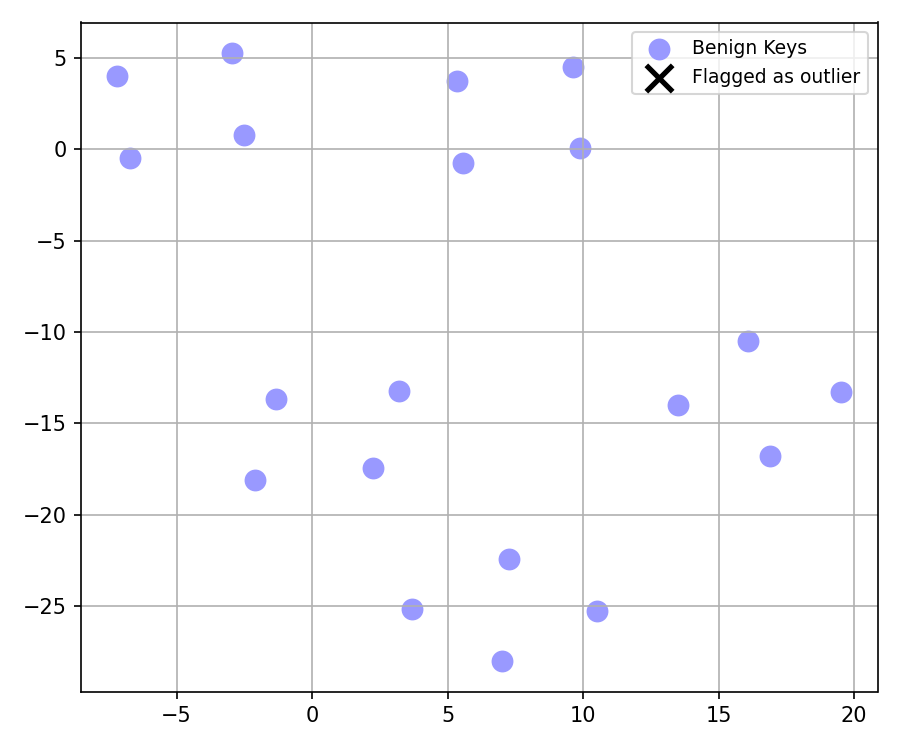}
    \caption{Local Outlier Factor}
  \end{subfigure}\hfill
  \begin{subfigure}{0.24\textwidth}
    \includegraphics[width=\linewidth]{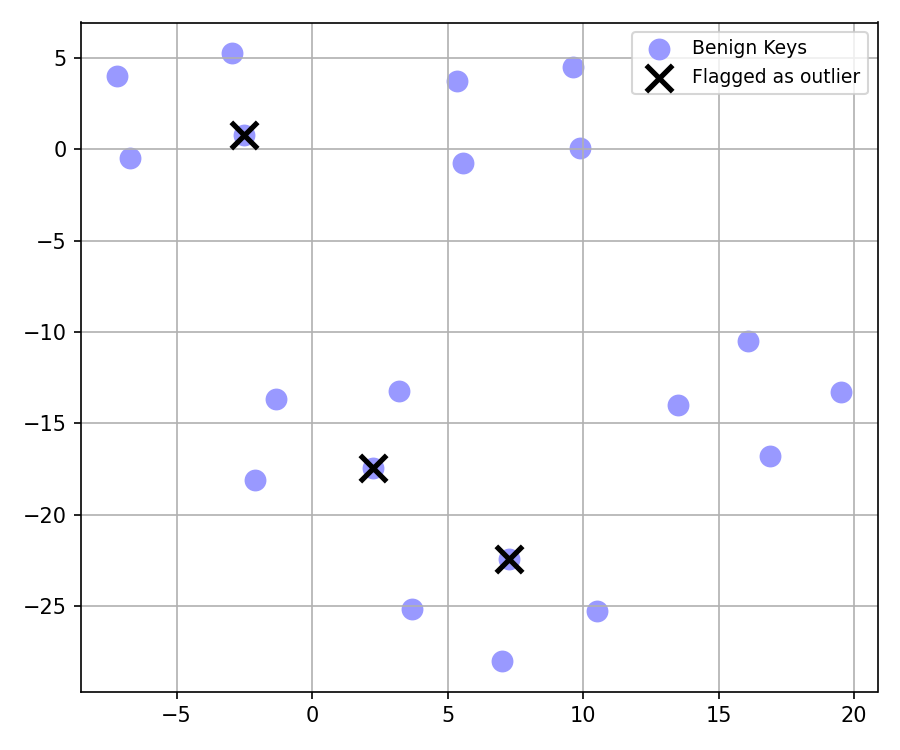}
    \caption{OneClassSVM}
  \end{subfigure}
  \begin{subfigure}{0.24\textwidth}
    \includegraphics[width=\linewidth]{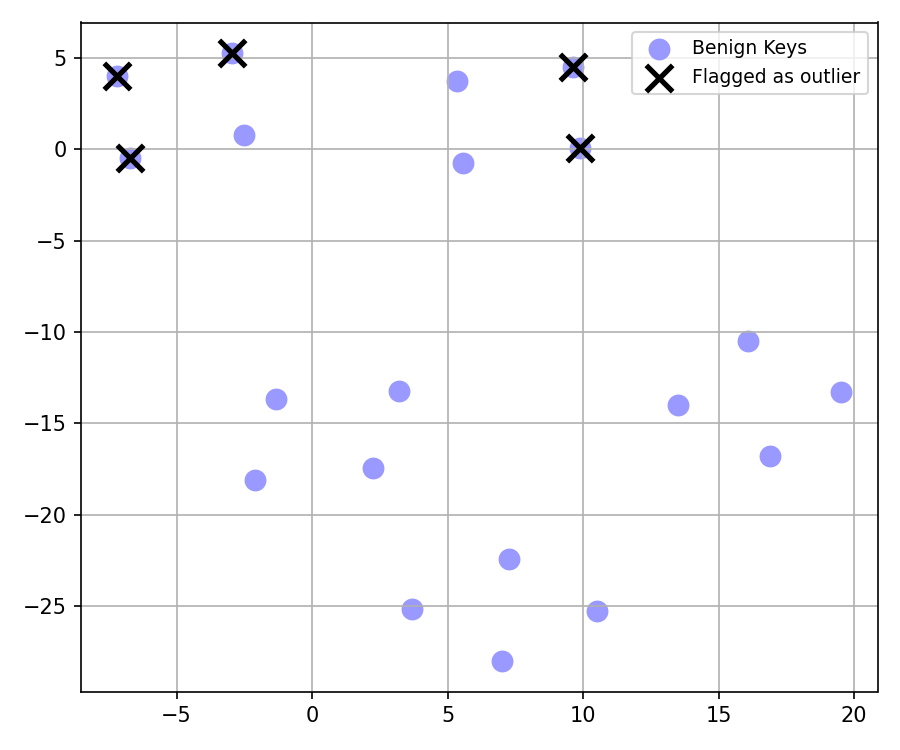}
    \caption{EllipticEnvelope}
  \end{subfigure}
  \caption{Visualization of outlier detection methods on CIFAR-10 trained ViT-B32. Blue keys are benign keys. Red keys are adversarial keys. Crossed keys are flagged as outliers from the corresponding algorithm. Outlier detection methods still falsely flag benign keys as outliers when no adversarial keys are present.}
  \label{fig:outlierbenign}
\end{figure}

Table~\ref{tab:outlier_all} reports the full precision, recall, and F1 scores of classical anomaly detectors applied to detecting adversarial keys in the global prompt pool, across datasets and backbones.We use an injection rate of $10\%$: each round is treated as attacked with probability $0.1$ and clean otherwise, and precision, recall, and F1 are computed from the expected counts, so clean rounds contribute the benign keys flagged on the un-injected pool as false positives. In practice, since \textsc{PromptMIA} identifies membership from a single injected round, the actual injection rate can be as low as one round out of the full schedule (e.g., $1/60 \approx 1.7\%$), “which further lowers precision at the same detector operating point.  Visualizations of these methods are given in Fig.~\ref{fig:outlier}. Moreover, these detectors consistently misclassify benign keys even when no adversarial keys are present (Fig.~\ref{fig:outlierbenign}).

\begin{table}[h!]
\centering
\scriptsize
\begin{tabular}{lllccc}
\toprule
Dataset & Model & Method & Precision & Recall & F1 \\
\midrule
\multirow{12}{*}{CIFAR10} 
 & ViT    & IsolationForest   & 0.0277 & 1.0000 & 0.0539 \\
 &        & LocalOutlierFactor & 0.0000 & 0.0000 & 0.0000 \\
 &        & OneClassSVM       & 0.0625 & 0.5000 & 0.1112 \\
 &        & EllipticEnvelope  & 0.0155 & 0.2275 & 0.0291 \\
 & ConViT & IsolationForest   & 0.0236 & 1.0000 & 0.0462 \\
 &        & LocalOutlierFactor & 0.0000 & 0.0000 & 0.0000 \\
 &        & OneClassSVM       & 0.0585 & 0.4925 & 0.1045 \\
 &        & EllipticEnvelope  & 0.0511 & 0.2875 & 0.0868 \\
 & DeiT   & IsolationForest   & 0.0461 & 1.0000 & 0.0881 \\
 &        & LocalOutlierFactor & 0.0000 & 0.0000 & 0.0000 \\
 &        & OneClassSVM       & 0.0690 & 0.5450 & 0.1225 \\
 &        & EllipticEnvelope  & 0.0182 & 0.2250 & 0.0337 \\
\midrule
\multirow{12}{*}{CIFAR100} 
 & ViT    & IsolationForest   & 0.0503 & 1.0000 & 0.0958 \\
 &        & LocalOutlierFactor & 0.0000 & 0.0000 & 0.0000 \\
 &        & OneClassSVM       & 0.0321 & 0.4800 & 0.0602 \\
 &        & EllipticEnvelope  & 0.0397 & 0.2200 & 0.0673 \\
 & DeiT   & IsolationForest   & 0.0377 & 1.0000 & 0.0727 \\
 &        & LocalOutlierFactor & 0.0000 & 0.0000 & 0.0000 \\
 &        & OneClassSVM       & 0.0440 & 0.4450 & 0.0801 \\
 &        & EllipticEnvelope  & 0.0195 & 0.2400 & 0.0360 \\
 & ConViT & IsolationForest   & 0.0507 & 1.0000 & 0.0965 \\
 &        & LocalOutlierFactor & 0.0000 & 0.0000 & 0.0000 \\
 &        & OneClassSVM       & 0.0334 & 0.4900 & 0.0625 \\
 &        & EllipticEnvelope  & 0.0352 & 0.2775 & 0.0625 \\
\midrule
\multirow{12}{*}{TinyImageNet} 
 & ViT    & IsolationForest   & 0.0215 & 1.0000 & 0.0421 \\
 &        & LocalOutlierFactor & 0.0000 & 0.0000 & 0.0000 \\
 &        & OneClassSVM       & 0.0379 & 0.4800 & 0.0702 \\
 &        & EllipticEnvelope  & 0.0186 & 0.2300 & 0.0343 \\
 & DeiT   & IsolationForest   & 0.0458 & 1.0000 & 0.0876 \\
 &        & LocalOutlierFactor & 0.0000 & 0.0000 & 0.0000 \\
 &        & OneClassSVM       & 0.0314 & 0.4625 & 0.0588 \\
 &        & EllipticEnvelope  & 0.0126 & 0.2125 & 0.0239 \\
 & ConViT & IsolationForest   & 0.0566 & 1.0000 & 0.1071 \\
 &        & LocalOutlierFactor & 0.0000 & 0.0000 & 0.0000 \\
 &        & OneClassSVM       & 0.0405 & 0.5075 & 0.0751 \\
 &        & EllipticEnvelope  & 0.0353 & 0.2775 & 0.0627 \\
\bottomrule
\end{tabular}
\caption{Precision, Recall, and F1 of outlier detection across datasets, models, and methods (per-model, pooled at a $10\%$ injection rate).}
\label{tab:outlier_all}
\end{table}

\newpage
\subsection{Hyperparameter Analysis}
\label{appx:hyperparam_analysis}
To isolate the effect of each hyperparameter, we vary one parameter at a time from the default setting ($M=20$, $N=4$, $\delta_{\min}=0.02$, $\Delta=0.05$, round 60). All experiments in this section use the ViT-B/32 model, and results are averaged over CIFAR-10, CIFAR-100, TinyImageNet, and FourDataset. Models are retrained for each value of $M$ and $N$, while $\delta_{\min}$ and $\Delta$ only change the attack on the trained default model.

\begin{figure}[ht]
  \centering
  \begin{subfigure}{0.3\textwidth}
    \includegraphics[width=\linewidth]{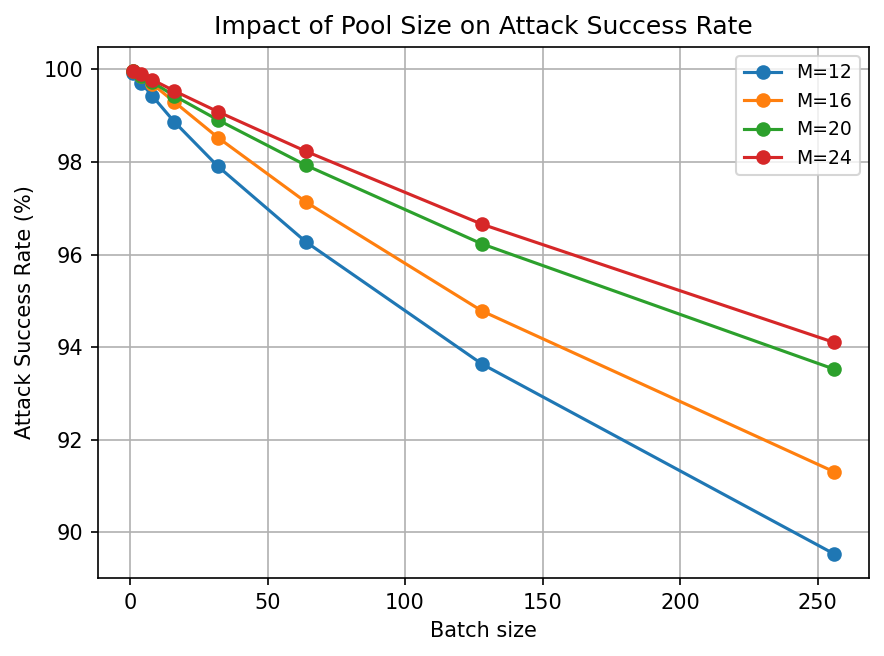}
    \caption{Impact of prompt pool size $M$}
    \label{fig:ablation_poolsize}
  \end{subfigure}\hfill
  \begin{subfigure}{0.3\textwidth}
    \includegraphics[width=\linewidth]{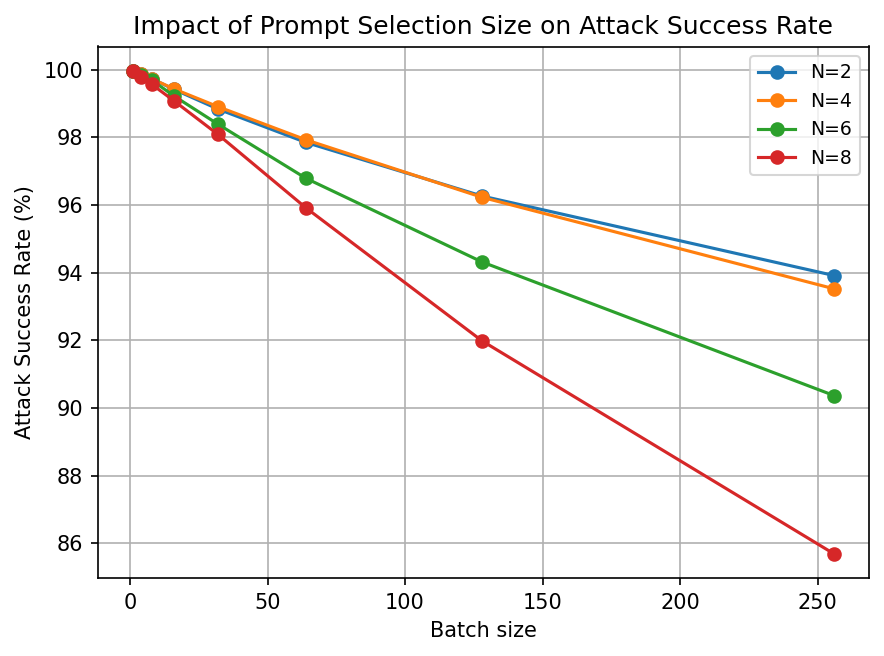}
    \caption{Impact of $N$}
    \label{fig:ablation_selectionsize}
  \end{subfigure}\hfill
  \begin{subfigure}{0.3\textwidth}
    \includegraphics[width=\linewidth]{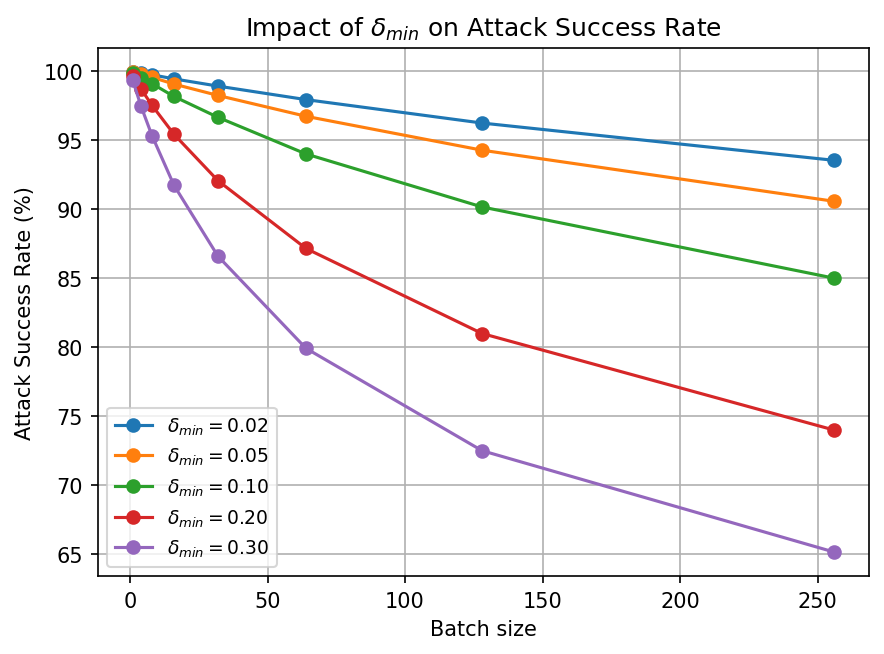}
    \caption{Impact of $\delta_{\min}$}
    \label{fig:ablation_deltamin}
  \end{subfigure}

  \begin{subfigure}{0.3\textwidth}
    \includegraphics[width=\linewidth]{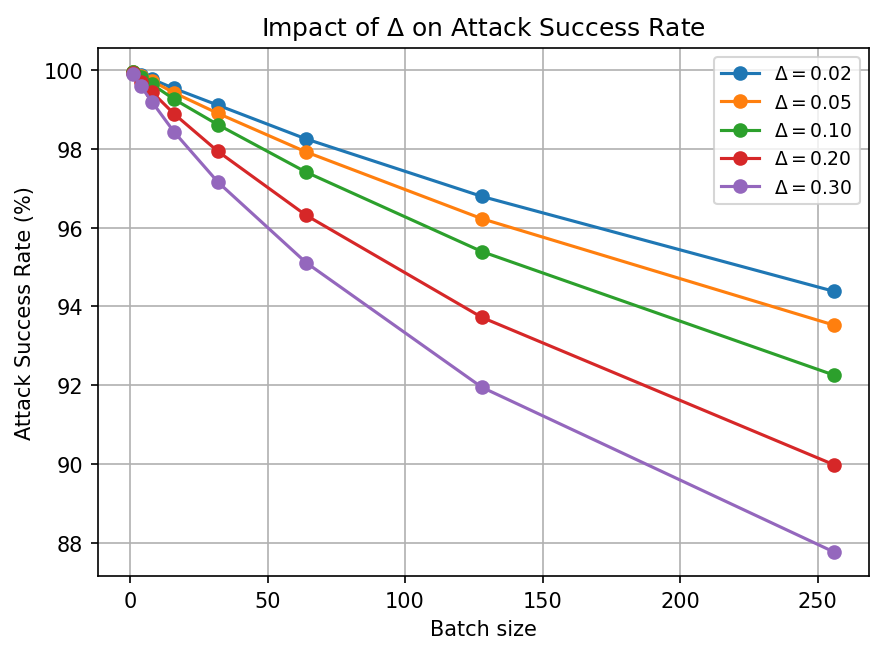}
    \caption{Impact of $\Delta$}
    \label{fig:ablation_delta}
  \end{subfigure}\hfill
  \begin{subfigure}{0.3\textwidth}
    \includegraphics[width=\linewidth]{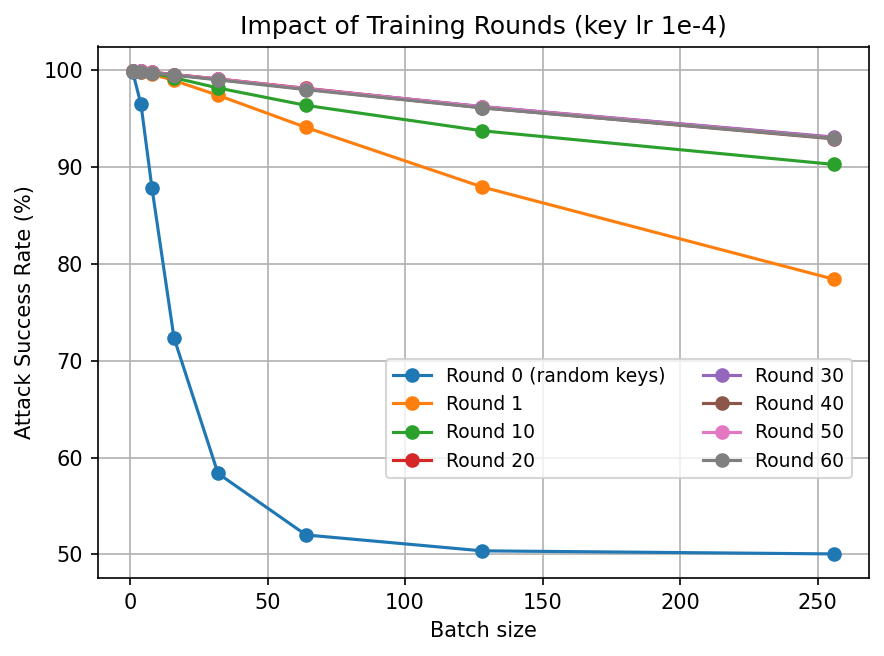}
    \caption{Impact of training rounds}
    \label{fig:ablation_traininground}
  \end{subfigure}\hfill
  \begin{subfigure}{0.3\textwidth}
   \includegraphics[width=\linewidth]{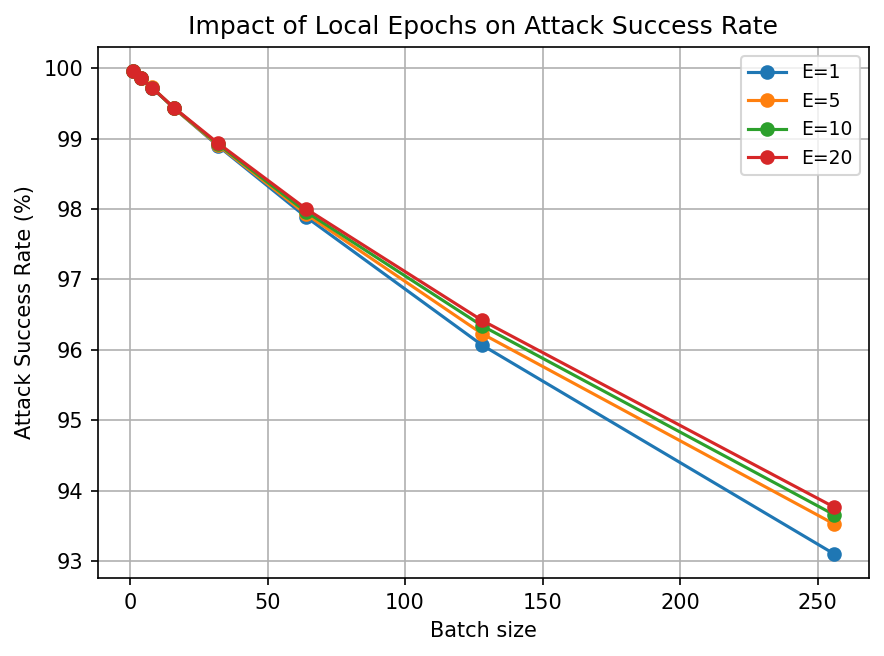}
   \caption{Impact of local epochs}
   \label{fig:ablation_localepochs}
  \end{subfigure}

  \caption{Ablation study on \textsc{PromptMIA}. 
  Each subfigure shows the effect of one parameter: (a) $M$, (b) $N$, (c) $\delta_{\min}$, (d) $\Delta$, (e) training rounds and f) local epochs.}
  \label{fig:ablation}
\end{figure}

\textbf{Global Prompt Pool size $M$:} Increasing the pool size strengthens the attack. 
Larger $M$ (e.g., $M=24$) yields consistently higher attack success rates across all batch sizes, while smaller pools (e.g., $M=12$) slightly weaken the attack as batch size grows. When the pool size increases, the probability of the adversarial keys being selected when $\mathcal{T} \notin \mathcal{D}$ decreases, decreasing FPR. Since the server is in control of the training protocol and the global prompt pool, they can choose the value $M$. See Fig. \ref{fig:ablation_poolsize}.

\textbf{Prompt selection size $N$:} 
Increasing $N$ slightly weakens the attack. Again, since the server controls the training protocol, they can also most likely dictate the choice of $N$. See Fig. \ref{fig:ablation_selectionsize}.

\textbf{Impact of $\delta_{\min}$:}
Without any defense, increasing \(\delta_{\min}\) reduces the attack success rate; therefore one may choose \(\delta_{\min}\) arbitrarily small (e.g.\ \(\delta_{\min}=0.02\)). See Fig. \ref{fig:ablation_deltamin}.

\textbf{Impact of $\Delta$:} 
Empirically, increasing \(\Delta\) reduces inference accuracy, and a relatively small \(\Delta\) does not make the attack detectable by traditional anomaly-detection methods. Therefore we choose \(\Delta\) to be modest (e.g.\ \(\Delta=0.05\)). However, \(\Delta\) should not be so small that the adversarial keys become indistinguishably close to one another. See Fig. \ref{fig:ablation_delta}.

\textbf{Impact of number of training rounds:} \textsc{PromptMIA} achieves higher ASR against models that have been trained for a few rounds compared to randomly initialized keys. This is reflected in our theorectical findings in Section \ref{sec:theory} that FPR is lower when the benign data forms tight and compact clusters in the query space around the benign keys, which happens naturally during the training process. We use a smaller key learning rate of $10^{-4}$ for this experiment to highlight per-round impact. See Fig. \ref{fig:ablation_traininground} and Fig. \ref{fig:trainround}. 

\textbf{Impact of number of local epochs:} We retrain the model with $E \in \{1, 5, 10, 20\}$ local epochs and attack the round-60 keys. The number of local epochs used in training has little effect on the attack. See Fig.~\ref{fig:ablation_localepochs}.

\subsection{Training Dynamics of Benign Keys}\label{appx:trainingdynamic}
\begin{figure}[h!]
  \centering
  \begin{subfigure}{0.24\textwidth}
    \includegraphics[width=\linewidth]{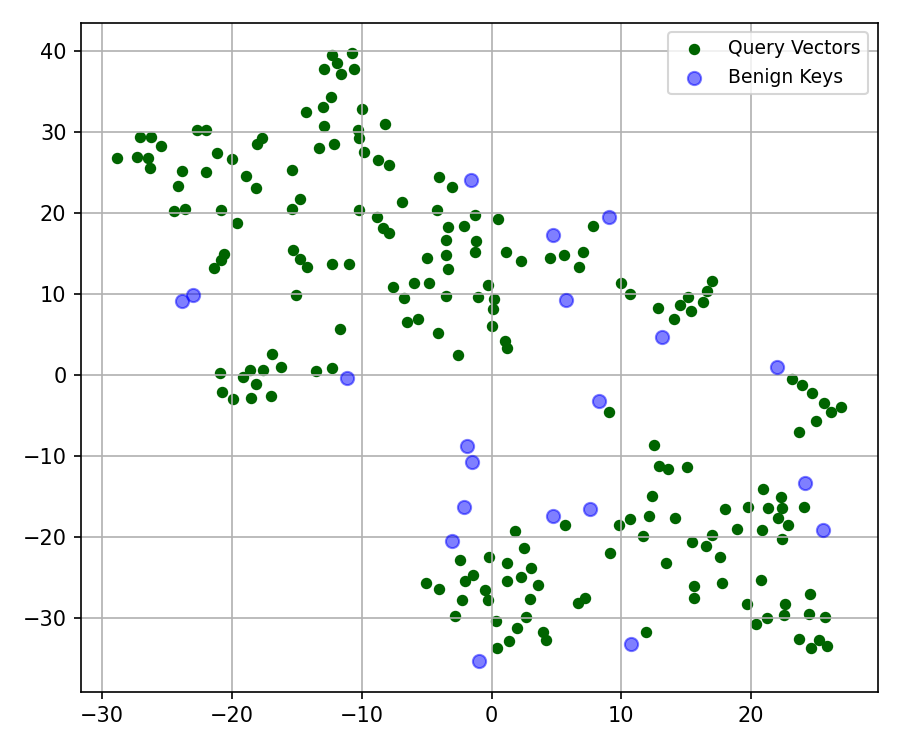}
    \caption{Random weight}
  \end{subfigure}\hfill
  \begin{subfigure}{0.24\textwidth}
    \includegraphics[width=\linewidth]{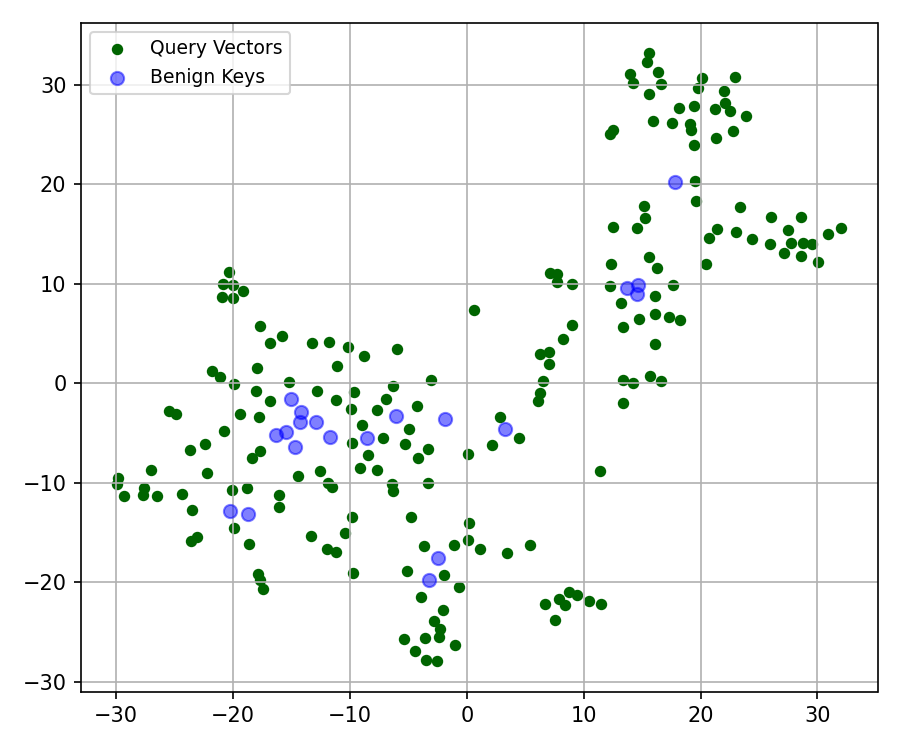}
    \caption{Round 1}
  \end{subfigure}\hfill
  \begin{subfigure}{0.24\textwidth}
    \includegraphics[width=\linewidth]{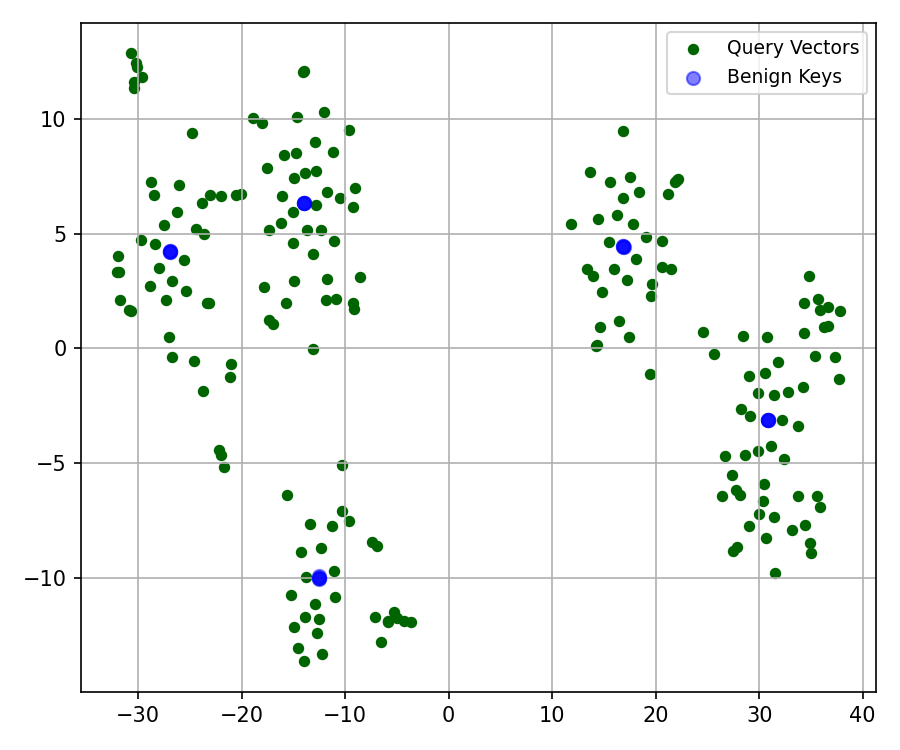}
    \caption{Round 20}
  \end{subfigure}
  \begin{subfigure}{0.24\textwidth}
    \includegraphics[width=\linewidth]{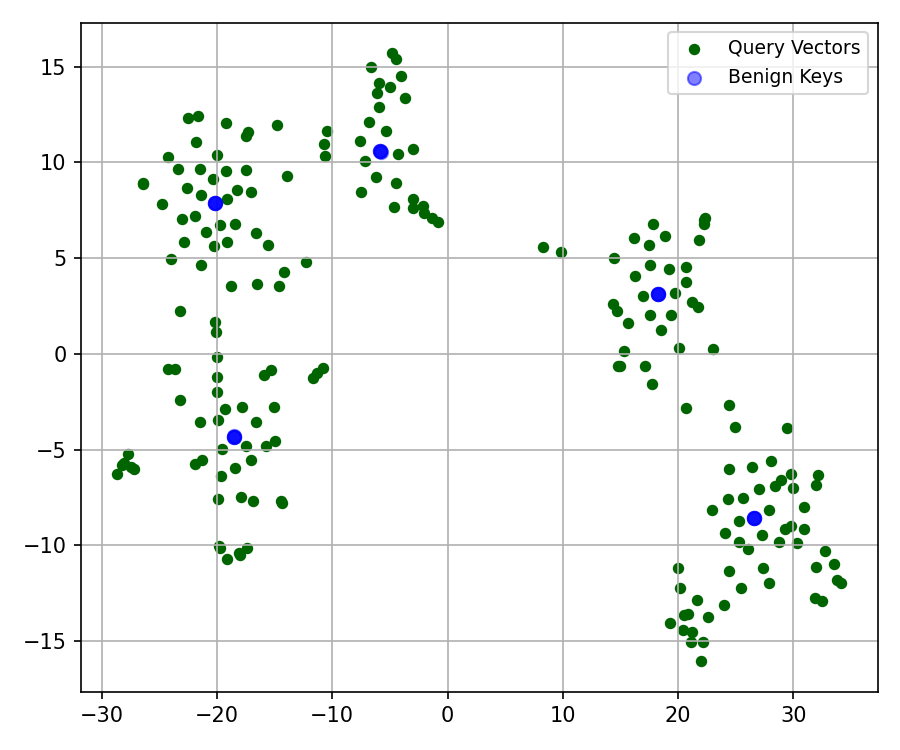}
    \caption{Round 60}
  \end{subfigure}
  \caption{Visualization of distribution of benign keys (blue) and query vectors $q(x)$ (green) across training rounds.}
  \label{fig:trainround}
\end{figure}

In federated prompt tuning, the keys in the global prompt pool gradually adapt to represent the distribution of clients' data. Early in training (Round~0), benign keys are randomly initialized and do not reflect the query feature space. As training proceeds, the selected keys are pulled toward their associated query vectors, causing the benign keys to migrate toward dense regions of the feature space. Over multiple communication rounds, these keys stabilize and effectively act as cluster centroids for groups of similar queries. Figure~\ref{fig:trainround} visualizes this process, showing how random keys become structured and aligned with the data distribution after the training process.
\subsection{Membership Inference under very large batch sizes}
\label{appx:larger_batch}
\begin{figure}[h!]
    \centering
    \includegraphics[width=1\linewidth]{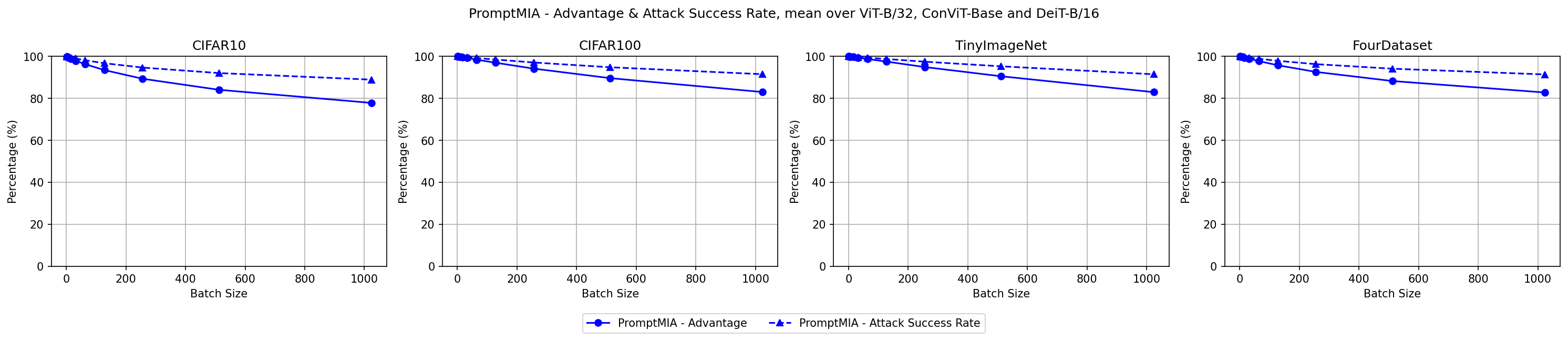}
    \caption{
        Advantage and attack success rate of \textsc{PromptMIA} under increasing batch size (up to 1024) across different datasets, averaged over ViT-B/32, ConViT, and DeiT. The attack success rate remains high even under extremely large batch sizes.
    }
    \label{fig:largebatch}
\end{figure}

To assess whether further increasing the batch size can mitigate membership leakage, we evaluate \textsc{PromptMIA} on the trained models of Fig.~\ref{fig:promptmia_vs_all_models} with client batch sizes of up to 1024. As seen in Fig.~\ref{fig:largebatch}, averaged over the three models, the attack success rate at batch size 1024 remains high. We also note that using such a batch size is often not possible in practical scenarios where low-resource edge devices cannot afford high VRAM consumption. Using a large batch size as a defense is therefore both ineffective and impractical.

\subsection{Performance of \textsc{PromptMIA} against text and multimodal data.}
\label{appx:modality}

\begin{figure}[h!]
    \centering
    \includegraphics[width=1\linewidth]{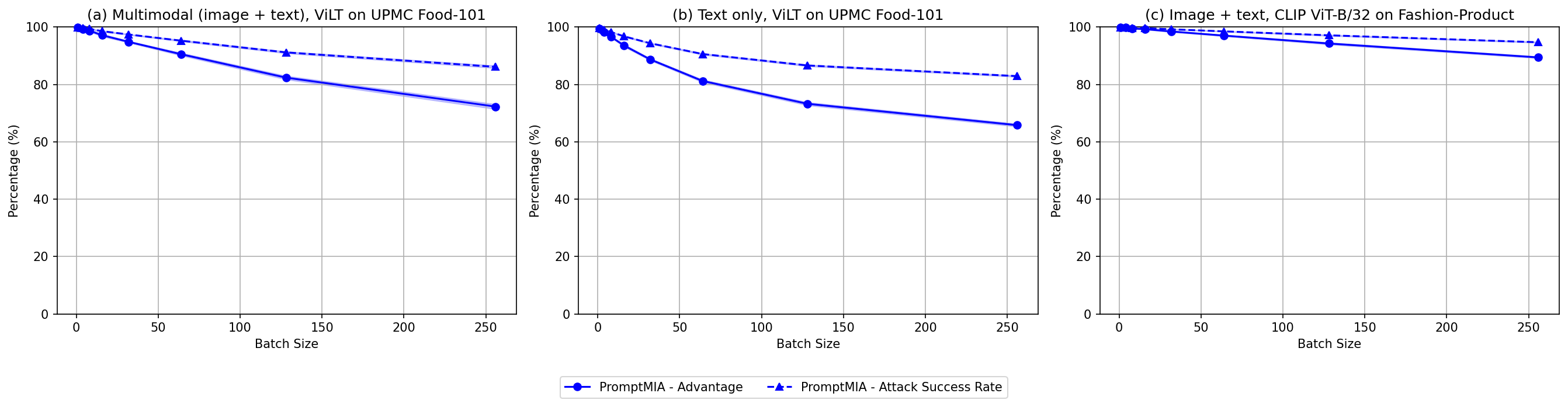}
    \caption{
        Advantage and attack success rate of \textsc{PromptMIA} under multimodal and text input modalities.
    }
    \label{fig:multimodal}
\end{figure}

To show that our attack is not restricted to the vision domain, we evaluate \textsc{PromptMIA} on multimodal and text data under the same paradigm of a frozen backbone paired with a shared, learnable prompt pool. We use the pretrained Vision-and-Language Transformer (ViLT) on an 8-class subset of UPMC Food-101 \cite{gallo2020image} with 20 clients, the max-pooled input embedding as the query \citep{phung2025federated}, and frequency-penalized prompt selection in training, in both multimodal and text-only configurations. We also use a frozen CLIP ViT-B/32 on Fashion-Product (images and titles) with 80 clients. See Fig.~\ref{fig:multimodal}.

\subsection{Impact of number of clients on \textsc{PromptMIA}}
\label{appx:numclients}

\begin{figure}[h!]
    \centering
    \includegraphics[width=0.8\linewidth]{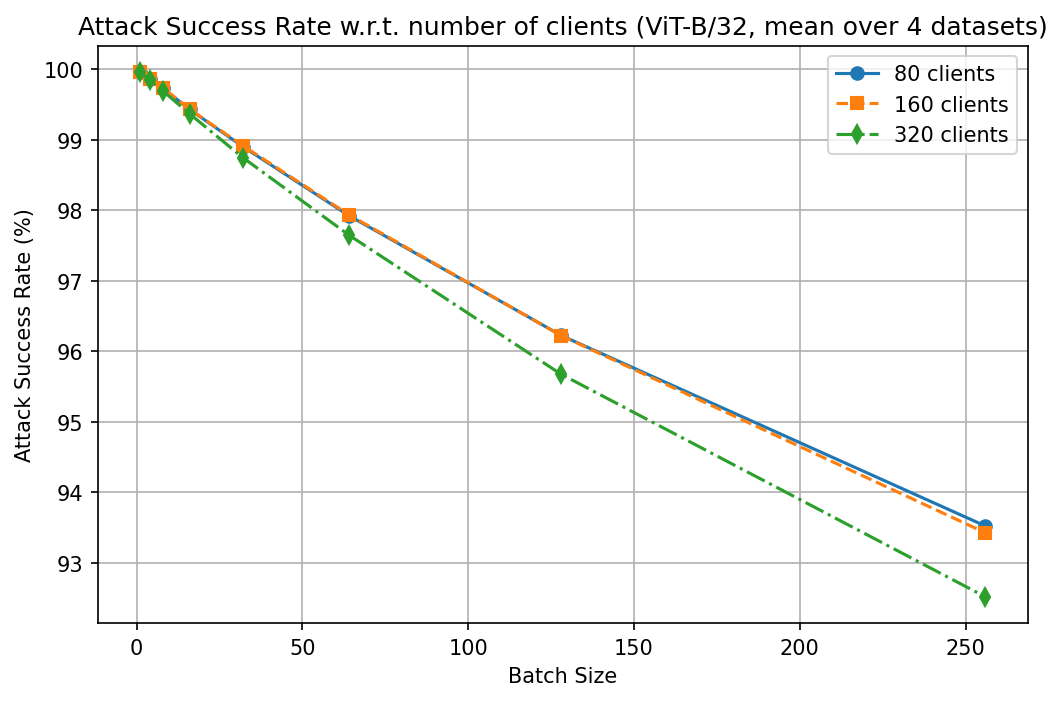}
    \caption{
        Impact of number of clients on \textsc{PromptMIA}. 
    }
    \label{fig:numclients}
\end{figure}

To further examine PromptMIA’s performance in larger-scale settings, we additionally evaluate it with up to 320 clients. In general, increasing the number of clients can slightly reduce the attack accuracy (Fig. \ref{fig:numclients}). This is due to increasing the number of clients can lead to slower convergence rate, making it harder for the benign key vectors to converge to true centroids of the benign data clusters. As a result, the cluster variance $\sigma^2$ in Lemma 3.2 increases, which in turn raises the FPR. However, the scalability issue w.r.t the number of clients is a general limitation of federated learning and is not specific to our attack.

\subsection{Impact of target distinctiveness on \textsc{PromptMIA}}
\label{appx:target_distinct}
We examine whether \textsc{PromptMIA} depends on the target being distinctive in the query space, using the round-60 keys of ViT-B/32 on the four datasets. We score each of $2{,}000$ random test targets by $s_{\max}=\max_{k_b}\kappa(q(\mathcal{T}),k_b)$, its similarity to the nearest benign key. A lower score means a more distinctive target. We group targets into quintiles of this score, from Q1 (most distinctive) to Q5 (least distinctive). As shown in Table~\ref{tab:target_distinct}, the FPR generally rises as targets become less distinctive, consistent with Theorem~\ref{thm:fpr_main}. Even so, for the least distinctive quintile at $B=256$, the ASR averaged over datasets remains $90.1\%$.

\begin{table}[h!]
\centering
\caption{FPR and ASR (\%) of \textsc{PromptMIA} by target distinctiveness ($s_{\max}$), averaged over the four datasets (ViT-B/32).}
\label{tab:target_distinct}
\small
\begin{tabular}{lcccc}
\toprule
 & \multicolumn{2}{c}{$B=64$} & \multicolumn{2}{c}{$B=256$} \\
\cmidrule(lr){2-3}\cmidrule(lr){4-5}
Group & FPR & ASR & FPR & ASR \\
\midrule
Q1 (most distinctive) & 1.58 & 99.21 & 4.72 & 97.64 \\
Q2 & 2.21 & 98.89 & 6.05 & 96.98 \\
Q3 & 3.10 & 98.45 & 8.16 & 95.92 \\
Q4 & 3.06 & 98.47 & 9.08 & 95.46 \\
Q5 (least distinctive) & 7.03 & 96.48 & 19.73 & 90.13 \\
\bottomrule
\end{tabular}
\end{table}

\subsection{Impact of \textsc{PromptMIA} on downstream task accuracy}
\label{appx:impact_acc}
Unlike prior MIAs, PromptMIA performs membership inference within a single communication round. As a result, an attacker could minimize the impact on model utility by launching the attack in one of the final training rounds. To provide a more rigorous evaluation, we inject the adversarial prompts at round 30 out of 60 total training rounds and measure the impact on the final model utility. For prompt injection rate, we set the number of injection prompts = N, where N is the number of prompts selected in top-N prompt selection mechanism. In general, increasing N reduces attack efficacy (Appendix \ref{appx:hyperparam_analysis}). Since the server controls the training protocol, they can also dictate the choice of N. Since prompt injection only modifies the key values and does not alter the prompt parameters, it has minimal impact on model accuracy:

\begin{table}[htbp]
\centering
\begin{tabular}{llcc}
\toprule
\textbf{Dataset} & \textbf{Model} & \textbf{Model Accuracy} & \textbf{Model Accuracy With PromptMIA} \\
\midrule
CIFAR10 & DeiT & 0.9480 & 0.9478 \\
FourDataset & ViT-B/32 & 0.6667 & 0.6654 \\
\bottomrule
\end{tabular}
\caption{Final test accuracy (round 60) with prompt injection at round 30.}
\label{tab:prompt_injection_accuracy}
\end{table}

\subsection{Additional Metrics}\label{appx:metrics}
\textbf{Attack Sparsity:} Table \ref{tab:model_sparsity} reports \textsc{PromptMIA}'s Attack Sparsity, which is defined as the percentage of the model parameters modified by PromptMIA. PromptMIA achieves very low attack spartsity of $\leq 0.0036 \%$ on all tested models.
\begin{table}[h!]
\centering
\begin{tabular}{lrrr}
\toprule
\textbf{Model} & \textbf{Total params} & \textbf{Modified params} & \textbf{Sparsity (\%)} \\
\midrule
vit\_b32     & 88,772,594 & 3,072 & 0.0035 \\
deit\_b16    & 86,228,234 & 3,072 & 0.0036 \\
convit\_base & 86,200,618 & 3,072 & 0.0036 \\
\bottomrule
\end{tabular}
\caption{Attack sparsity of \textsc{PromptMIA}}
\label{tab:model_sparsity}
\end{table}

\textbf{Computational Overhead:} The only overhead that \textsc{PromptMIA} incurs is the adversarial key generation process; all other procedures follow standard prompt-tuning training, so there is no additional overhead associated with maintaining a prompt pool for each dataset or client. Generating the $N=4$ adversarial keys of dimension $D=768$ against the $M-N=16$ benign keys requires approximately $122{,}000$ FLOPs (122 KFLOPs), about $0.003\%$ of a single ViT-B/32 forward pass on one image (4.4 GFLOPs). The overhead is therefore negligible.

\textbf{Scalability: } Our attack scales linearly w.r.t to the number of adversarial prompts ( $O(N)$) and the dimension of the prompt embedding ( $O(D)$). 	

\subsection{Limitations and Future Works}\label{appx:limitations}

We discuss the current limitations of \textsc{PromptMIA} as follows:
\begin{itemize}
    \item Unlike other MIAs designed for more general federated learning settings, \textsc{PromptMIA} specifically targets the federated prompt-tuning paradigm. As a result, its applicability is limited to prompt-tuning-based scenarios. However, since prompt tuning has become an increasingly common fine-tuning approach \citep{weng2024probabilistic, phung2025federated}, we believe this setting is both practical and important.
    \item \textsc{PromptMIA} relies on the server knowing the content of each individual client's updated prompt pool. Protocols that hide individual updates, such as secure aggregation via MPC/HE, therefore prevent a client-level inference, and the attack becomes a group-level membership inference attack. However, secure aggregation incurs high computational and communication costs that hinder practical deployment \citep{zhu2025fedmia}, and it does not extend to the probabilistic aggregation methods commonly used in federated prompt tuning \citep{weng2024probabilistic}.
    \item The attack requires an active, malicious server that alters and distributes modified keys, which is a stronger threat model than the honest-but-curious setting. Furthermore, concurrent tracking of multiple targets within a single round remains an open challenge.
    \item Our evaluation of defenses specifically designed against \textsc{PromptMIA} is not exhaustive, and we do not claim resistance to every future defense mechanism.
\end{itemize}
Based on these observations, one concrete future research direction is to design a secure aggregation-based protocol that works with probabilistic aggregation methods to effectively defend against prompt-based MIAs such as \textsc{PromptMIA}. A promising direction for tracking multiple targets within a single round is to craft, in place of $N$ keys tuned to one target, a set of $N$ keys that activates partially on any member of a set of target samples. To differentiate between the signals of these targets, each target can be assigned a distinct activation signature, so that membership is decoded from which adversarial keys were updated rather than from whether all $N$ were. Finally, angular or directional drift audits are a promising defense direction against \textsc{PromptMIA}.

\subsection{Comparision with broader active attacks}\label{appx:backdoor}
In general, other active attacks are orthogonal to Membership Inference attacks. The goal of MIA is to infer whether a specific target sample belongs to a client’s private dataset, whereas attacks such as backdoor injection or representation steering are typically designed to manipulate model behavior or internal representations rather than infer membership. We summarize the difference as follow:

\textbf{Active Membership Inference Attack}: We formalize the AMI threat models as in the standard settings of existing works \citep{vu2024analysis,carlini2022membership}. First, a randomly generated bit $b$ is used to determine whether the client's data $D$ contains a target sample $T$. If the protocol allows the server to decide the training architecture, the initialization step $A_{\text{INIT}}$ selects a model $\Phi$ for training. Otherwise, $A_{\text{INIT}}$ simply obtains information about the model from the protocol. Next, $A_{\text{ATTACK}}$ crafts the model parameters $\theta$ based on the target $T$ and the chosen architecture. Upon receiving $\Phi$ and $\theta$, the client computes the updated version $\dot{\theta}$ based on local training protocols and sends them to the server. Using $\dot{\theta}$, $A_{\text{GUESS}}$ outputs a guess for the value of $b$. Correctly inferring $b$ is equivalent to determining whether $T \in D$.

\vspace{1em}

\textbf{Backdoor attacks \citep{zhao2024survey}}: Let the standard training dataset be denoted by $D_{\text{train}}$. The attacker partitions $D_{\text{train}}$ into a clean subset $D_{\text{train}}^{\text{clean}}$ and a poisoned subset $D_{\text{train}}^{\text{poison}}$. The victim language model is then trained on the poisoned dataset $D_{\text{train}}^*$, yielding poisoned parameters:

$$
\theta_p = \arg\min_\theta \frac{1}{N} \sum_{i=1}^N [\mathcal{L}(f(x_i; \theta), y_i) + \mathcal{L}(f(x_i^*; \theta), y_{b,i})]
$$

where $D_{\text{train}}^* = \{ \{(x_i, y_i), (x_i^*, y_{b,i})\} \}_{i=1}^N$, $\mathcal{L}$ denotes the loss function, $\theta_p$ represents the poisoned model parameters,
$D_{\text{train}}^{\text{clean}}$ denotes clean samples, $x^* \in D_{\text{train}}^{\text{poison}}$ denotes poisoned samples containing the trigger, and $y_b$ is the attacker's target label. Through training, the model learns an association between the trigger and the target label, causing it to produce the attacker's intended output. During inference, if

$$
f(x^*; \theta_p) = y_b,
$$

then the backdoor attack is considered successful.

\vspace{1em}

\textbf{Representation steering \citep{wu2025improved,wu2025axbench}:} Representation steering aims to guide the output of an instruction-tuned language model toward a desired concept while still following the original instruction. Given an input instruction $x$ and a steering concept $c$, the goal is to generate a response $\hat{y}_i^c$ that both satisfies the instruction and incorporates the concept. Steering may be achieved through intervention-based methods or prompting-based methods. To support this task, a small training dataset is constructed as:

$$
D_{\text{Train}} = (x_i, y_i, y_i^c)_{i=1}^n,
$$

where each example consists of an instruction $x_i$, a standard response $y_i$, and a steered response $y_i^c$ containing the concept $c$. For intervention-based representation steering, the method modifies internal hidden representations of a Transformer during the forward pass. Let $h^l$ denote the sequence of $d$-dimensional representations at some model component in layer $l$. A steering intervention $\Phi_{\text{Steer}}$ edits these representations according to:

$$
h^l \leftarrow \Phi_{\text{Steer}}(\cdot; \alpha)
$$

where $\alpha$ is an optional parameter controlling the strength of the intervention.

\subsection{Broader Impacts}\label{appx:impact}

This research exposes critical privacy vulnerabilities in Federated Prompt Tuning (FPT). While federated parameter-efficient fine-tuning is gaining popularity, our work demonstrates that the exchange of soft prompts can leak sensitive client data, posing severe privacy risks if exploited by a malicious or compromised central server. 

 While publishing this methodology inherently carries dual-use risks, detailing \textsc{PromptMIA} is necessary to dispel the dangerous assumption that federated fine-tuning guarantees data protection. Because studied standard privacy defenses currently fail to mitigate this threat effectively, our findings serve as a call to action. There is an urgent need for the research community to design specialized, FPT-aware defense mechanisms (such as novel probabilistic Secure Aggregation protocols) to ensure the secure and ethical deployment of collaborative foundation models

\newpage
\section*{NeurIPS Paper Checklist}

\begin{enumerate}

\item {\bf Claims}
    \item[] Question: Do the main claims made in the abstract and introduction accurately reflect the paper's contributions and scope?
    \item[] Answer: \answerYes{} 
    \item[] Justification: Our contributions can be found in Sections \ref{sec:mia} and \ref{sec:fullexp}.

\item {\bf Limitations}
    \item[] Question: Does the paper discuss the limitations of the work performed by the authors?
    \item[] Answer: \answerYes{}{} 
    \item[] Justification: Our paper's limitations are clearly stated in Appendix \ref{appx:limitations}.

\item {\bf Theory assumptions and proofs}
    \item[] Question: For each theoretical result, does the paper provide the full set of assumptions and a complete (and correct) proof?
    \item[] Answer: \answerYes{} 
    \item[] Justification: All assumptions are given in Section \ref{sec:theory}. Complete proofs can be found in Appendix \ref{appx:proofs}.

    \item {\bf Experimental result reproducibility}
    \item[] Question: Does the paper fully disclose all the information needed to reproduce the main experimental results of the paper to the extent that it affects the main claims and/or conclusions of the paper (regardless of whether the code and data are provided or not)?
    \item[] Answer: \answerYes{} 
    \item[] Justification:  Our experiment protocols and hyperparameters are described in Section \ref{sec:fullexp} and Appendix \ref{appx:experiments}.

\item {\bf Open access to data and code}
    \item[] Question: Does the paper provide open access to the data and code, with sufficient instructions to faithfully reproduce the main experimental results, as described in supplemental material?
    \item[] Answer: \answerYes{} 
    \item[] Justification: All datasets used in this paper are public datasets. We release the code at \url{https://github.com/GivralNguyen/Prompt-MIA}.

\item {\bf Experimental setting/details}
    \item[] Question: Does the paper specify all the training and test details (e.g., data splits, hyperparameters, how they were chosen, type of optimizer) necessary to understand the results?
    \item[] Answer: \answerYes{} 
    \item[] Justification: Our experiment settings are described in Section \ref{sec:fullexp} and Appendix \ref{appx:experiments}.

\item {\bf Experiment statistical significance}
    \item[] Question: Does the paper report error bars suitably and correctly defined or other appropriate information about the statistical significance of the experiments?
    \item[] Answer: \answerYes{} 
    \item[] Justification:  We have provided standard deviations/ error bands for all main results.

\item {\bf Experiments compute resources}
    \item[] Question: For each experiment, does the paper provide sufficient information on the computer resources (type of compute workers, memory, time of execution) needed to reproduce the experiments?
    \item[] Answer: \answerYes{} 
    \item[] Justification: Details regarding computing resources are given in \ref{appx:experiments}.
    
\item {\bf Code of ethics}
    \item[] Question: Does the research conducted in the paper conform, in every respect, with the NeurIPS Code of Ethics \url{https://neurips.cc/public/EthicsGuidelines}?
    \item[] Answer: \answerYes{} 
    \item[] Justification: We have reviewed and do not violate the NeurIPS Code of Ethics.

\item {\bf Broader impacts}
    \item[] Question: Does the paper discuss both potential positive societal impacts and negative societal impacts of the work performed?
    \item[] Answer: \answerYes{} 
    \item[] Justification: We discuss the work's impact in Appendix \ref{appx:impact}.
    
\item {\bf Safeguards}
    \item[] Question: Does the paper describe safeguards that have been put in place for responsible release of data or models that have a high risk for misuse (e.g., pre-trained language models, image generators, or scraped datasets)?
    \item[] Answer: \answerNA{} 
    \item[] Justification: In this work, we do not introduce novel datasets or pre-trained models. All experiments rely exclusively on existing, publicly available datasets, their combinations, and established pre-trained NLP and vision models

\item {\bf Licenses for existing assets}
    \item[] Question: Are the creators or original owners of assets (e.g., code, data, models), used in the paper, properly credited and are the license and terms of use explicitly mentioned and properly respected?
    \item[] Answer: \answerYes{} 
    \item[] Justification: All datasets and pre-trained models used in our empirical evaluations are properly cited.

\item {\bf New assets}
    \item[] Question: Are new assets introduced in the paper well documented and is the documentation provided alongside the assets?
    \item[] Answer: \answerNA{} 
    \item[] Justification: We do not release new assests.

\item {\bf Crowdsourcing and research with human subjects}
    \item[] Question: For crowdsourcing experiments and research with human subjects, does the paper include the full text of instructions given to participants and screenshots, if applicable, as well as details about compensation (if any)? 
    \item[] Answer: \answerNA{} 
    \item[] Justification: Our paper does not involve crowdsourcing with human subjects.

\item {\bf Institutional review board (IRB) approvals or equivalent for research with human subjects}
    \item[] Question: Does the paper describe potential risks incurred by study participants, whether such risks were disclosed to the subjects, and whether Institutional Review Board (IRB) approvals (or an equivalent approval/review based on the requirements of your country or institution) were obtained?
    \item[] Answer: \answerNA{} 
    \item[] Justification: Justification: Our paper does not involve crowdsourcing with human subjects.

\item {\bf Declaration of LLM usage}
    \item[] Question: Does the paper describe the usage of LLMs if it is an important, original, or non-standard component of the core methods in this research? Note that if the LLM is used only for writing, editing, or formatting purposes and does \emph{not} impact the core methodology, scientific rigor, or originality of the research, declaration is not required.
    \item[] Answer: \answerNA{} 
    \item[] Justification: The core method development in this research does not involve LLMs as any important, original, or non-standard components.

\end{enumerate}
\end{document}